\documentclass{article}

\PassOptionsToPackage{numbers, compress}{natbib}
\usepackage{fix-cm}
\usepackage{titlesec}

\usepackage[main, final]{neurips_2026}

\usepackage[utf8]{inputenc} 
\usepackage[T1]{fontenc}    
\usepackage[colorlinks=true, citecolor=blue, linkcolor=blue, urlcolor=blue]{hyperref}       
\hypersetup{
  pdftitle={Spectral Prioritized Sweeping in Non-stationary Reinforcement Learning},
  pdfauthor={Gia-Hung Pham, Tuan Quang Dam}
}
\usepackage{url}            
\usepackage{booktabs}       
\usepackage{amsfonts}       
\usepackage{nicefrac}       
\usepackage{microtype}      
\usepackage{xcolor}         
\usepackage{amsmath,amssymb,amsthm}
\usepackage{mathtools}
\usepackage{algorithm}
\usepackage{algpseudocode}
\usepackage{enumitem}
\usepackage{tikz}
\usetikzlibrary{arrows.meta,positioning}
\usepackage{caption}
\usepackage{titletoc}
\newcommand\DoToC{%
  \startcontents
  \printcontents{}{1}{\textbf{Table of Contents}\vskip3pt\hrule\vskip5pt}
  \vskip3pt\hrule\vskip5pt
}

\BeforeBeginEnvironment{table}{\vspace{0.3cm}}
\AfterEndEnvironment{table}{\vspace{0.3cm}}
\BeforeBeginEnvironment{figure}{\vspace{0.3cm}}
\AfterEndEnvironment{figure}{\vspace{0.3cm}}
\BeforeBeginEnvironment{itemize}{\vspace{0.15cm}}
\AfterEndEnvironment{itemize}{\vspace{0.15cm}}
\BeforeBeginEnvironment{enumerate}{\vspace{0.15cm}}
\AfterEndEnvironment{enumerate}{\vspace{0.15cm}}

\newtheorem{assumption}{Assumption}
\newtheorem{proposition}{Proposition}[section]
\newtheorem{lemma}{Lemma}[subsection]
\newtheorem{theorem}{Theorem}[section]
\newtheorem{remark}{Remark}
\newtheorem{corollary}{Corollary}

\BeforeBeginEnvironment{lemma}{\vspace{0.15cm}}
\AfterEndEnvironment{lemma}{\vspace{0.15cm}}
\BeforeBeginEnvironment{corollary}{\vspace{0.15cm}}
\AfterEndEnvironment{corollary}{\vspace{0.15cm}}
\BeforeBeginEnvironment{assumption}{\vspace{0.15cm}}
\AfterEndEnvironment{assumption}{\vspace{0.15cm}}
\BeforeBeginEnvironment{definition}{\vspace{0.15cm}}
\AfterEndEnvironment{definition}{\vspace{0.15cm}}
\BeforeBeginEnvironment{remark}{\vspace{0.15cm}}
\AfterEndEnvironment{remark}{\vspace{0.15cm}}

\newcommand{\calS}{\mathcal{S}}
\newcommand{\calA}{\mathcal{A}}
\newcommand{\calM}{\mathcal{M}}
\newcommand{\bbR}{\mathbb{R}}
\newcommand{\bbE}{\mathbb{E}}
\newcommand{\norm}[1]{\left\lVert#1\right\rVert}
\newcommand{\abs}[1]{\left|#1\right|}
\newcommand{\Lout}{L^{\mathrm{out}}}
\newcommand{\Lin}{L^{\mathrm{in}}}
\newcommand{\fout}{\mathbf{f}^{\mathrm{out}}}
\newcommand{\fin}{\mathbf{f}^{\mathrm{in}}}
\newcommand{\slem}{\mathrm{SLEM}}
\newcommand{\prio}{\mathrm{priority}}
\newcommand{\lemref}[1]{Lemma~\ref{#1}}
\newcommand{\propref}[1]{Prop.~\ref{#1}}
\newcommand{\thmref}[1]{Thm.~\ref{#1}}
\newcommand{\corref}[1]{Cor.~\ref{#1}}

\title{Spectral Prioritized Sweeping in \\
  Non-stationary Reinforcement Learning}

\author{%
  Gia-Hung Pham \\
  SOICT, Hanoi University of Science and Technology \\
  \texttt{hungpg230036@sis.hust.edu.vn} \\
  \And
  Tuan Quang Dam \\
  SOICT, Hanoi University of Science and Technology \\
  \texttt{tuandq@soict.hust.edu.vn} \\
}

\begin{document}

\maketitle

\begin{abstract}
Prioritized Sweeping (PS) accelerates model-based reinforcement learning by selecting backups according to residual magnitude. In nonstationary reward settings, however, the canonical priority score is shortsighted: after a localized reward shift, residuals propagate only through realized backups, so clogged or topologically distant state estimates may become idle under a limited replanning budget. We introduce \textbf{Spectral Prioritized Sweeping} (SPS, also known as \textbf{Graph Topology Augmentation for Prioritized Sweeping}, GTA-PS), a drop-in ordering score for model-based planning under fixed dynamics and changing state rewards. Grounded in our broader \textbf{Graph Topology Augmentation} (GTA) framework, SPS utilizes the graph resolvent and directed Laplacian spectrum to diffuse residual information bidirectionally across the policy-induced transition chain. The topology weight is modulated by a scheduler based on the Second Largest Eigenvalue Modulus (SLEM), allowing priority propagation to adapt to the chain's spectral mixing regime. We prove that the forward potential coincides with discounted residual propagation at a canonical regularization parameter, and show that SPS gives nonzero priority to states that standard PS can leave blocked after sparse reward shifts. Tabular experiments on FourRooms and GARNET demonstrate improved replanning efficiency over vanilla PS under both exact DP and Dyna-style host planners.
\end{abstract}

\section{Introduction}
\label{sec:intro}

Model-based reinforcement learning improves general behavior by using a known or
learned transition model for planning
\citep[Ch.~8]{sutton2018reinforcement}. The agent benefits most when planning effort can be directed to the states that matter most, especially within a brittle budget \citep{sutton1990dyna,janner2019trust}.
\textbf{Prioritized Sweeping} addresses this by maintaining a
queue of states keyed by Bellman or Temporal-Difference residual magnitude \citep{moore1993prioritized,peng1993efficient}.
For each planning step, the largest-priority state has its value estimate updated.
After that, predecessors' value estimates often increase in magnitude; their refreshed priorities have higher chance to be queried next.
This effect encourages information propagation through the predecessor support direction.
Although conceptually simple, the resulting sweep can improve
update efficiency by several orders of magnitude compared to uniform random backups \citep{wingate2005prioritization}.

We focus our study in the nonstationarity case, where the transition dynamics are fixed but the state reward changes
over time \citep{padakandla2020reinforcement}. This setting appears whenever the navigation structure is comparatively
stable but the task objective changes, as in goal-conditioned robotics
\citep{plappert2018multigoal}, dynamic pricing and logistic operations
\citep{aviv2005pomdp}, and sequential recommendation model
\citep{zou2019engagement}.

The standard approach using residual priority is both \emph{shortsighted} and \emph{reactive}, as most state
order remain stale until the discrepancy wave has propagated to them through realized backups.
Under scarce number of updates between reward shifts, some topologically distant states may
never be reached, regardless of the host planner \citep{sutton2018reinforcement}.
This creates a hindsight effect: by the time the residual signal reaches distant regions, the reward information that triggered it may already be stale, and the resulting value errors feed back into the planning--action loop as outdated priorities \citep{cheung2020nonstationary, khetarpal2022continual}.

\begin{figure}[t]
  \centering
  \includegraphics[width=0.85\linewidth]{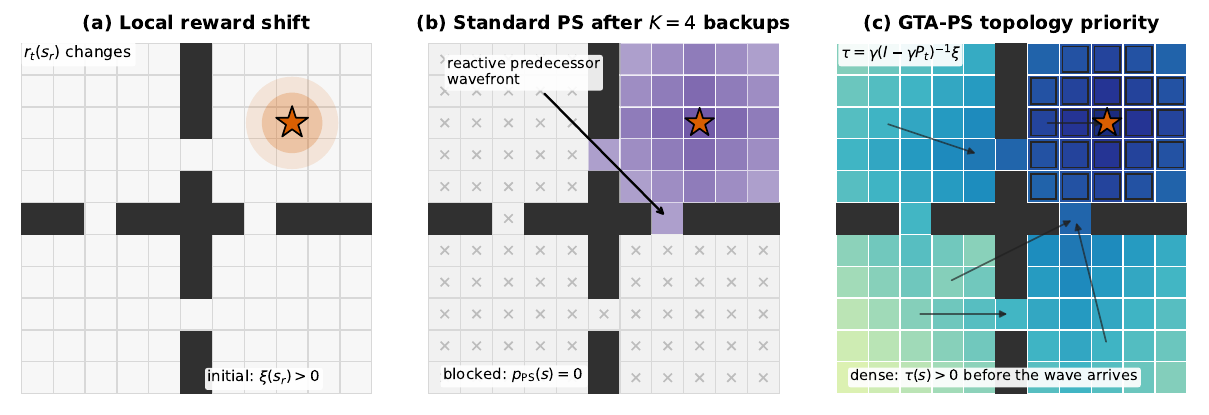}
  \caption{
    \footnotesize
    A localized reward shift creates sparse residual at $s_r$. Standard PS is
    shortsighted: priority of distant states receives no information hints.
    Spectral Prioritized Sweeping (SPS / GTA-PS) adds a topology potential on the policy-induced graph, assigning nonzero
    topology scores to upstream states before predecessor pushes reach them.
  }
  \label{fig:gta-frontpage-story}
\end{figure}

We propose to address the credit propagation bottleneck by utilizing the transition model as spectral assignment guidance.
We introduce the \textbf{Graph Topology Augmentation} (GTA) framework, which captures the structure of the agent's navigation field via directed spectral operators. This yields rich representations and structural preconditioning for score-based decision making. Our primary application, \textbf{Spectral Prioritized Sweeping} (SPS, also termed \textbf{GTA-PS}), instantiates the GTA principle for nonstationary replanning by augmenting the residual-keyed priority with an additive mixed potential built from regularized directed Laplacian inverses. The method is host-agnostic and drop-in: it intervenes exclusively in the priority ordering of planning backups, leaving the underlying planner architecture untouched.

\paragraph{Reading guide.}
Section~\ref{sec:related} positions the method against successor-based, abstraction-based, and graph-structural planners. Section~\ref{sec:gta-ps} develops the spectral topology signal, states the residual gate, and formalizes the SPS priority score; Section~\ref{sec:theory} gives the theoretical analysis; Section~\ref{sec:experiments} reports empirical benchmarks, ablations, and post-review audits.

\paragraph{Our contributions.}
\begin{itemize}
    \item We formalize priority-based planning as a host-agnostic queue-management problem,
    making the priority definition a modular design choice. The same SPS module plugs
    into both exact tabular (DP) and Dyna-style planners without modifying the host.
    \item We introduce \textbf{Graph Topology Augmentation} (GTA), a general RL framework
    that applies spectral operators on policy-induced chains to produce topology-aware
    planning signals.
    \item We present \textbf{Spectral Prioritized Sweeping} (SPS, or \textbf{GTA-PS}), a concrete realization of the GTA framework for nonstationary rewards. It
    augments residual priority with a directed regularized-Laplacian potential and a
    fixed forward/backward weight aggregation, together with a SLEM-driven $\beta_t$
    scheduler (the SPS$^\beta$ / GTA-PS$^\beta$ variant) that reads SLEM as a global ventilation-confidence
    signal rather than a per-state bottleneck detector.
    \item We prove key properties of the topology signal, including its numerical stability,
    the geometric discount and implicit importance sampling effects, plus the
    instance-dependent extent to which the dense signal can reach beyond the reactive
    wavefront.
    \item We report a focused experimental study, varying chain statistics across environments and planner architectures, with surgical ablations isolating each cause's effectiveness, followed by post-review audits: an equal-wall-clock comparison, a partitioned GPS-H2 baseline, a compact-abstraction study, and frontier and cutoff diagnostics.
\end{itemize}

\section{Related Work}
\label{sec:related}

In model-based planning, three lines of work provide richer guidance than raw residual magnitude.
\emph{Successor-based signals} spread value information over the state graph rather than waiting for local error cascades
\citep{dayan1993improving,stachenfeld2017hippocampus}.
\emph{Learned abstractions} compress the planning space, as in the abstract tabular model of \emph{VaST} \citep{corneil2018vast}.
\emph{Graph-structural heuristics} exploit an auxiliary graph representation, as in the strongly connected component schedule of \emph{TVI/FTVI} \citep{dai2011topological}. Together they show that good planning requires more than raw residuals and that model knowledge is a powerful source of guidance.

Spectral graph theory has a long history of applications in RL representation learning.
\emph{Proto-value functions} \citep{mahadevan2005proto,mahadevan2007proto} use undirected
Laplacian eigenvectors as smooth value-approximation bases
\citep{shuman2013emerging,belkin2003laplacian}; the same eigenvectors underpin \emph{option
discovery} \citep{machado2017laplacian}, its \emph{successor-representation} analogue \citep{machado2018eigenoption}, and \emph{cover-time options}
\citep{jinnai2019discovering}. Scalable computation uses \emph{Krylov methods} such as Arnoldi and GMRES \citep{golub2013matrix, saad1986gmres},
\emph{graph-drawing objectives} with unique solutions \citep{wang2021laplacian},
and recent \emph{exact recovery methods} \citep{gomez2024proper}; \emph{effective resistance} has also been proposed as an intrinsic reward
\citep{chauhan2026effective}. In these works, spectral operators serve offline representation or filtering rather than online backup selection.

Our \textbf{Graph Topology Augmentation} framework brings analytical Markov-chain tools into an RL planning module.
It builds on the Laplacian-potential intuition from electrical networks \citep{doyle1984random},
treating the policy-induced chain as the Laplacian kernel \citep{chung2005laplacians} and its
Green's function as the potential operator \citep{chung2000discrete}, with directed-chain eigenvalues \citep{levin2017markov} used as a scheduling signal. Analytical tools such as biorthogonal decomposition \citep{horn1985matrix}, additive reversibility decomposition \citep{bierkens2016non},
and classical spectral graph bounds \citep{chung1997spectral} further shape the framework.

\section{Preliminary}
\label{sec:preliminary}

\paragraph{Markov Decision Processes.}
\label{par:mdp}
We consider a finite discounted MDP
\[
  \calM = (\calS, \calA, p, r_t, \gamma),
\]
with $|\calS|<\infty$, $|\calA|<\infty$, fixed discount $\gamma \in (0,1)$, and
a \emph{fixed} transition kernel $p(s'\mid s,a)$. The reward is a bounded,
state-based vector $r_t : \calS \to \bbR$ that \textbf{may change over time}. Non-station is introduced in the modeling process as demand permutations captured by the reward signal, while the navigation skeleton is asserted to be static.

\paragraph{Priority-based planning.}
A serial planning sweep executes as a sequence of backups, and in-place updates let later backups immediately use earlier ones \citep[Ch.4]{sutton2018reinforcement}. Ordering the backups to maximize this overlap is the role of a priority queue, which always returns the entry with the largest weight. Section~\ref{subsec:gta-ps} describes the queue interface we adopt.

\paragraph{Reward setting.}
\label{par:reward-setting}
Let $\{\tau_m\}_{m\geq 0}$ denote the reward-shift times and let $N_{\mathrm{shift}}(n)$ be the
number of shifts up to step $n$. The reward is piecewise stationary on each interval
$[\tau_m, \tau_{m+1})$ and may shift exogenously at the times $\tau_m$ with maximum magnitude $\Delta_r$.

Sparse or localized reward changes are the main regime of interest. Let \(e_s \in \mathbb{R}^{|\mathcal{S}|}\) denote the standard basis vector associated with state \(s\), i.e.,
\[
  (e_s)_{s'} =
  \begin{cases}
    1, & s'=s,\\
    0, & s'\neq s.
  \end{cases}
\]
A rank-one reward shift at time \(t\) is a localized perturbation of the form $\delta \mathbf{r}_t = \Delta_{r,t}\, e_{s_{r,t}}$
where \(s_{r,t}\in\mathcal{S}\) is the affected state and \(\Delta_{r,t}\) is the shift amplitude.
A shift batch aggregates multiple localized reward shifts occurring at the same time step:
\[
  \delta\mathbf{r}_t
  =
  \sum_{i=1}^{m_t}
  \Delta_{r,t}^{(i)}\,e_{s_{r,t}}^{(i)},
  \quad
  m_t \leq m_{\max} \ll |\mathcal{S}|.
\]
This holds in almost all logistical reward design settings.

\paragraph{Agent setting.}
\label{par:agent-setting}
The host planner can vary: exact asynchronous value iteration or a Dyna-style planning loop. In both cases the planner maintains a value estimate, requiring efficient backup ordering. Each such backup is called a \emph{planning step}, planned before the host decides to take a committal action (denoted as a \emph{host step}). In our paper, the number of planning steps is fixed to $K$ per host step for fair comparisons. Our setting deliberately limits the use of prior knowledge, so that the agent is designed under minimal assumptions. The information available at initialization is either an oracle tabular model in the DP setting, or an immature learned model in the learning setting. Details of component settings in each host algorithm are discussed in \ref{app:b-host-intel}.

\paragraph{Residuals as inferred belief.}
Residual (in short for \emph{value estimation residual})  measures the discrepancy between
the planner's current estimate and new, implied one by latest evidence. Formally,
\[
    \xi_t(s)
    :=
    \left|
    \mathcal{T}_t V_t(s)-V_t(s)
    \right|,
\]
where $\mathcal{T}_t V_t(s)$ is the value estimate inferred from the
current reward and transition information.
Prioritized sweeping treats residual magnitude as a belief-correction score:
states with larger discrepancies are backed up first.
The important caution is that the inferred estimation should be both \emph{unbiased} and \emph{robust}
to variance (see \ref{app:proof-gtaps-infinite-visit}), which limits residual application using only one-step samples.

\paragraph{Paper scope.}
The theory assumes an exact model, a fixed transition kernel, and a tabular value function; reward nonstationarity is the only exogenous change covered by the guarantees. Dyna efficiency is verified empirically in Section~\ref{sec:experiments}, and the learned-model runs do not inherit the exact-model guarantees. Core notation is collected at the head of \ref{app:a}.

\section{Graph Topology Augmentation and Spectral Prioritized Sweeping}
\label{sec:gta-ps}

\subsection{Graph Topology Augmentation framework}
\paragraph{Policy-induced random walk.}
\label{par:policy-op}
For any policy $\mu(a\mid s)$, the induced transition matrix
\[
  P^\mu = \sum_a \mathrm{diag}(\mu(\cdot,a))\,P^a, \qquad P^\mu(s, s') = \sum_a \mu(a\mid s)\,p(s'\mid s,a),
\]
defines a random walk operator on the state graph. GTA treats $P^\mu$ as the soft adjacency matrix of the MDP that captures both the transitional rule and the navigational preference of the policy.

\paragraph{Directed topology operators.}
The row-stochastic matrix $P_t$ defines a directed state graph. Spectral Prioritized Sweeping (SPS / GTA-PS) uses two
regularized directed Laplacian systems:
\[
  \Lout_t = I-P_t,
  \qquad
  \Lin_t = D_t^{\mathrm{in}}-P_t^\top,
  \qquad
  D_t^{\mathrm{in}}=\operatorname{diag}(P_t^\top\mathbf{1}).
  \label{eq:gta-directed-laplacians}
\]
The out-potential measures how much residual a walk starting from $s$ expects
to encounter under the topology chain. The in-potential applies the analogous
construction on the edge-reversed flow. For a residual vector
$\xi_t$, define the diffused signal
\[
  \fout_t = (\Lout_t+\eta I)^{-1}|\xi_t|,
  \qquad
  \fin_t = (\Lin_t+\eta I)^{-1}|\xi_t|,
  \qquad
  \eta>0.
  \label{eq:gta-resolvent-potentials}
\]
where the regularizer $\eta I$ makes both systems invertible [\lemref{lemma:invertible-laplacian}].

The forward potential has a particularly grounded interpretation. At the canonical regularization
$
  \eta^\star=\frac{1-\gamma}{\gamma}.
$
Since $\Lout_t=I-P_t$,
$
  (\Lout_t+\eta^\star I)^{-1}
  =
  \gamma(I-\gamma P_t)^{-1}.
$
Therefore
\begin{align}
  & \fout_t
  =
  \gamma(I-\gamma P_t)^{-1}\xi_t
  =
  \gamma
  \sum_{k\ge 0}(\gamma P_t)^k\xi_t,
  \label{eq:gta-discounted-residual-propagation} \\
  \Rightarrow & \fout_t(s) =
  \gamma\,
  \bbE_{P_t}\!\left[
    \sum_{k\ge 0}\gamma^k\xi_t(X_k)
    \mid X_0=s
  \right]. \notag
\end{align}
\label{eq:gta-forward-expectation}
\noindent Thus the diagonal regularizer is not an arbitrary feature: $\eta$ effectively controls the geometric discount horizon. With $\eta = \eta^\star$, it equals exactly the reward discount rate $\gamma$. Other diffuser designs also carry effective-horizon readings, as discussed further in~\ref{par:why-resolvent}.

\paragraph{Second Largest Eigenvalue Modulus.}
\label{par:slem}
The eigenvalues of a stochastic matrix are ordered by modulus as
$1=\kappa_1 \geq |\kappa_2| \geq \cdots \geq |\kappa_n|$, and the Second
Largest Eigenvalue Modulus $\slem(P_t)=|\kappa_2|$ governs the asymptotic
contraction of the chain toward its stationary distribution \citep{levin2017markov}.
We read it as a \emph{global ventilation-confidence} statistic rather than a
local bottleneck detector: a value near one reports slow mixing, so an injected
residual spreads slowly and the diffused field deserves less confidence, while
a small value reports fast mixing and longer-range reliability.

\paragraph{GTA's rationale}
Where standard planning techniques transmit the reactive signal through mixing or representational kernels, GTA conditions the signal on topological operators, serving four roles: a proactive priority signal (\ref{par:topo-augment}), a detector of high-impact structure (\ref{app:e-ablations}), a gain scheduler (\ref{par:topo-scheduler}), and a diagnostic of the planning geometry (\ref{app:a-analytic-tools}).

\subsection{The GTA-PS instantiation.}
\label{subsec:gta-ps}

\paragraph{Topology policy.}
\label{par:topo-policy}
We design the \emph{topology policy} $\pi_t^{\mathrm{PS}}$, separate from but aligned with the host policy. Convergence tolerates any choice that stays reasonably aligned and preserves infinite-visit exploration [\lemref{lemma:infinite-visit-convergence}], so we use a finite-temperature softmax over centered action advantages. With the one-step action-value estimate
\[
  Q_t(s,a)=r_t(s) + \gamma \sum_{s'\in\calS} P_t^a(s,s')V_t(s')
\]
and the centered score $A_t(s,a) = Q_t(s,a) - \max_{b\in\calA} Q_t(s,b)$, we form the Boltzmann policy
\[
  \pi_t^{\mathrm{PS}}(a\mid s) = \frac{\exp(A_t(s,a)/\lambda)}{\sum_{a'}\exp(A_t(s,a')/\lambda)},
  \qquad
  \lambda >0
  \qquad
  P_t = \sum_a \mathrm{diag}(\pi_t^{\mathrm{PS}}(\cdot,a))\,P^a.
\]
The temperature keeps the ordering of action preferences while softening the policy, and finite $\lambda$ makes the logits scale-free; the induced chain then satisfies the infinite-visit requirement of \ref{prop:gtaps-infinite-visit}.

\paragraph{Augmented priority score.}
\label{par:topo-augment}
Standard PS keys the queue by the Bellman residual $p_t^{\mathrm{PS}}(s)=|\xi_t(s)|$. Spectral Prioritized Sweeping (SPS, or GTA-PS) keeps this base and adds the structure-aware spectral correction, with $\fout_t(s)$ quantifying the forward influence of the residual at $s$, and $\fin_t(s)$ the backward influence from predecessors through the edge-reversed graph.

The topology signal is:
\[
  \boldsymbol{\tau}_t \;\triangleq\;
  (1-\alpha)\,\left|\fout_t\right|
  \;+\;
  \alpha\,\left|\fin_t\right|,
  \qquad \alpha\in[0,1],
\]
where $\alpha \equiv \alpha_t$ is held fixed in deployed variants and small values keep the forward field dominant, matching the algorithm and the calibration in Appendix~\ref{app:d-additional-results}.

Spectral Prioritized Sweeping replaces the queue key with:
\[
  \mathrm{priority}_t(s) = |\xi_t(s)| + \beta_t\,\tau_t(s).
\]

\paragraph{Residual gate and top-$q$ frontier.}

Because the mixed field $\boldsymbol{\tau}_t$ built from the two resolvent solves is dense (see \ref{lemma:resolvent}), inserting it wholesale would crowd the queue with
low-priority states and add an excessive $O(|\mathcal{S}|)$ multiplier. The
module therefore separates signal availability from backup eligibility. The
residual gate
\[
  \mathcal{I}_t=\{s\in\mathcal{S}:|\xi_t(s)|>0\}
\]
is the executable candidate set, and the selection rule is
\[
  s_t\in\arg\max_{s\in\mathcal{I}_t}
  \left\{|\xi_t(s)|+\beta_t\,\tau_t(s)\right\}.
\]
Under the standing scheduler condition $\beta_1>0,\beta_2>0$ we have
$\beta_t\geq\beta_1\beta_2>0$ and $\tau_t\geq0$, so the topology term lifts
scores and never demotes a residual-active state below its Standard-PS key.
A bounded frontier $\mathcal{F}_q=\textsc{TopQ}(\tau_t,q)$ and its heap $\mathcal{H}_q$ 
store the dense field for rekeying; the deeper description resides in \ref{app:b-elitist-two-queue} and \ref{app:b-gating-mechanism}.
Setting $q_t=|\mathcal{S}|$ recovers full visibility for the field.
Positive $\tau_t$ outside the residual support proves nonlocal availability, not eligibility.

\paragraph{SLEM-based $\beta$ scheduler}
\label{par:topo-scheduler}
The forward/backward weight $\alpha$ is held fixed for a fair credit assignment (\ref{app:c-failed-approaches}); we vary only the topology gain.

In the SPS$^\beta$ (GTA-PS$^\beta$) variant, we use the SLEM-driven linear scheduler
$\beta_t = \beta_1(1 - \mathrm{SLEM}(P_t) + \beta_2)$. The gain amplifies the topology term in fast-mixing regimes and contracts it toward its positive floor $\beta_1\beta_2$ as the chain mixes more slowly (SLEM near one); the contraction reduces the topology bonus without demoting any residual-active state below its Standard-PS key, and setting $\beta_1=0$ recovers Standard-PS exactly.

In slowly mixing regimes the queue therefore leans back toward the reactive residual instead of over-trusting a diffused field. In sparse, low-volume regions a small top-$q$ frontier can retain substantial mass even at high SLEM, so the product $\beta_t\tau_t$, not $\beta_t$ alone, sets the queue contribution; checkpoint measurements of SLEM, $\beta_t$, and both priority masses in Appendix~\ref{app:d-post-review} support this joint reading. \ref{app:c} expands the design rationale.

Algorithm~1 in Appendix~\ref{app:b-algorithm-pseudocode} lists the complete refresh and planning procedures; all variants share it and differ only in the priority key.

\begin{figure}[t]
\centering
\resizebox{\linewidth}{!}{%
\begin{tikzpicture}[
  x=1cm, y=1cm,
  box/.style={draw, rounded corners=2pt, align=center, inner sep=4pt, font=\small},
  arr/.style={-{Stealth[length=2mm]}, thick}
]
\node[box, fill=blue!6] (xi) at (0,1.25) {$\xi_t$: Bellman residual};
\node[font=\scriptsize\itshape, above=1mm of xi] {reactive};
\node[box, fill=orange!10] (qxi) at (4.1,1.25) {$\mathcal{Q}_\xi$ over $\mathcal{I}_t$};
\node[box, fill=blue!6] (tau) at (0,-1.25) {$\tau_t$ from the Laplacian pair};
\node[font=\scriptsize\itshape, below=1mm of tau] {proactive};
\node[box, fill=orange!10] (hq) at (4.1,-1.25) {$\mathcal{H}_q=\textsc{TopQ}(\tau_t,q)$};
\node[box, fill=green!10] (pop) at (8.3,0) {\textsc{PopAndPrioritize}\\ residual-first over $\mathcal{I}_t$};
\node[box] (back) at (12.1,0) {Bellman backup $s^\star$};
\draw[arr] (xi) -- (qxi);
\draw[arr] (tau) -- (hq);
\draw[arr] (qxi.east) -- (pop.west);
\draw[arr] (pop) -- (back);
\draw[arr, dashed] (hq.north) -- node[right, font=\scriptsize, pos=0.4] {rekey, storage only} (qxi.south);
\draw[arr, dashed] (back.north) -- ++(0,1.6) -| node[pos=0.72, above, font=\scriptsize] {\textsc{PushPredecessors} refreshes keys} (qxi.north);
\end{tikzpicture}%
}
\caption{
\footnotesize
The two-queue organization in Spectral Prioritized Sweeping (SPS / GTA-PS). The reactive lane feeds the executable residual queue; the proactive lane maintains the bounded topology frontier. \textsc{PopAndPrioritize} scores only residual-active states; the frontier supplies the exploratory fallback once the residual queue is exhausted.}
\label{fig:two-queue}
\end{figure}

\section{Theoretical Analysis of Spectral Prioritized Sweeping}
\label{sec:theory}

The general identities and convergence statements below use the standing assumptions collected in Appendix~\ref{app:assumptions}.
Results about blocked states and cumulative tracking additionally require bounded, sparse, ergodic
reward shifts, as made explicit in the theorem statement and Appendix~\ref{app:assumptions}.
All full proofs are given in Appendix~\ref{app:proofs}.

\subsection{Convergence-guarantee of SPS on DP}
\label{subsec:convergence}
We first ensure that Spectral Prioritized Sweeping (SPS, or GTA-PS) is not worse than Standard-PS in the stationary limit: value iteration must still converge to the optimal policy, at least when reward shifts are sparse enough. The key observation is that the ordering rule only affects the transient. As long as every state is selected for backup infinitely often, the queue key cannot move the fixed point, and the contraction factor $\gamma<1$ delivers the same limit.

\begin{lemma}[Convergence under infinite state visits]
\label{lemma:infinite-visit-convergence}
Let $\mathcal{T}$ be the Bellman optimality operator, a $\gamma$-contraction in $\norm{\cdot}_\infty$. Suppose an asynchronous backup scheme selects a sequence of states $\{s_k\}_{k\geq 0}$ such that every state $s\in\mathcal{S}$ is selected infinitely often. Then $V_k\to V^*$ in $\norm{\cdot}_\infty$, independently of the backup-ordering rule.
\end{lemma}

We claim that SPS maintains lifelong exploration via infinite visit. 
\begin{proposition}[SPS satisfies infinite-visit condition]
\label{prop:gtaps-infinite-visit}
Under GTA-PS$^\beta$ with $\beta_2>0$, $\beta_1(1+\beta_2)<\eta^*$, and random tie-breaking among equal-priority states in the queue,
then every state $s\in\mathcal{S}$ is selected for backup infinitely often.
Consequently, GTA and GTA-PS$^\beta$ converge to $V^*$ in $\norm{\cdot}_\infty$.
\end{proposition}

\subsection{Dense support and the reactive-PS barrier}
\label{subsec:barrier}

\emph{Intuition.} Irreducibility gives every state a path to the shift site, so the Neumann series in Section~\ref{sec:gta-ps} assigns positive field mass everywhere immediately, while reactive PS must deliver that mass through one predecessor backup per hop. This is what can leave blocked states unqueued under a fixed budget.

\begin{proposition}[Dense support and PS barrier]
\label{prop:dense-support}
Assume a single-site reward shift at $s_r$, let $h^*(s;s_r)$ denote the backward predecessor distance (the length of the shortest directed path from $s$ to $s_r$ under the support of $P_t$), and let
\[
  \mathcal{B}_K
  \triangleq
  \{s\in\mathcal{S}: h^*(s;s_r)>K\}
\]
be the blocked set beyond the $K$-backup predecessor wavefront. Then: $(i)$ $\tau_t(s)>0$ for every state $s$; and $(ii)$ after at most $K$ reactive PS backups, every blocked state satisfies $p_t^{\mathrm{PS}}(s)=0$ while $\tau_t(s)>0$. This reveals the reactive barrier, alongside bottleneck hindrance effect formally.
\end{proposition}

An important note is that availability is not translated from the bare priority magnitude. How much useful lookahead survives the reactive barrier is instance-dependent, quantified by the horizon-significance account in Appendix~\ref{app:a-horizon}.

\subsection{Nonstationary Performance Analysis}
\label{subsec:nonstationary}

Finally, we state the cumulative, inter-episode guarantee for GTA-PS$^\beta$ over Standard-PS. We apply the assumption~\ref{app:a-assm-slow-drift} that the evolving topology kernel remains close to a frozen reference kernel $\bar P$ in the converging region:
\[
\|P_t-\bar P\|_\infty \le \varepsilon_{\mathrm{pol}}.
\]

Throughout this subsection, $\tau$ denotes the forward field at the canonical regularizer, $\tau(P,\zeta)=\gamma(I-\gamma P)^{-1}\zeta$; the deployed mixture adds the backward component with the small weight $\alpha$ calibrated in Appendix~\ref{app:d-additional-results}, outside the closed-form statements below.

\begin{lemma}[Lipschitz priority mass advantage]
\label{lemma:adv-lipschitz}
For a fixed single-site episode-start residual field $\zeta=\Delta_r e_{s_r}$, define the proactive reward shift vector:
\[
  \tau(P,\zeta)\triangleq \gamma(I-\gamma P)^{-1}\zeta .
\]
Let $\tau_{(q+1)}(P,\zeta)$ denote the $(q+1)$-st largest entry of $\tau(P,\zeta)$,
using the conservative convention that ties at the cutoff are excluded from the top-$q$ set. Define
\[
  \mathrm{adv}(P,\zeta)
  \triangleq
  \frac{1}{\gamma}\sum_{s\in\mathcal{B}_K(s_r)}
  \bigl(\tau(P,\zeta)(s)-\tau_{(q+1)}(P,\zeta)\bigr)_+
\]
as the excess priority mass beyond the $K$-backup predecessor wavefront, above the top-$q$ cutoff.

Then, $\mathrm{adv}(P,\zeta)$ is uniformly Lipschitz in $P$:
\[
  \abs{\mathrm{adv}(P,\zeta)-\mathrm{adv}(P',\zeta)}
  \leq
  L\norm{P-P'}_\infty,
  \qquad
  L\triangleq
  \frac{2|\mathcal{S}|\gamma\Delta_{\max}}{(1-\gamma)^2}.
\]
\end{lemma}

\emph{Intuition.} Smoothness of the advantage function, together with a controllable choice of $\bar P$ along the convergence trajectory, provides the positive guarantee in a local region.

\vspace{0.2cm}
\paragraph{Tracking advantage} The anchor, given a suitable choice formulated in~\ref{app:a-assm-positive-local-advantage}, provides a positive advantage over a policy point. Lipschitzness then extrapolates the gain over the local convergence region. Thus, we attain:

\begin{proposition}[Cumulative ergodic advantage]
\label{prop:cumulative-advantage}
Under the ergodic-reward, positive-local-advantage, and slow-drift assumptions~\ref{app:a-assm-ergodic-reward}, \ref{app:a-assm-positive-local-advantage}, and \ref{app:a-assm-slow-drift}
\[
  \frac{1}{N}\sum_{t=1}^N \mathrm{adv}_t(P_t,\zeta_t)
  \geq
  \bar{A}-L\varepsilon_{\mathrm{pol}}+o(1)
  \qquad\text{a.s.},
\]
for some $\bar A>0$.
\end{proposition}

\emph{Intuition.} Three effects combine across shifts: above the frontier cutoff, GTA-PS holds a positive mass advantage $\bar A$ per shift; policy drift erodes that advantage at a Lipschitz rate in the kernel perturbation; and the ergodic shift process averages the per-shift gains, so the cumulative separation grows linearly while the drift loss stays below the advantage.

\vspace{0.2cm}
\paragraph{Global advantage} Aggregating tracking across timesteps, diffusing over the successor representation, we realize the cumulative all-horizon advantage.

\begin{theorem}[Cumulative tracking advantage of GTA-PS over reactive PS]
\label{thm:cumulative-tracking}
Under the assumptions of Proposition~\ref{prop:cumulative-advantage} and the bounded-reward assumption~\ref{app:a-assm-bounded-reward}, with budget $q=K$.
Let $\mathcal{B}_K^{\mathrm{sel},t}\triangleq\{s\in\mathcal{B}_K(s_r):\tau_t(s)>\tau_{(q+1),t}\}$ be the blocked states above the top-$q$ cutoff, and assume $|\mathcal{B}_K^{\mathrm{sel},t}|\geq 1$ at positive reward-shift frequency $\mu_R$.
Define $\mathrm{err}^{\mathrm{GTA}}_t \triangleq \tau_{(q+1),t}/\gamma$, $\mathrm{TE}^{\mathrm{GTA}}(N)\triangleq\sum_{t=1}^N \mathrm{err}^{\mathrm{GTA}}_t$, and
\[
  \mathrm{TE}^{\mathrm{PS}}(N)\triangleq \sum_{t=1}^N \sum_{s^*\in\mathcal{B}_K^{\mathrm{sel},t}} \Phi_t(s^*),
\]
where $\Phi_t \triangleq (I-\gamma P_t)^{-1}\xi_t$.
Then the cumulative tracking error of GTA-PS is strictly smaller than reactive PS by at least
$N(\bar{A}-L\varepsilon_{\mathrm{pol}})+o(N)$ almost surely.
In particular, if
$L\varepsilon_{\mathrm{pol}}<\bar{A}$ then
$\mathrm{TE}^{\mathrm{PS}}(N)-\mathrm{TE}^{\mathrm{GTA}}(N)\to+\infty$ almost surely.
\end{theorem}

\noindent\textbf{Significance.}
\thmref{thm:cumulative-tracking} is the centerpiece of our analysis: it provides, to our knowledge, the first cumulative guarantee for topology-augmented planning, showing that GTA-PS's advantage over reactive PS grows without bound with the number of episodes whenever the drift bound satisfies $L\varepsilon_{\mathrm{pol}}<\bar{A}$.

\begin{figure}[t]
\centering
\includegraphics[width=0.85\linewidth]{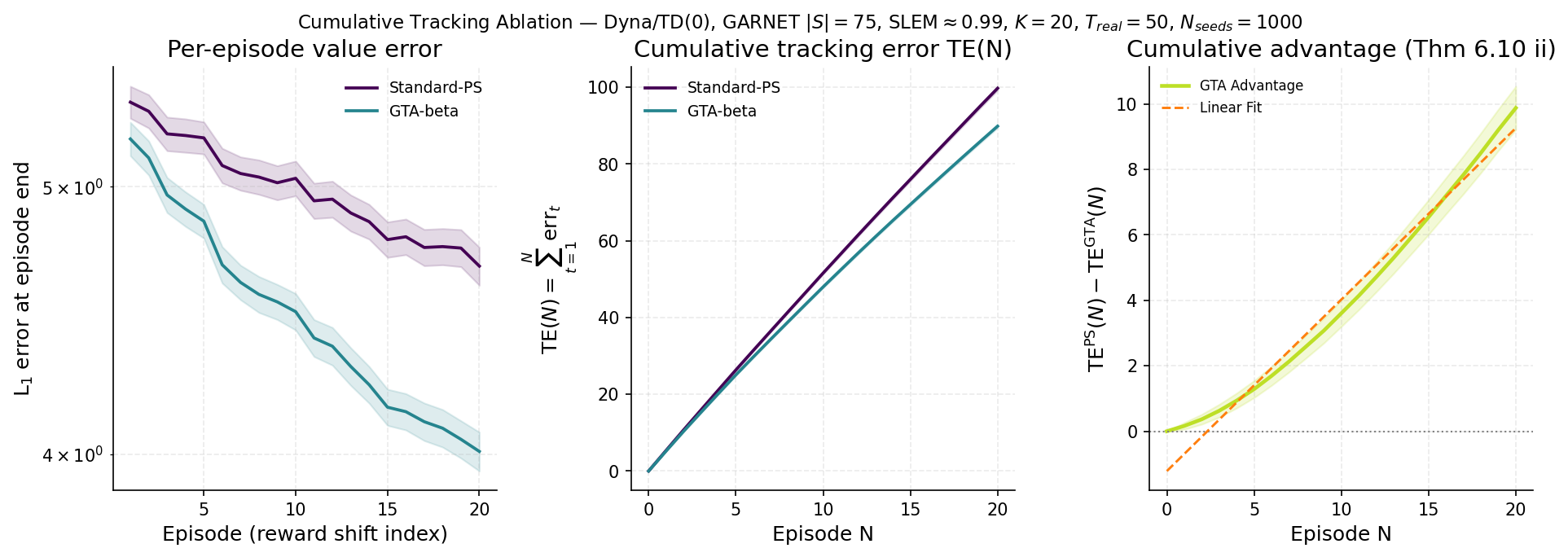}
\caption{\footnotesize Cumulative tracking ablation under repeated reward shifts in GARNET (Dyna/TD(0) host, $|\mathcal{S}|=75$, $K=20$, 1000 seeds). GTA-PS$^\beta$ lowers the per-episode terminal error, reduces cumulative tracking error $\mathrm{TE}(N)$, and produces a steadily increasing cumulative advantage over Standard-PS, matching the direction of the linear separation in \thmref{thm:cumulative-tracking}.}
\label{fig:cumulative-ablation}
\end{figure}

\subsection{Additional lens}
\label{subsec:lens}

This subsection collects the operator facts that motivate the design; the statements and proofs live in \ref{app:a}. It is interpretive: none of its content enters the proofs of Propositions~\ref{prop:gtaps-infinite-visit}, \ref{prop:dense-support}, and \ref{prop:cumulative-advantage} or of Theorem~\ref{thm:cumulative-tracking}.

\paragraph{Biorthogonal spectrum and the slow modes.} For a diagonalizable kernel with $P_t=\sum_k\kappa_kv_ku_k^\top$ and $u_j^\top v_k=\delta_{jk}$,
\[
  \fout_t=\sum_k\frac{u_k^\top|\xi_t|}{1-\kappa_k+\eta}\,v_k,
\]
since the resolvent acts spectrally as $(\Lout_t+\eta I)^{-1}v_k=(1-\kappa_k+\eta)^{-1}v_k$ (Lemma~\ref{lemma:green}). Slow modes near one receive weight up to $1/\eta$, so the field reaches farthest exactly when the chain mixes slowly; the first non-Perron mode is the component the SLEM schedule reads. Non-normality makes the reading delicate: the coefficients $u_k^\top|\xi_t|$ are left-eigenvector functionals, and transient amplification need not follow the modulus ordering, which is why the scheduler treats SLEM as a confidence signal rather than a guarantee.

\paragraph{The Doob tilt and its importance weights.} At the canonical regularizer, the predecessor entry of the forward Green's function is
\[
  \psi(s)\triangleq G^{out}_{\eta^\star,t}(s,s_r)
  =\gamma\bigl[(I-\gamma P_t)^{-1}\bigr]_{s,s_r}>0,
\]
and the tilted kernel $\tilde P_t(s,s')\triangleq\gamma P_t(s,s')\,\psi(s')/\psi(s)$ satisfies the row-sum identity $\sum_{s'}\tilde P_t(s,s')=1-\gamma\,\delta_{s,s_r}/\psi(s)$: it is stochastic on $\mathcal S\setminus\{s_r\}$ and absorbing at the shift site. Along any positive path, the change of measure with respect to $\gamma P_t$ telescopes,
\[
  \prod_{j=0}^{k-1}\frac{\tilde P_t(X_j,X_{j+1})}{\gamma P_t(X_j,X_{j+1})}
  =\prod_{j=0}^{k-1}\frac{\psi(X_{j+1})}{\psi(X_j)}
  =\frac{\psi(X_k)}{\psi(X_0)},
\]
so ranking by $\psi$ is an implicit importance-sampling rule toward the shift site (Remark~\ref{prop:importance-sampling}).

\section{Experiments}
\label{sec:experiments}

Under repeated reward shifts, we evaluate Spectral Prioritized Sweeping (SPS / GTA-PS) on tabular GARNET \citep{archibald1995generation} and Minigrid FourRooms \citep{chevalier2023minigrid}, with an oracle DP host and a Dyna host under a fixed budget $K$. Setup, reward protocol, and computational details live in Appendix~\ref{app:d-environment-setup}--\ref{app:d-computational-details}.

\begin{figure*}[t]
\centering
\includegraphics[width=0.85\linewidth]{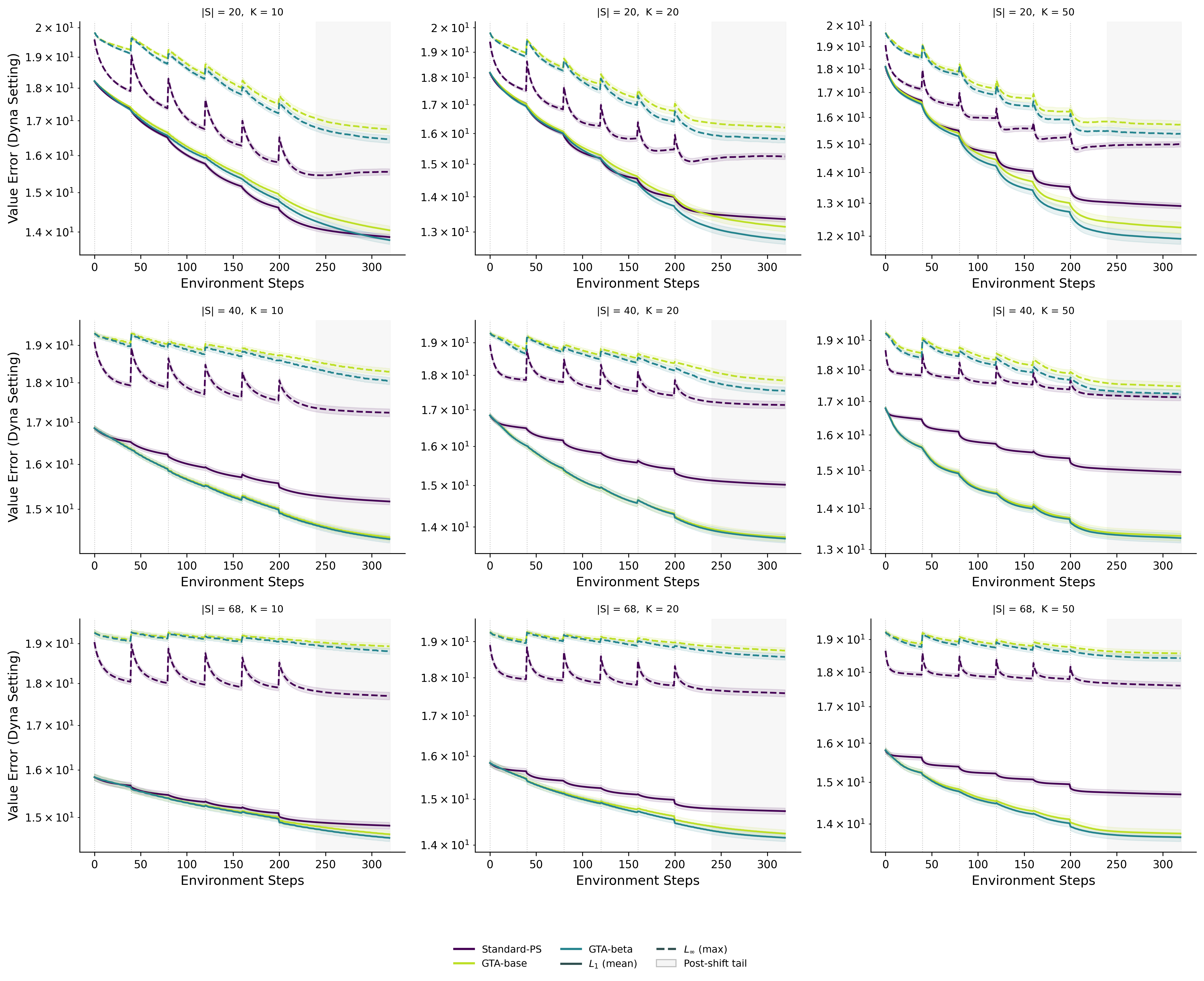}
\caption{\footnotesize Repeated reward-shift recovery under a Dyna host. Each panel reports value-error traces after multiple reward shifts for a fixed planning budget $K$ and state size $|\mathcal{S}|$. Solid lines denote mean $L_1$ error and dashed lines denote mean $L_\infty$ error. Across all tested sizes and budgets, both SPS variants recover faster than Standard-PS, with SPS$^\beta$ (GTA-PS$^\beta$) typically matching or slightly improving over the fixed-$\beta$ variant.}
\label{fig:rq1-dyna}
\end{figure*}

\subsection{Recovery and planning efficiency}
Under a fixed planning budget, reactive propagation reaches only states near the shifted reward, and SPS closes the gap by ordering backups with the spectral topology field. In Fig.~\ref{fig:rq1-dyna}, both SPS variants reduce value error faster than Standard-PS at every tested size and budget, and the gain concentrates right after each shift. Because the gap reappears at every shift, it accumulates into a persistent advantage (Fig.~\ref{fig:cumulative-ablation}), as \thmref{thm:cumulative-tracking} predicts. So the same budget yields faster recovery, more reward, and higher planning accuracy.

\begin{table}[H]
\centering
\small
\caption{\footnotesize Equal-wall-clock recovery on GARNET-200. Entries are means with 95\% Student-$t$ intervals over 100 paired seeds. AUC summarizes the error integral over each recovery window. End-point column records the final error, and backups/s reports achieved planning throughput.}
\label{tab:d-wall-clock}
\resizebox{\linewidth}{!}{%
\begin{tabular}{llccccc}
\toprule
Host & Method & $L_1$ AUC & $L_\infty$ AUC & End $L_1$ & End $L_\infty$ & Backups/s \\
\midrule
DP & Standard-PS & $0.923\pm0.033$ & $1.342\pm0.052$ & $0.024\pm0.002$ & $0.026\pm0.002$ & $1771\pm37$ \\
DP & SPS$^\beta$ (GTA-PS$^\beta$) & $1.840\pm0.063$ & $2.783\pm0.106$ & $0.345\pm0.018$ & $0.395\pm0.021$ & $785\pm13$ \\
Dyna & Standard-PS & $12.178\pm0.097$ & $19.313\pm0.155$ & $12.144\pm0.097$ & $19.096\pm0.158$ & $1647\pm40$ \\
Dyna & SPS$^\beta$ (GTA-PS$^\beta$) & $12.159\pm0.093$ & $19.411\pm0.151$ & $12.127\pm0.094$ & $19.209\pm0.154$ & $970\pm21$ \\
\bottomrule
\end{tabular}
}
\end{table}

Table~\ref{tab:d-wall-clock} tests the claim under a shared wall-clock budget instead of a shared backup count. Because refreshes and queue maintenance consume time, SPS buys fewer backups per second. Standard-PS ends lower on the exact-DP boundary, and the Dyna endpoints are close since both methods converge inside the window. So the audit supports expanded backup allocation, and it leaves a universal wall-clock speedup unestablished.

\subsection{Robustness and Recommendations}
Because the topology term enters as a bias, its weight decides how well it helps (Appendix~\ref{app:d-additional-results}). Moderate $\beta$ aligns the field with the realized pop order, while oversized $\beta$ sharply degrades that alignment. Since the term should surface blocked states early, the pattern matches \propref{prop:gtaps-infinite-visit}. The $\alpha$ sweeps point the same way, so we tune with small backward mixing and moderate $\beta$.

SPS stays effective in stochastic GARNET and bottlenecked FourRooms, because the topology field still surfaces blocked states there (Appendix~\ref{app:d-additional-results}). Under adversarial or drift-heavy transport, however, the chain stops matching the needed paths, and guidance weakens. So the topology signal works best as an adaptive bias, not a replacement for residual priority.

\section{Conclusion}

GTA builds representation operators from graph theory, so the framework captures patterns that degree counts or path-based metrics miss, including connectivity, influence, and community structure. On that spectral foundation, Spectral Prioritized Sweeping (SPS, or GTA-PS) maps beliefs in both the model and the policy into a structural priority score. Theory and experiments support the design, and both hosts show the module transfers. Future work can extend the framework to higher-dimensional settings, continuous control, and combine it with advances in representation learning, exploration, and scalable planning.

\subsection*{Limitations and Future work}
Past small-to-medium tabular domains, the scaling frontier of the mechanism remains unmeasured. The shifts are also localized and positive, so the setting stays narrow. Adversarial, drift-heavy, or spark-like transport can still distort the topology signal and weaken guidance. We validate DP and Dyna hosts only, leaving integrations such as Dyna-2 open. Because refreshes and queue maintenance cost real time, our gain is replanning quality under fixed budgets, not lower wall-clock cost.

Three extensions look most promising. In the fast-shift regime $\rho=T/t_{\mathrm{mix}}(1/4)\ll1$, we conjecture that the mean advantage grows as $1/\rho$, so a scaling argument would locate the largest benefit relative to cost. Because the analysis assumes an exact model, extending it to learned models would cover Dyna-style deployment under model noise, and the SLEM and resolvent potentials could inform exploration or model-learning budgets.

\bibliographystyle{plainnat}
\bibliography{references}

@book{bertsekas1989parallel,
  author    = {Bertsekas, Dimitri P. and Tsitsiklis, John N.},
  title     = {Parallel and Distributed Computation: Numerical Methods},
  publisher = {Prentice Hall},
  address   = {Englewood Cliffs, NJ},
  year      = {1989},
  isbn      = {978-0-13-648700-5}
}

@article{tsitsiklis1994asynchronous,
  author    = {Tsitsiklis, John N.},
  title     = {Asynchronous Stochastic Approximation and {Q}-Learning},
  journal   = {Machine Learning},
  volume    = {16},
  number    = {3},
  pages     = {185--202},
  year      = {1994},
  doi       = {10.1007/BF00993306}
}

@book{sutton2018reinforcement,
  author    = {Sutton, Richard S. and Barto, Andrew G.},
  title     = {Reinforcement Learning: An Introduction},
  edition   = {2nd},
  publisher = {MIT Press},
  year      = {2018},
  url       = {http://incompleteideas.net/book/the-book-2nd.html}
}

@article{plappert2018multigoal,
  author    = {Plappert, Matthias and Andrychowicz, Marcin and Ray, Alex and McGrew, Bob and Baker, Bowen and Powell, Glenn and Schneider, Jonas and Tobin, Josh and Chociej, Maciek and Welinder, Peter and Kumar, Vikash and Zaremba, Wojciech},
  title     = {Multi-Goal Reinforcement Learning: Challenging Robotics Environments and Request for Research},
  journal   = {arXiv preprint arXiv:1802.09464},
  year      = {2018},
  url       = {https://arxiv.org/abs/1802.09464},
}

@article{aviv2005pomdp,
  author    = {Aviv, Yossi and Pazgal, Amit},
  title     = {A Partially Observed {Markov} Decision Process for Dynamic Pricing},
  journal   = {Management Science},
  volume    = {51},
  number    = {9},
  pages     = {1400--1416},
  year      = {2005},
  doi       = {10.1287/mnsc.1050.0393},
}

@inproceedings{zou2019engagement,
  author    = {Zou, Lixin and Xia, Long and Ding, Zhuoye and Song, Jiaxing and Liu, Weidong and Yin, Dawei},
  title     = {Reinforcement Learning to Optimize Long-term User Engagement in Recommender Systems},
  booktitle = {Proceedings of the 25th {ACM} {SIGKDD} International Conference on Knowledge Discovery \& Data Mining},
  pages     = {2810--2818},
  year      = {2019},
  doi       = {10.1145/3292500.3330668},
  url       = {https://dl.acm.org/doi/10.1145/3292500.3330668},
}

@book{doyle1984random,
  author    = {Doyle, Peter G. and Snell, J. Laurie},
  title     = {Random Walks and Electric Networks},
  publisher = {Mathematical Association of America},
  series    = {Carus Mathematical Monographs},
  volume    = {22},
  year      = {1984},
  url       = {https://arxiv.org/abs/math/0001057},
}

@book{chung1997spectral,
  author    = {Chung, Fan R. K.},
  title     = {Spectral Graph Theory},
  series    = {{CBMS} Regional Conference Series in Mathematics},
  number    = {92},
  publisher = {American Mathematical Society},
  year      = {1997},
}

@article{chung2005laplacians,
  author    = {Chung, Fan},
  title     = {Laplacians and the {Cheeger} Inequality for Directed Graphs},
  journal   = {Annals of Combinatorics},
  volume    = {9},
  number    = {1},
  pages     = {1--19},
  year      = {2005},
  doi       = {10.1007/s00026-005-0237-z},
}

@book{levin2017markov,
  author    = {Levin, David A. and Peres, Yuval},
  title     = {Markov Chains and Mixing Times},
  edition   = {2nd},
  publisher = {American Mathematical Society},
  year      = {2017},
}

@book{woess2000random,
  author    = {Woess, Wolfgang},
  title     = {Random Walks on Infinite Graphs and Groups},
  series    = {Cambridge Tracts in Mathematics},
  volume    = {138},
  publisher = {Cambridge University Press},
  address   = {Cambridge, UK},
  year      = {2000},
  isbn      = {978-0-521-55292-9}
}

@article{bierkens2016non,
  title={Non-reversible metropolis-hastings},
  author={Bierkens, Joris},
  journal={Statistics and Computing},
  volume={26},
  number={6},
  pages={1213--1228},
  year={2016},
  publisher={Springer},
}

@article{chung2000discrete,
  author    = {Chung, Fan and Yau, Shing-Tung},
  title     = {Discrete {Green}'s functions},
  journal   = {Journal of Combinatorial Theory, Series A},
  volume    = {91},
  number    = {1-2},
  pages     = {191--214},
  year      = {2000},
  doi       = {10.1006/jcta.2000.3094},
}

@book{golub2013matrix,
  author    = {Golub, Gene H. and Van Loan, Charles F.},
  title     = {Matrix Computations},
  edition   = {4th},
  publisher = {Johns Hopkins University Press},
  year      = {2013},
}

@book{horn1985matrix,
  author    = {Horn, Roger A. and Johnson, Charles R.},
  title     = {Matrix Analysis},
  publisher = {Cambridge University Press},
  year      = {1985},
}

@article{saad1986gmres,
  author    = {Saad, Youcef and Schultz, Martin H.},
  title     = {{GMRES}: A Generalized Minimal Residual Algorithm for Solving
               Nonsymmetric Linear Systems},
  journal   = {{SIAM} Journal on Scientific and Statistical Computing},
  volume    = {7},
  number    = {3},
  pages     = {856--869},
  year      = {1986},
  doi       = {10.1137/0907058},
}

@article{meijerink1977iterative,
    author    = {Meijerink, J. A. and van der Vorst, H. A.},
    title     = {An Iterative Solution Method for Linear Systems of Which the Coefficient Matrix Is a Symmetric {$M$}-Matrix},
    journal   = {Mathematics of Computation},
    volume    = {31},
    number    = {137},
    pages     = {148--162},
    year      = {1977},
    doi       = {10.1090/S0025-5718-1977-0438681-4},
}

@article{moore1993prioritized,
  author    = {Moore, Andrew W. and Atkeson, Christopher G.},
  title     = {Prioritized Sweeping: Reinforcement Learning with Less Data and Less Time},
  journal   = {Machine Learning},
  volume    = {13},
  number    = {1},
  pages     = {103--130},
  year      = {1993},
  publisher = {Springer},
  doi       = {10.1007/BF00993104}
}

@article{peng1993efficient,
  author    = {Peng, Jing and Williams, Ronald J.},
  title     = {Efficient Learning and Planning Within the {Dyna} Framework},
  journal   = {Adaptive Behavior},
  volume    = {1},
  number    = {4},
  pages     = {437--454},
  year      = {1993},
  publisher = {SAGE Publications},
  doi       = {10.1177/105971239300100403}
}

@article{wingate2005prioritization,
  author    = {Wingate, David and Seppi, Kevin D.},
  title     = {Prioritization Methods for Accelerating {MDP} Solvers},
  journal   = {Journal of Machine Learning Research},
  volume    = {6},
  pages     = {851--881},
  year      = {2005},
  url       = {https://jmlr.org/papers/v6/wingate05a.html},
}

@inproceedings{sutton1990dyna,
  author    = {Sutton, Richard S.},
  title     = {Integrated Architectures for Learning, Planning, and Reacting Based on Approximating Dynamic Programming},
  booktitle = {Proceedings of the Seventh International Conference on Machine Learning (ICML)},
  pages     = {216--224},
  year      = {1990},
  publisher = {Morgan Kaufmann}
}

@inproceedings{silver2008dyna2,
  author    = {Silver, David and Sutton, Richard S. and M\"{u}ller, Martin},
  title     = {Sample-Based Learning and Search with Permanent and Transient Memories},
  booktitle = {Proceedings of the 25th International Conference on Machine Learning ({ICML})},
  pages     = {968--975},
  year      = {2008},
  publisher = {{ACM}},
  doi       = {10.1145/1390156.1390280},
}

@article{dai2011topological,
  author    = {Dai, Peng and Mausam and Weld, Daniel S. and Goldsmith, Judy},
  title     = {Topological Value Iteration Algorithms},
  journal   = {Journal of Artificial Intelligence Research},
  volume    = {42},
  pages     = {181--209},
  year      = {2011},
  doi       = {10.1613/jair.3390},
  url       = {https://arxiv.org/abs/1401.3910},
}

@inproceedings{corneil2018vast,
  author    = {Corneil, Dane and Gerstner, Wulfram and Brea, Johanni},
  title     = {Efficient Model-Based Deep Reinforcement Learning with Variational State Tabulation},
  booktitle = {Proceedings of the 35th International Conference on Machine Learning ({ICML})},
  pages     = {1049--1058},
  year      = {2018},
  publisher = {{PMLR}},
  url       = {https://arxiv.org/abs/1802.04325}
}

@article{dayan1993improving,
  author    = {Dayan, Peter},
  title     = {Improving Generalization for Temporal Difference Learning: The Successor Representation},
  journal   = {Neural Computation},
  volume    = {5},
  number    = {4},
  pages     = {613--624},
  year      = {1993},
  doi       = {10.1162/neco.1993.5.4.613}
}

@inproceedings{mahadevan2005proto,
  author    = {Mahadevan, Sridhar},
  title     = {Proto-Value Functions: Developmental Reinforcement Learning},
  booktitle = {Proceedings of the 22nd International Conference on Machine Learning ({ICML})},
  pages     = {553--560},
  year      = {2005},
  doi       = {10.1145/1102351.1102421},
}

@article{mahadevan2007proto,
  author    = {Mahadevan, Sridhar and Maggioni, Mauro},
  title     = {Proto-Value Functions: {A} {Laplacian} Framework for Learning Representation and Control in {Markov} Decision Processes},
  journal   = {Journal of Machine Learning Research},
  volume    = {8},
  pages     = {2169--2231},
  year      = {2007}
}

@article{shuman2013emerging,
  author    = {Shuman, David I. and Narang, Sunil K. and Frossard, Pascal and Ortega, Antonio and Vandergheynst, Pierre},
  title     = {The Emerging Field of Signal Processing on Graphs},
  journal   = {{IEEE} Signal Processing Magazine},
  volume    = {30},
  number    = {3},
  pages     = {83--98},
  year      = {2013},
  doi       = {10.1109/MSP.2012.2235192},
}

@article{belkin2003laplacian,
  author    = {Belkin, Mikhail and Niyogi, Partha},
  title     = {Laplacian Eigenmaps for Dimensionality Reduction and Data Representation},
  journal   = {Neural Computation},
  volume    = {15},
  number    = {6},
  pages     = {1373--1396},
  year      = {2003},
  doi       = {10.1162/089976603321780317},
}

@article{stachenfeld2017hippocampus,
  author    = {Stachenfeld, Kimberly L. and Botvinick, Matthew M. and Gershman, Samuel J.},
  title     = {The Hippocampus as a Predictive Map},
  journal   = {Nature Neuroscience},
  volume    = {20},
  number    = {11},
  pages     = {1643--1653},
  year      = {2017},
  publisher = {Nature Publishing Group},
  doi       = {10.1038/nn.4650}
}

@inproceedings{machado2018eigenoption,
  author    = {Machado, Marlos C. and Rosenbaum, Clemens and Guo, Xiaoxiao and Liu, Miao and Tesauro, Gerald and Campbell, Murray},
  title     = {Eigenoption Discovery Through the Deep Successor Representation},
  booktitle = {International Conference on Learning Representations (ICLR)},
  year      = {2018},
  url       = {https://arxiv.org/abs/1710.11089}
}

@inproceedings{jinnai2019discovering,
  author    = {Jinnai, Yuu and Park, Jee Won and Abel, David and Konidaris, George},
  title     = {Discovering Options for Exploration by Minimizing Cover Time},
  booktitle = {Proceedings of the 36th International Conference on Machine Learning ({ICML})},
  pages     = {3130--3139},
  year      = {2019},
  publisher = {PMLR}
}

@inproceedings{machado2017laplacian,
  author    = {Machado, Marlos C. and Bellemare, Marc G. and Bowling, Michael},
  title     = {A {Laplacian} Framework for Option Discovery in Reinforcement Learning},
  booktitle = {Proceedings of the 34th International Conference on Machine Learning ({ICML})},
  pages     = {2295--2304},
  year      = {2017},
  publisher = {{PMLR}},
  url       = {https://proceedings.mlr.press/v70/machado17a.html},
}

@inproceedings{wang2021laplacian,
  author    = {Wang, Kaixin and Zhou, Kuangqi and Zhang, Qixin and Shao, Jie and Hooi, Bryan and Feng, Jiashi},
  title     = {Towards Better {Laplacian} Representation in Reinforcement Learning with Generalized Graph Drawing},
  booktitle = {Proceedings of the 38th International Conference on Machine Learning ({ICML})},
  pages     = {11003--11012},
  year      = {2021},
  publisher = {{PMLR}},
  url       = {https://proceedings.mlr.press/v139/wang21ae.html},
}

@inproceedings{gomez2024proper,
  author    = {Gomez, Diego and Bowling, Michael and Machado, Marlos C.},
  title     = {Proper {Laplacian} Representation Learning},
  booktitle = {International Conference on Learning Representations ({ICLR})},
  year      = {2024},
  url       = {https://openreview.net/forum?id=7gLfQT52Nn},
}

@inproceedings{chauhan2026effective,
  author    = {Chauhan, Jatin and Bhardwaj, Shivam and Saibewar, Aditya and
               Ramesh, Aditya and Babar, Sadbhavana and Kaul, Manohar},
  title     = {Graph-Theoretic Intrinsic Reward: Guiding {RL} with Effective Resistance},
  booktitle = {International Conference on Learning Representations ({ICLR})},
  year      = {2026},
  url       = {https://openreview.net/forum?id=W8bKDPf1Ko},
}

@inproceedings{cheung2020nonstationary,
  author    = {Cheung, Wang Chi and Simchi-Levi, David and Zhu, Ruihao},
  title     = {Reinforcement Learning for Non-Stationary {Markov} Decision Processes: The Blessing of (More) Optimism},
  booktitle = {Proceedings of the 37th International Conference on Machine Learning ({ICML})},
  pages     = {1843--1854},
  year      = {2020},
  publisher = {{PMLR}},
  url       = {https://arxiv.org/abs/1906.02922}
}

@article{khetarpal2022continual,
  author    = {Khetarpal, Khimya and Riemer, Matthew and Rish, Irina and Precup, Doina},
  title     = {Towards Continual Reinforcement Learning: {A} Review and Perspectives},
  journal   = {Journal of Artificial Intelligence Research},
  volume    = {75},
  pages     = {1401--1476},
  year      = {2022},
  doi       = {10.1613/jair.1.13673}
}

@article{padakandla2020reinforcement,
  author    = {Padakandla, Sindhu and J, Prabuchandran K. and Bhatnagar, Shalabh},
  title     = {Reinforcement Learning Algorithm for Non-Stationary Environments},
  journal   = {Applied Intelligence},
  volume    = {50},
  pages     = {3590--3606},
  year      = {2020},
  publisher = {Springer},
  doi       = {10.1007/s10489-020-01758-5}
}

@inproceedings{janner2019trust,
  author    = {Janner, Michael and Fu, Justin and Zhang, Marvin and Levine, Sergey},
  title     = {When to Trust Your Model: Model-Based Policy Optimization},
  booktitle = {Advances in Neural Information Processing Systems ({NeurIPS})},
  volume    = {32},
  year      = {2019},
  url       = {https://arxiv.org/abs/1906.08253}
}

@article{archibald1995generation,
  author    = {Archibald, T. W. and McKinnon, K. I. M. and Thomas, L. C.},
  title     = {On the Generation of {Markov} Decision Processes},
  journal   = {Journal of the Operational Research Society},
  volume    = {46},
  number    = {3},
  pages     = {354--361},
  year      = {1995},
  doi       = {10.1057/jors.1995.55},
}

@inproceedings{chevalier2023minigrid,
  author    = {Chevalier-Boisvert, Maxime and Dai, Bolun and Towers, Mark and
               {Perez-Vicente}, Rodrigo and Willems, Lucas and Lahlou, Salem and
               Pal, Suman and Castro, Pablo Samuel and Precup, Doina},
  title     = {{Minigrid} \& {Miniworld}: Modular \& Customizable Reinforcement Learning
               Environments for Goal-Oriented Tasks},
  booktitle = {Advances in Neural Information Processing Systems ({NeurIPS})},
  volume    = {36},
  year      = {2023},
  url       = {https://arxiv.org/abs/2306.13831},
}

@inproceedings{akiba2019optuna,
  author    = {Akiba, Takuya and Sano, Shotaro and Yanase, Toshihiko and Ohta, Takeru and Koyama, Masanori},
  title     = {Optuna: A Next-generation Hyperparameter Optimization Framework},
  booktitle = {Proceedings of the 25th {ACM} {SIGKDD} International Conference on Knowledge Discovery \& Data Mining},
  pages     = {2623--2631},
  year      = {2019},
  doi       = {10.1145/3292500.3330701},
  url       = {https://doi.org/10.1145/3292500.3330701},
}

\newpage
\newcommand{\appendixheading}[1]{%
  \refstepcounter{section}%
  \setcounter{subsection}{0}%
  \phantomsection%
  \section*{%
    {\fontsize{16pt}{18pt}\selectfont Appendix \thesection. #1}%
  }%
  \addcontentsline{toc}{section}{Appendix \thesection. #1}%
}

\newcommand{\appendixsubsection}[1]{%
  \refstepcounter{subsection}%
  \subsection*{%
    {\fontsize{13pt}{15pt}\selectfont \thesubsection\ #1}%
  }%
}

\newcommand{\appendixsubsubsection}[1]{%
  \refstepcounter{subsubsection}%
  \subsubsection*{%
    {\fontsize{11pt}{13pt}\selectfont \thesubsubsection\ #1}%
  }%
}

\appendix

\begin{center}
{\bf \Large{Appendix of ``Spectral Prioritized Sweeping in Non-stationary Reinforcement Learning''}}
\end{center}
\DoToC
\medskip

\makeatletter
\renewcommand{\p@section}{Appendix~}
\makeatother

\appendixheading{Proofs of Theoretical Results}
\label{app:a}

\appendixsubsection{Core notation}
\label{app:a-notation}

\begin{table}[h]
\centering\small
\caption{\footnotesize Core notation used throughout the paper and the appendix.}
\label{tab:notation}
\begin{tabular}{p{0.30\linewidth}p{0.63\linewidth}}
\toprule
Symbol & Meaning \\
\midrule
$\calM=(\calS,\calA,p,r_t,\gamma)$ & Finite discounted tabular MDP with stable transition skeleton and piecewise-stationary reward \\
$\pi,\,\pi^{\mathrm{PS}}$ & Host policy and topology policy, respectively \\
$P_t^{\mathrm{PS}} \; (\text{e.g., } P_t)$ & Topology-policy-induced transition matrix \\
$\Lout_t = I - P_t$ & Out-Laplacian; governs forward diffusion \\
$\Lin_t = D_t^{\mathrm{in}} - (P_t)^\top$ & In-Laplacian; governs backward attribution \\
$\fout_t,\,\fin_t$ & Forward and backward residual potentials \\
$\slem(P_t) \; (\text{e.g., } \slem)$ & Second largest eigenvalue modulus of the topology chain \\
$K, T_\mathrm{shift}$ & Fixed planning budget per host step and host step budget per reward shift respectively \\
\bottomrule
\end{tabular}
\end{table}

\appendixsubsection{Analytical assumptions}
\label{app:assumptions}
First we address some stronger behaviorial assumptions or workarounds applied to ensure robustness throughout our analytics.

\begin{assumption}[Communicating MDP]\label{app:a-assm-communicating}
The MDP $\mathcal{M}=(\mathcal{S},\mathcal{A},p,\mathcal{R},\gamma)$ has finite
state and action spaces and is \emph{communicating}: for any two states
$s,s'\in\mathcal{S}$, there exists a stationary policy under which $s'$ is
reachable from $s$ with positive probability. Equivalently, every
$\varepsilon$-soft stationary policy induces an irreducible transition matrix
on $\mathcal{S}$. Since the Boltzmann policy design in \ref{par:topo-policy} assigns nonzero probability to every action,
the induced transition matrix $P_t$ is \textbf{irreducible} for all $t$.
\end{assumption}

\begin{assumption}[Bounded reward signal]\label{app:a-assm-bounded-reward}
Rewards may be stochastic and may change exogenously over time, but their absolute value do not exceed $\Delta_{\max}$.
This keeps residuals, value targets, and topology scores finite.
\end{assumption}

\begin{assumption}[Ergodic reward shifts]\label{app:a-assm-ergodic-reward}
The sequence $\{(s_{r,t},\,\Delta_{r,t})\}_{t\geq 1}$ is \textbf{stationary-ergodic} under a measure $\mu_R$, with mean shift amplitude $\bar{\Delta}_r \triangleq \mathbb{E}_{\mu_R}[\Delta_{r,t}] > 0$ and cumulative total variation $B_N \triangleq \sum_{t=1}^N \Delta_{r,t} = N\bar{\Delta}_r(1+o(1))$ a.s.
\end{assumption}

\begin{assumption}[Slow topology drift for cumulative tracking]\label{app:a-assm-slow-drift}
  For the cumulative-tracking analysis, there exist a reference kernel $\bar P$
  and a constant $\varepsilon_{\mathrm{pol}}\ge 0$ such that throughout the
  analysis window,
  \[
    \|P_t-\bar P\|_\infty \le \varepsilon_{\mathrm{pol}}.
  \]
  This is the frozen-kernel / slow-drift envelope used to compare the evolving
  process against a fixed-kernel ergodic reference.
\end{assumption}

\begin{assumption}[Frozen-kernel positive local advantage]
\label{app:a-assm-positive-local-advantage}
Under the frozen reference kernel $\bar P$ from
Assumption~\ref{app:a-assm-slow-drift}, the reward-shift process has
positive probability of placing strictly above-cutoff topology mass inside
the $K$-neighborhood of the shift site:
\[
  \mu_R\!\left(
    \exists s\in\mathcal B_K(s_r):
    \tau(\bar P,\zeta)(s)>\tau_{(q+1)}(\bar P,\zeta)
  \right)>0.
\]
\end{assumption}

\begin{corollary}[Well-definedness of the kernel $P_t$]
\label{cor:kernel-well-defined}
The existence of a unique stationary distribution $\mu^{\mathrm{PS}}_t$ requires the
policy-induced kernel $P_t$ to be irreducible and aperiodic. When this condition is not
met exactly, it can be enforced by adding an auxiliary bridge edge with arbitrarily small
probability $\varepsilon>0$. Likewise, the Green's-function expansion
in~\lemref{lemma:green} is simplest when $P_t$ is diagonalizable, which can be enforced by
an arbitrarily small generic perturbation of one transition entry.
\end{corollary}

\appendixsubsection{Supporting Lemmas}
\label{app:a-analytic-tools}

\begin{theorem}[Levy--Desplanques Strict Diagonal Dominance Invertibility]\label{thm:levy-desplanques}
Let $A \in \bbR^{n \times n}$. If $A$ is strictly row diagonally dominant (SDD), i.e.,
\begin{equation}\label{eq:sdd-condition}
  \abs{a_{ii}} > \sum_{j \ne i} \abs{a_{ij}} \quad \text{for all } i = 1, 2, \dots, n,
\end{equation}
then $A$ is non-singular (invertible).
\end{theorem}
\begin{proof}
We prove this theorem by contradiction. Suppose $A$ is singular. Then $\kappa = 0$ is an eigenvalue of $A$, which implies there exists a non-zero eigenvector $\mathbf{v} \in \bbR^n \setminus \{\mathbf{0}\}$ such that $A \mathbf{v} = \mathbf{0}$.

Let $k \triangleq \mathrm{argmax}_{1 \le i \le n} \abs{v_i}$. Since $\mathbf{v} \ne \mathbf{0}$, we have $\abs{v_k} > 0$ and $\abs{v_k} \ge \abs{v_j}$ for all $j = 1, \dots, n$.
Evaluating the $k$-th equation of $A \mathbf{v} = \mathbf{0}$:
\[
  \sum_{j=1}^n a_{kj} v_j = 0 \implies -a_{kk} v_k = \sum_{j \ne k} a_{kj} v_j.
\]
Taking absolute values on both sides and applying the triangle inequality:
\[
  \abs{v_k} \abs{a_{kk}} = \abs{ \sum_{j \ne k} a_{kj} v_j } \le \sum_{j \ne k} \abs{a_{kj}} \abs{v_j} \le \abs{v_k} \sum_{j \ne k} \abs{a_{kj}}.
\]
Dividing both sides by $\abs{v_k} > 0$ yields $\abs{a_{kk}} \le \sum_{j \ne k} \abs{a_{kj}}$, which directly contradicts the strict row diagonal dominance condition~\eqref{eq:sdd-condition}. Therefore, $A$ must be invertible.
\end{proof}

\begin{lemma}[Invertible regularized Laplacian]
\label{lemma:invertible-laplacian}
For any time step $t$ and any $\eta > 0$, both shifted directed Laplacians $M_t^{\mathrm{out}} = \Lout_t + \eta I$ and $M_t^{\mathrm{in}} = \Lin_t + \eta I$ are strictly row diagonally dominant with a uniform diagonal gap of $\eta > 0$, and are therefore invertible.
\end{lemma}
\begin{proof}
We verify the strict row diagonal dominance for both systems:

\textbf{Case 1: Out-degree system $M_t^{\mathrm{out}} = \Lout_t + \eta I$.}
Recall $\Lout_t = I - P_t$, so $M_t^{\mathrm{out}} = (1 + \eta) I - P_t$. The diagonal and off-diagonal entries are:
\[
  a_{ii} = 1 + \eta - P_t(s_i, s_i), \qquad a_{ij} = -P_t(s_i, s_j) \le 0 \quad (j \ne i).
\]
Since $P_t$ is row-stochastic ($\sum_{j=1}^n P_t(s_i, s_j) = 1$), the diagonal entry satisfies:
\[
  a_{ii} = 1 + \eta - P_t(s_i, s_i) = \eta + \sum_{j \ne i} P_t(s_i, s_j) = \eta + \sum_{j \ne i} \abs{a_{ij}}.
\]
Thus the uniform row diagonal gap is $\abs{a_{ii}} - \sum_{j \ne i} \abs{a_{ij}} = \eta > 0$. By Theorem~\ref{thm:levy-desplanques}, $M_t^{\mathrm{out}}$ is invertible.

\textbf{Case 2: In-degree system $M_t^{\mathrm{in}} = \Lin_t + \eta I$.}
Recall $\Lin_t = D_t^{\mathrm{in}} - P_t^\top$, where $D_t^{\mathrm{in}} = \mathrm{diag}(c_{1,t}, \dots, c_{n,t})$ with $c_{i,t} = \sum_{j=1}^n P_t(s_j, s_i)$. The entries of $M_t^{\mathrm{in}} = D_t^{\mathrm{in}} + \eta I - P_t^\top$ are:
\[
  b_{ii} = c_{i,t} + \eta - P_t(s_i, s_i) = \eta + \sum_{j \ne i} P_t(s_j, s_i), \qquad b_{ij} = -P_t(s_j, s_i) \le 0 \quad (j \ne i).
\]
The sum of off-diagonal magnitudes in row $i$ is $\sum_{j \ne i} \abs{b_{ij}} = \sum_{j \ne i} P_t(s_j, s_i)$. Thus the diagonal gap is:
\[
  \abs{b_{ii}} - \sum_{j \ne i} \abs{b_{ij}} = \left( \eta + \sum_{j \ne i} P_t(s_j, s_i) \right) - \sum_{j \ne i} P_t(s_j, s_i) = \eta > 0.
\]
By Theorem~\ref{thm:levy-desplanques}, $M_t^{\mathrm{in}}$ is also invertible.
\end{proof}

\begin{lemma}[Resolvent norm bounds]
\label{lemma:resolvent}
Let $\eta > 0$. The $\ell_\infty$-operator norms of both regularized resolvent operators satisfy:
\[
  \norm{(\Lout_t+\eta I)^{-1}}_\infty = \frac{1}{\eta} \quad \text{and} \quad \norm{(\Lin_t+\eta I)^{-1}}_\infty = \frac{1}{\eta}.
\]
\end{lemma}
\begin{proof}
Let $A_t \in \{ \Lout_t + \eta I, \, \Lin_t + \eta I \}$. We carry out the derivation in three steps:

\textbf{Non-singularity and entrywise non-negativity ($A_t^{-1} \ge 0$).}
We split $A_t = D - B$, where $D = \mathrm{diag}((A_t)_{11}, \dots, (A_t)_{nn})$ is the diagonal matrix and $B$ contains the negated off-diagonal entries:
\[
  B_{ij} = \begin{cases} 0, & i = j, \\ -(A_t)_{ij}, & i \ne j. \end{cases}
\]
Since the off-diagonal entries of $A_t$ are non-positive ($(A_t)_{ij} \le 0$ for $i \ne j$), we have $B \ge 0$.
Moreover, $(A_t)_{ii} > 0$, so $D$ is invertible and $D^{-1} = \mathrm{diag}\left(\frac{1}{(A_t)_{11}}, \dots, \frac{1}{(A_t)_{nn}}\right) \ge 0$.
Factoring $A_t$:
\[
  A_t = D (I - D^{-1} B).
\]
Let $T \triangleq D^{-1} B \ge 0$. Its entries are $T_{ii} = 0$ and $T_{ij} = \frac{-(A_t)_{ij}}{(A_t)_{ii}} \ge 0$ for $i \ne j$.
Evaluating the induced matrix $\ell_\infty$-norm of $T$:
\[
  \norm{T}_\infty = \norm{D^{-1} B}_\infty = \max_{1 \le i \le n} \sum_{j \ne i} \frac{\abs{(A_t)_{ij}}}{(A_t)_{ii}}.
\]
By the strict row diagonal dominance established in Lemma~\ref{lemma:invertible-laplacian}, $(A_t)_{ii} - \sum_{j \ne i} \abs{(A_t)_{ij}} = \eta > 0$, which gives $\sum_{j \ne i} \abs{(A_t)_{ij}} < (A_t)_{ii}$. Thus:
\[
  \norm{D^{-1} B}_\infty = \max_{1 \le i \le n} \frac{\sum_{j \ne i} \abs{(A_t)_{ij}}}{(A_t)_{ii}} < 1.
\]
Since $\norm{T}_\infty < 1$, the Neumann series for $(I - T)^{-1}$ converges and all its terms are entrywise non-negative ($T \ge 0$):
\[
  (I - D^{-1} B)^{-1} = \sum_{m=0}^\infty (D^{-1} B)^m \ge 0.
\]
Consequently, the inverse of $A_t$ exists and is entrywise non-negative:
\begin{equation}\label{eq:app-inv-nonneg}
  A_t^{-1} = (I - D^{-1} B)^{-1} D^{-1} = \sum_{m=0}^\infty (D^{-1} B)^m D^{-1} \ge 0.
\end{equation}

\textbf{Annihilation of the constant all-ones vector.}
Evaluating the action of the unshifted directed Laplacians on the all-ones vector $\mathbf{1} = (1, \dots, 1)^\top$:
\begin{align}
  \Lout_t \mathbf{1} &= (I - P_t) \mathbf{1} = \mathbf{1} - P_t \mathbf{1} = \mathbf{1} - \mathbf{1} = \mathbf{0}, \\
  \Lin_t \mathbf{1} &= (D_t^{\mathrm{in}} - P_t^\top) \mathbf{1} = D_t^{\mathrm{in}} \mathbf{1} - P_t^\top \mathbf{1} = P_t^\top \mathbf{1} - P_t^\top \mathbf{1} = \mathbf{0},
\end{align}
since $D_t^{\mathrm{in}} \mathbf{1} = \mathrm{diag}(P_t^\top \mathbf{1}) \mathbf{1} = P_t^\top \mathbf{1}$ by definition.

\textbf{Exact row-sum evaluation and $\ell_\infty$-norm computation.}
Adding the regularizer $\eta I$:
\[
  A_t \mathbf{1} = (L_t + \eta I) \mathbf{1} = L_t \mathbf{1} + \eta \mathbf{1} = \mathbf{0} + \eta \mathbf{1} = \eta \mathbf{1}.
\]
Multiplying both sides from the left by $A_t^{-1}$:
\[
  A_t^{-1} (A_t \mathbf{1}) = A_t^{-1} (\eta \mathbf{1}) \implies \mathbf{1} = \eta A_t^{-1} \mathbf{1} \implies A_t^{-1} \mathbf{1} = \frac{1}{\eta} \mathbf{1}.
\]
Thus, every row $i$ of $A_t^{-1}$ has exact row sum $\sum_{j=1}^n (A_t^{-1})_{ij} = \frac{1}{\eta}$. Because $A_t^{-1} \ge 0$ by~\eqref{eq:app-inv-nonneg}, all entries equal their absolute values. By definition of the induced matrix $\ell_\infty$-norm:
\[
  \norm{A_t^{-1}}_\infty = \max_{1 \le i \le n} \sum_{j=1}^n \abs{(A_t^{-1})_{ij}} = \max_{1 \le i \le n} \sum_{j=1}^n (A_t^{-1})_{ij} = \norm{A_t^{-1}\mathbf{1}}_\infty = \norm{\frac{1}{\eta}\mathbf{1}}_\infty = \frac{1}{\eta}.
\]
Applying this identity to $A_t = \Lout_t + \eta I$ and $A_t = \Lin_t + \eta I$ yields the exact equality.
\end{proof}

\begin{lemma}[Geometric discount horizon]
\label{lemma:geometric-discount-horizon}
Let $P_t$ be row-stochastic, $\gamma\in(0,1)$, and
$\eta^\star=(1-\gamma)/\gamma$. Then
\[
  (\Lout_t+\eta^\star I)^{-1}
  =
  \gamma(I-\gamma P_t)^{-1}
  =
  \gamma\sum_{k\geq0}(\gamma P_t)^k .
\]
Thus, for any $z_t\geq0$,
\[
  \bigl((\Lout_t+\eta^\star I)^{-1}z_t\bigr)(s)
  =
  \gamma\,
  \bbE_{P_t}\!\left[
    \sum_{k\geq0}\gamma^k z_t(X_k)\mid X_0=s
  \right].
\]
\end{lemma}

\begin{proof}
Since $\Lout_t=I-P_t$,
\[
  \Lout_t+\eta^\star I
  =
  \frac{1}{\gamma}(I-\gamma P_t).
\]
The inverse identity follows. Since $\norm{\gamma P_t}_\infty<1$, the
Neumann series applies. Finally,
$(P_t^kz_t)(s)=\bbE_{P_t}[z_t(X_k)\mid X_0=s]$.
\end{proof}

\begin{corollary}[Exact topology-priority sorter]
\label{cor:exact-sorter}
At $\eta=\eta^\star$, taking $z_t=|\xi_t|$ in
Lemma~\ref{lemma:geometric-discount-horizon} gives
\[
  \fout_t(s)
  =
  \bigl((\Lout_t+\eta^\star I)^{-1}|\xi_t|\bigr)(s)
  =
  \gamma\,
  \bbE_{P_t}\!\left[
    \sum_{k\geq0}\gamma^k|\xi_t(X_k)|\mid X_0=s
  \right].
\]
Hence the forward potential $\fout_t$ ranks states by expected discounted future Bellman residual, reflecting the proactive amount of future Bellman updating that will be applied through the backup chain.
\end{corollary}

\begin{lemma}\label{lemma:bellman-contraction}
The Bellman optimality operator $\mathcal{T}V(s) = r_t(s) + \gamma\max_a\sum_{s'}p(s'|s,a)V(s')$ is a $\gamma$-contraction in $\norm{\cdot}_\infty$, independent of the topology policy $\pi^{\mathrm{PS}}_t$.
\end{lemma}

\begin{proof}
Fix $V,V':\mathcal{S}\to\mathbb{R}$ and a state $s\in\mathcal{S}$. Let
\[
  a_s \in \arg\max_a \sum_{s'} p(s'|s,a)V(s'),
  \qquad
  a'_s \in \arg\max_a \sum_{s'} p(s'|s,a)V'(s').
\]
Using $a_s$ as a feasible action for the maximization that defines $\mathcal{T}V'(s)$,
\begin{align*}
  (\mathcal{T}V)(s)-(\mathcal{T}V')(s)
  &= \gamma \max_a \sum_{s'} p(s'|s,a)V(s')
     - \gamma \max_a \sum_{s'} p(s'|s,a)V'(s') \\
  &\leq \gamma \sum_{s'} p(s'|s,a_s)\bigl(V(s')-V'(s')\bigr) \\
  &\leq \gamma \norm{V-V'}_\infty \sum_{s'} p(s'|s,a_s)
   = \gamma \norm{V-V'}_\infty.
\end{align*}
Exchanging the roles of $V$ and $V'$ gives
\[
  (\mathcal{T}V')(s)-(\mathcal{T}V)(s) \leq \gamma \norm{V-V'}_\infty.
\]
Hence
\[
  \abs{(\mathcal{T}V)(s)-(\mathcal{T}V')(s)} \leq \gamma \norm{V-V'}_\infty.
\]
Taking the supremum over $s\in\mathcal{S}$ yields
\[
  \norm{\mathcal{T}V-\mathcal{T}V'}_\infty \leq \gamma \norm{V-V'}_\infty.
\]
The formula for $\mathcal{T}$ depends only on $p$ and $r_t$, so this contraction property is independent of the topology policy $\pi_t^{\mathrm{PS}}$.
\end{proof}

\appendixsubsection{Proofs of Theoretical Results}
\label{app:proofs}

\subsection*{Proof of Lemma~\ref{lemma:infinite-visit-convergence}}
\label{app:proof-infinite-visit-convergence}
\begin{proof}
Let
\[
  E_k \triangleq \norm{V_k-V^*}_\infty.
\]
At backup step $k$, only the selected state $s_k$ is updated:
\[
  V_{k+1}(s_k)=(\mathcal{T}V_k)(s_k),
  \qquad
  V_{k+1}(s)=V_k(s)\ \text{for } s\neq s_k.
\]
Since $V^*=\mathcal{T}V^*$ and $\mathcal{T}$ is a $\gamma$-contraction by \lemref{lemma:bellman-contraction},
\[
  \abs{V_{k+1}(s_k)-V^*(s_k)}
  = \abs{(\mathcal{T}V_k)(s_k)-(\mathcal{T}V^*)(s_k)}
  \leq \gamma E_k.
\]
For every $s\neq s_k$, the error is unchanged:
\[
  \abs{V_{k+1}(s)-V^*(s)}=\abs{V_k(s)-V^*(s)}\leq E_k.
\]
Hence $E_{k+1}\leq E_k$ for every $k$, so the error sequence is nonincreasing.

Now define a sequence of cover times. Let $N_0=0$. Since every state is selected infinitely often and $\mathcal{S}$ is finite, there exists a finite time $N_1>N_0$ such that every state appears at least once among
\[
  s_{N_0}, s_{N_0+1}, \dots, s_{N_1-1}.
\]
Inductively, once $N_m$ is defined, there exists $N_{m+1}>N_m$ such that every state appears at least once among
\[
  s_{N_m}, s_{N_m+1}, \dots, s_{N_{m+1}-1}.
\]

Fix $m\geq 0$ and a state $s\in\mathcal{S}$. Let $j_s\in\{N_m,\dots,N_{m+1}-1\}$ be the last time in this block at which state $s$ is updated. Immediately after that update,
\[
  \abs{V_{j_s+1}(s)-V^*(s)} \leq \gamma E_{j_s} \leq \gamma E_{N_m},
\]
because $E_k$ is nonincreasing. Since $s$ is not updated again between times $j_s+1$ and $N_{m+1}$, its value stays fixed on that interval, so
\[
  \abs{V_{N_{m+1}}(s)-V^*(s)}
  = \abs{V_{j_s+1}(s)-V^*(s)}
  \leq \gamma E_{N_m}.
\]
This bound holds for every state $s$, therefore
\[
  E_{N_{m+1}} \leq \gamma E_{N_m}.
\]
Iterating gives
\[
  E_{N_m} \leq \gamma^m E_0 \xrightarrow[m\to\infty]{} 0.
\]
Finally, if $N_m \leq k < N_{m+1}$, then $E_k\leq E_{N_m}$ because the error sequence is nonincreasing. Hence $E_k\to 0$ as $k\to\infty$, i.e.
\[
  V_k \to V^* \quad\text{in } \norm{\cdot}_\infty.
\]
\end{proof}

\subsection*{Proof of Proposition~\ref{prop:gtaps-infinite-visit}}
\label{app:proof-gtaps-infinite-visit}

\begin{proof}
Assume that no further reward shifts occur, so the Bellman operator
\(\mathcal T\) is fixed.

\medskip
\noindent\textbf{Queue mechanism}
We restate the notations in
\ref{app:b-gating-mechanism}. Let
\[
  \xi_t(s) \triangleq |(\mathcal T V_t)(s)-V_t(s)|,
  \qquad
  \mathcal I_t \triangleq \{s\in\mathcal S:|\xi_t(s)|>0\}.
\]
For convenience, the argument below treats the residual-active steps; when $\mathcal I_t=\varnothing$ the gate triggers the exploratory fallback of \ref{app:b-gating-mechanism}, and fallback steps only add selections to the accounting.

The implementation maintains the residual queue exactly over the nonzero
reactive residuals,
\[
  \mathrm{Keys}(\mathcal Q_\xi(t))=\mathcal I_t,
\]
and a separate topology queue \(\mathcal Q_\tau(t)\). 
Backups are selected by the combined priority
\[
  s_t \in
  \arg\max_{s\in\mathcal I_t}
  \left\{
    |\xi_t(s)|+\beta_t\tau_t(s)
  \right\}.
\]

The topology score is computed from the full residual field through the
outgoing and incoming directed Laplacian resolvents:
\[
  \mathbf f^{out}_t
  \triangleq
  (L^{out}_t+\eta^* I)^{-1}|\boldsymbol\xi_t|,
  \qquad
  \mathbf f^{in}_t
  \triangleq
  (L^{in}_t+\eta^* I)^{-1}|\boldsymbol\xi_t|.
\]
The realized topology signal is the in--out mixture
\begin{equation}\label{eq:full-inout-signal}
  \tau_t(s)
  =
  (1-\alpha_t)\mathbf |f^{out}_t(s)|
  +
  \alpha_t\mathbf |f^{in}_t(s)|,
  \qquad
  \alpha_t\in[0,1].
\end{equation}

\medskip
\noindent\textbf{Topology signal bound}
By Assumption~\ref{app:a-assm-bounded-reward}, the reward signal is uniformly
bounded; hence the Bellman targets and residuals remain finite. Moreover, by
\lemref{lemma:resolvent},
\[
  \norm{(L^{out}_t+\eta^* I)^{-1}}_\infty
  =
  \frac{1}{\eta^*},
  \qquad
  \norm{(L^{in}_t+\eta^* I)^{-1}}_\infty
  =
  \frac{1}{\eta^*}.
\]
Therefore,
\begin{equation}\label{eq:tau-bound}
  0\leq \tau_t(s)
  \leq
  \frac{1}{\eta^*}\|\boldsymbol\xi_t\|_\infty
  <\infty .
\end{equation}

The mixing-time modulation is
\[
  \beta_t = \beta_1(1 - \mathrm{SLEM}(P_t) + \beta_2).
\]
Since $\mathrm{SLEM}(P_t) \geq 0$ and $\mathrm{SLEM}(P_t) \leq 1$, we have $\beta_1\beta_2 \leq \beta_t \leq \beta_1(1 + \beta_2) \triangleq B_\beta$. The strictly positive lower bound $\beta_t \geq \beta_1\beta_2 > 0$ ensures $\beta_t\tau_t(s) \geq 0$ for every $s$. Combining this with \eqref{eq:tau-bound} gives
\begin{equation}\label{eq:beta-tau-bound}
  0 \leq \beta_t\tau_t(s)
  \leq
  \frac{B_\beta}{\eta^*}\|\boldsymbol\xi_t\|_\infty
  \triangleq
  C_\tau\|\boldsymbol\xi_t\|_\infty .
\end{equation}

\medskip
\noindent\textbf{Infinite-visit property.} Let
\begin{align}
  F &\triangleq \{ s \in \calS \mid s \text{ is selected for backup only finitely many times} \}, \\
  U &\triangleq \calS \setminus F = \{ u \in \calS \mid u \text{ is selected for backup infinitely often} \}.
\end{align}
If $F = \emptyset$, every state is selected infinitely often, and \lemref{lemma:infinite-visit-convergence} immediately guarantees convergence $V_t \to V^*$.

Suppose, for contradiction, that $F \neq \emptyset$.
Since states in $F$ are updated only finitely many times, there exists a finite timestamp $N < \infty$ such that no state in $F$ is selected for backup at any time step $t \ge N$. Consequently, the value estimates for states in $F$ remain permanently frozen:
\begin{equation}\label{eq:V-frozen}
  V_t(f) = V_N(f) \quad \text{for all } f \in F \text{ and all } t \ge N.
\end{equation}

Regard $F$ as fixed boundary data and define the restricted Bellman operator $\mathcal{T}_F : \bbR^{\abs{U}} \to \bbR^{\abs{U}}$ on $U$ by:
\begin{equation}\label{eq:TF-operator-def}
  (\mathcal{T}_F W)(u) \triangleq (\mathcal{T} V)(u) \quad \text{for } u \in U, \text{ where } V = W \oplus V_N|_F.
\end{equation}
Updates to states $u \in U$ observe frozen boundary values $V_N|_F$. Thus, $\mathcal{T}_F$ acts as a fixed $\gamma$-contraction on $\bbR^{\abs{U}}$ for all $t \ge N$. Since every state in $U$ is updated infinitely often, applying \lemref{lemma:infinite-visit-convergence} to this restricted subsystem guarantees:
\begin{equation}\label{eq:U-residual-decay}
  \lim_{t \to \infty} \max_{u \in U} \abs{\xi_t(u)} = 0.
\end{equation}

Now suppose, toward a contradiction, that $F$ retains a persistent non-zero residual: there exists at least one state $f_1 \in F$ whose residual does not vanish, i.e., there exists a constant $\delta > 0$ such that at infinitely many timestamps $t_k \to \infty$:
\begin{equation}\label{eq:persistent-residual-F}
  M_F(t_k) \triangleq \max_{f \in F} \abs{\xi_{t_k}(f)} = \abs{\xi_{t_k}(f_1)} \ge \delta > 0.
\end{equation}

Set $\varepsilon \triangleq \frac{(1 - C_\tau) \delta}{2} > 0$. By~\eqref{eq:U-residual-decay}, there exists a timestamp $N_1 \ge N$ such that for all $t \ge N_1$:
\begin{equation}\label{eq:U-residual-bound}
  \max_{u \in U} \abs{\xi_t(u)} < \frac{(1 - C_\tau) \delta}{2}.
\end{equation}

Fix an arbitrary timestamp $t_k \ge N_1 > N$ where $\abs{\xi_{t_k}(f_1)} = M_F(t_k) \ge \delta$. Because $\max_{u \in U} \abs{\xi_{t_k}(u)} < \frac{(1 - C_\tau)\delta}{2} < \delta \le M_F(t_k)$, the overall supremum norm of the residual vector is dominated entirely by the frozen state $f_1$:
\begin{equation}\label{eq:norm-eval-f1}
  \norm{\boldsymbol{\xi}_{t_k}}_\infty = \max \left\{ \max_{u \in U} \abs{\xi_{t_k}(u)}, \, M_F(t_k) \right\} = M_F(t_k) = \abs{\xi_{t_k}(f_1)}.
\end{equation}

We now evaluate and compare the priority scores at time step $t_k$:
\begin{itemize}
  \item \textbf{For the state $f_1 \in F$:} Since $\beta_{t_k} \tau_{t_k}(f_1) \ge 0$, its priority score satisfies:
  \begin{equation}\label{eq:priority-f1-eval}
    \prio_{t_k}(f_1) = \abs{\xi_{t_k}(f_1)} + \beta_{t_k} \tau_{t_k}(f_1) \ge \abs{\xi_{t_k}(f_1)}.
  \end{equation}

  \item \textbf{For any state $u \in U$:} Applying~\eqref{eq:beta-tau-bound} and~\eqref{eq:norm-eval-f1}:
  \begin{equation}\label{eq:priority-u-eval}
    \prio_{t_k}(u) = \abs{\xi_{t_k}(u)} + \beta_{t_k} \tau_{t_k}(u) \le \abs{\xi_{t_k}(u)} + C_\tau \norm{\boldsymbol{\xi}_{t_k}}_\infty = \abs{\xi_{t_k}(u)} + C_\tau \abs{\xi_{t_k}(f_1)}.
  \end{equation}
\end{itemize}

Computing the priority gap between $f_1 \in F$ and any $u \in U$:
\begin{align}
  \prio_{t_k}(f_1) - \prio_{t_k}(u) 
  &\ge \abs{\xi_{t_k}(f_1)} - \Bigl( \abs{\xi_{t_k}(u)} + C_\tau \abs{\xi_{t_k}(f_1)} \Bigr) \nonumber \\
  &= (1 - C_\tau) \abs{\xi_{t_k}(f_1)} - \abs{\xi_{t_k}(u)}. \label{eq:priority-gap-identity}
\end{align}
Substituting $\abs{\xi_{t_k}(f_1)} \ge \delta$ and $\abs{\xi_{t_k}(u)} < \frac{(1 - C_\tau)\delta}{2}$ into~\eqref{eq:priority-gap-identity}:
\begin{equation}\label{eq:priority-strict-positive}
  \prio_{t_k}(f_1) - \prio_{t_k}(u) > (1 - C_\tau) \delta - \frac{(1 - C_\tau) \delta}{2} = \frac{(1 - C_\tau) \delta}{2} > 0 \quad \text{for all } u \in U.
\end{equation}

Since $C_\tau < 1$ (guaranteed by the hypothesis $\beta_1(1 + \beta_2) < \eta^*$), the gap $\frac{(1 - C_\tau) \delta}{2}$ is strictly positive. Thus:
\begin{equation}
  \prio_{t_k}(f_1) > \max_{u \in U} \prio_{t_k}(u).
\end{equation}
Since GTA-PS selects the state with the maximum priority at each backup step, the algorithm must select state $f_1 \in F$ for backup at time step $t_k \ge N_1 > N$.

This directly contradicts the definition of $N$, which asserted that no state in $F$ is ever selected after step $N$. Therefore, $F = \emptyset$, proving that every state $s \in \calS$ is visited infinitely often. Consequently, $\norm{\boldsymbol{\xi}_t}_\infty \to 0$.

\medskip
\noindent\textbf{Convergence to $V^*$.}
Using $V^*=\mathcal T V^*$ and the $\gamma$-contractivity of $\mathcal T$,
\[
\begin{aligned}
  \|V_t-V^*\|_\infty
  &\leq
  \|V_t-\mathcal T V_t\|_\infty
  +\|\mathcal T V_t-\mathcal T V^*\|_\infty  \\
  &\leq
  \|\boldsymbol\xi_t\|_\infty
  +\gamma\|V_t-V^*\|_\infty.
\end{aligned}
\]
Rearranging,
\[
  (1-\gamma)\|V_t-V^*\|_\infty
  \leq
  \|\boldsymbol\xi_t\|_\infty
  \to0.
\]
Therefore GTA-PS$^\beta$ converges deterministically to $V^*$ in the
no-shift regime.
\end{proof}

\paragraph{Interpretive items.} The two items below record the operator facts behind the design of Section~\ref{sec:gta-ps} and the lens of Section~\ref{subsec:lens}; the guarantees of Section~\ref{sec:theory} stand on the independent chain in Sections~\ref{subsec:convergence}--\ref{subsec:nonstationary}.

\begin{lemma}[Green's function spectral representation]
\label{lemma:green}
Given that the kernel $P_t$ is diagonalizable (\corref{cor:kernel-well-defined}) with a biorthogonal eigendecomposition
$P_t = \sum_k \kappa_k v_k u_k^\top$ and $u_j^\top v_k = \delta_{jk}$.
Then for any $\eta>0$, the forward potential has the spectral expansion
\[
  \mathbf{f}^{out}_t = \sum_k \frac{u_k^\top |\xi_t|}{1-\kappa_k+\eta}\,v_k,
\]
so slow modes are amplified by at most $1/\eta$. The statement is relocated here from the main text so that Section~\ref{subsec:lens} stays interpretive.
\end{lemma}

\subsection*{Proof of Lemma~\ref{lemma:green}}
\label{app:proof-green}
\begin{proof}
The biorthogonal decomposition gives
\[
  P_t = \sum_k \kappa_k v_k u_k^\top,
  \qquad
  u_j^\top v_k = \delta_{jk}.
\]
Let
\[
  M_t \triangleq L_t^{out}+\eta I = (1+\eta)I-P_t.
\]
For each eigenpair $(\kappa_k,v_k)$ of $P_t$,
\[
  M_t v_k
  = \bigl((1+\eta)I-P_t\bigr)v_k
  = (1+\eta-\kappa_k)v_k
  = (1-\kappa_k+\eta)v_k.
\]
Since $P_t$ is stochastic, every eigenvalue satisfies $|\kappa_k|\leq 1$. Therefore
\[
  \mathrm{Re}(1-\kappa_k+\eta)=1-\mathrm{Re}(\kappa_k)+\eta \geq \eta > 0,
\]
so $1-\kappa_k+\eta\neq 0$ for every $k$. Thus $M_t$ is invertible on every eigendirection and
\[
  M_t^{-1}v_k = \frac{1}{1-\kappa_k+\eta}v_k.
\]

Because $P_t$ is diagonalizable, the set of right eigenvectors $\{v_k\}_{k=1}^n$ forms a basis for $\bbR^n$. Any residual vector $\xi_t \in \bbR^n$ can thus be uniquely decomposed as a linear combination $\xi_t = \sum_{k=1}^n c_k v_k$. Multiplying from the left by $u_j^\top$ and using the biorthogonality relation $u_j^\top v_k = \delta_{jk}$:
\[
  u_j^\top \xi_t = u_j^\top \left( \sum_{k=1}^n c_k v_k \right) = \sum_{k=1}^n c_k (u_j^\top v_k) = c_j.
\]
Substituting $c_k = u_k^\top \xi_t$ gives the modal expansion $\xi_t = \sum_k (u_k^\top \xi_t)\,v_k$.
Applying $M_t^{-1}$ termwise and using $M_t^{-1}v_k = \frac{1}{1-\kappa_k+\eta}v_k$ yields:
\begin{align*}
  \mathbf{f}_t^{out}
  &= M_t^{-1}\xi_t \\
  &= \sum_k (u_k^\top\xi_t)\,M_t^{-1}v_k \\
  &= \sum_k \frac{u_k^\top\xi_t}{1-\kappa_k+\eta}\,v_k.
\end{align*}
This proves the claimed spectral expansion.

For the amplification bound, each modal coefficient is multiplied by $(1-\kappa_k+\eta)^{-1}$, and
\[
  \abs{1-\kappa_k+\eta} \geq \mathrm{Re}(1-\kappa_k+\eta) \geq \eta.
\]
Hence
\[
  \frac{1}{\abs{1-\kappa_k+\eta}} \leq \frac{1}{\eta},
\]
so no mode is amplified by more than $1/\eta$. The same argument applies to periodic chains: even if a nontrivial eigenvalue lies on the unit circle, the positive regularization parameter $\eta$ keeps the denominator away from zero.

\end{proof}

\begin{remark}[Doob $h$-transform reading]
\label{prop:importance-sampling}
Fix a single-site shift at $s_r$ and define the Green's predecessor entry $\psi(s)\triangleq G^{out}_{\eta^\star,t}(s,s_r)>0$. Then
\[
  \tilde P_t(s,s')
  \triangleq
  \frac{\gamma P_t(s,s')\,\psi(s')}{\psi(s)}
\]
is the Doob $h$-transform of the $\gamma$-killed topology chain. Along any path $(X_0,\ldots,X_k)$, the associated change-of-measure weight telescopes as $\psi(X_k)/\psi(X_0)$. In this sense, ranking states by $\psi(s)$ is an implicit \textbf{importance-sampling} rule over propagation paths toward the shifted reward site. The reading is relocated here from the main text so that Section~\ref{subsec:lens} stays interpretive.
\end{remark}

\subsection*{Proof of the telescoping identity in the Doob reading}
\label{app:proof-importanct-sampling}
\begin{proof}
Set $P^\star_t\triangleq\gamma P_t$. By the Neumann series,
$\psi(s)=G_{P^\star_t}(s,s_r)$.
Positivity $\psi(s)>0$ follows from
Assumption~\ref{app:a-assm-communicating}: for every $s$ there exists
$m\geq1$ with $P_t^m(s,s_r)>0$, so
$G_{P^\star_t}(s,s_r)\geq\gamma^m P_t^m(s,s_r)>0$.
The identity $\eta^*=(1-\gamma)/\gamma$ gives
$L^{out}_t+\eta^*I=\tfrac{1}{\gamma}(I-\gamma P_t)$, hence
$(L^{out}_t+\eta^*I)^{-1}=\gamma(I-\gamma P_t)^{-1}$, confirming
$\psi(s)=G^{out}_{\eta^*,t}(s,s_r)$.

\medskip\noindent\textbf{Harmonicity.}
The resolvent identity $Q(I-Q)^{-1}=(I-Q)^{-1}-I$ with $Q=\gamma P_t$
gives
\begin{equation}\label{eq:harm}
  (P^\star_t\psi)(s)
  =\gamma\bigl[\gamma P_t(I-\gamma P_t)^{-1}\bigr]_{s,s_r}
  =\gamma\bigl[(I-\gamma P_t)^{-1}-I\bigr]_{s,s_r}
  =\psi(s)-\gamma\delta_{s,s_r},
\end{equation}
so $P^\star_t\psi=\psi$ pointwise on $\mathcal S\setminus\{s_r\}$.

\medskip\noindent\textbf{Row sums of $\tilde P_t$.}
From \eqref{eq:harm},
\[
  \sum_{s'}\tilde P_t(s,s')
  =\frac{(P^\star_t\psi)(s)}{\psi(s)}
  =1-\frac{\gamma\,\delta_{s,s_r}}{\psi(s)}.
\]
Hence $\tilde P_t$ is stochastic on $\mathcal S\setminus\{s_r\}$ and
has row-sum defect $\gamma/\psi(s_r)$ at $s_r$, which is the standard
Doob absorption at the target site \citep[Thm.~7.2]{woess2000random}.

\medskip\noindent\textbf{Telescoping and importance sampling.}
For any $P^\star_t$-positive path $(X_0,\ldots,X_k)$,
\[
  \frac{d\tilde{\mathbb P}}{d\mathbb P}
  \bigl[(X_0,\ldots,X_k)\bigr]
  \;=\;
  \prod_{j=0}^{k-1}
  \frac{\tilde P_t(X_j,X_{j+1})}{P^\star_t(X_j,X_{j+1})}
  \;=\;
  \prod_{j=0}^{k-1}
  \frac{\psi(X_{j+1})}{\psi(X_j)}
  \;=\;
  \frac{\psi(X_k)}{\psi(X_0)}.
\]
The initial factor $\psi(X_0)^{-1}$ is the exact importance weight for
the path-measure change. Since $\psi(s)=G_{P^\star_t}(s,s_r)$ equals the
$\tilde P_t$-expected discounted hitting mass at $s_r$
\citep[Prop.~7.4]{woess2000random}, ranking by $\psi(s)$ concentrates
backups on the states with maximal $\tilde P_t$-path weight toward the
reward-shift site.
\end{proof}

\subsection*{Proof for Proposition~\ref{prop:dense-support}}
\begin{proof}
\textit{(i).}
At episode start after a single-site reward shift,
\[
  \xi_t = \Delta_r e_{s_r}.
\]
By \corref{cor:exact-sorter},
\[
  \tau_t
  = \gamma(I-\gamma P_t)^{-1}\xi_t
  = \gamma\Delta_r\sum_{k=0}^{\infty}(\gamma P_t)^k e_{s_r}.
\]
Evaluating the $s$-th component gives
\[
  \tau_t(s)
  = \gamma\Delta_r\sum_{k=0}^{\infty}\gamma^k [P_t^k]_{s,s_r}.
\]
Every term in this series is nonnegative. Because $P_t$ is finite and irreducible by Assumption~\ref{app:a-assm-communicating}, for every state $s \in \calS$, there exists a finite path length $k^* = k^*(s, s_r) \ge 0$ such that $[P_t^{k^*}]_{s, s_r} > 0$.
Retaining the $k^*$-th term yields:
\begin{equation}\label{eq:tau-strict-positive}
  \tau_t(s) \ge \gamma \Delta_r \, \gamma^{k^*} [P_t^{k^*}]_{s, s_r} > 0 \quad \text{for all } s \in \calS.
\end{equation}
This proves dense priority support.

\textit{(ii).}
Fix $s\in\mathcal{B}_K$, by single-site reward shift assumption, for $s \in \mathcal{B}_K$
\[
  h^*(s;s_r)>K.
\]
By definition of the backward predecessor distance, there is no path
\[
  s=s_0,s_1,\dots,s_\ell=s_r
\]
with $\ell\leq K$ and $P_t(s_{i+1},s_i)>0$ for all $i$. Reactive PS propagates nonzero priority from the shifted reward state $s_r$ only through predecessor pushes, one predecessor edge per backup. Therefore after at most $K$ reactive backups, the residual wavefront can reach only states at backward predecessor distance at most $K$ from $s_r$. Since $s$ lies outside that $K$-step predecessor set, its reactive PS priority remains zero:
\[
  p_t^{\mathrm{PS}}(s)=0.
\]
By part~(i), the GTA-PS topology signal still satisfies $\tau_t(s)>0$. Hence GTA-PS assigns nonzero topology priority to every blocked state while reactive PS does not. This proves part~(ii).
\end{proof}

\begin{remark}[Blocked-set non-emptiness]\label{rem:mfpt-lb}
Under the communicating assumption~\ref{app:a-assm-communicating}, define the backward directed eccentricity of the shifted reward state by
\[
  D^*(s_r)\triangleq \max_{s\in\mathcal{S}} h^*(s;s_r).
\]
Then the blocked set $\mathcal{B}_K=\{s:h^*(s;s_r)>K\}$ is non-empty whenever
$K<D^*(s_r)$.
In bottlenecked environments, $D^*(s_r)$ is large relative to the backup budget, so blocked states exist for the budgets where GTA-PS is intended to help.

For periodic chains, the same structural claim still holds because it depends only on the support graph. Independently, the regularized resolvent remains well-defined because $\eta^\star=(1-\gamma)/\gamma>0$.
\end{remark}

\subsection*{Proof for Lemma~\ref{lemma:adv-lipschitz} and Proposition~\ref{prop:cumulative-advantage}}
\label{app:proof-cumulative-advantage}

\begin{proof}
We separate the argument into the fixed-kernel ergodic average and the drift perturbation.
 
\textbf{Fixed-kernel ergodic average.}
For a fixed stochastic kernel $P$, Assumption~\ref{app:a-assm-ergodic-reward} makes the reward-shift descriptor
$Z_t=(s_{r,t},\Delta_{r,t})$, and hence $\zeta_t=\Delta_{r,t}e_{s_{r,t}}$,
stationary-ergodic.
\[
  \zeta \longmapsto \mathrm{adv}_t(P,\zeta)
\]
is measurable and integrable, then Birkhoff's ergodic theorem gives
\[
  \frac{1}{N}\sum_{t=1}^{N}\mathrm{adv}_t(P,\zeta_t)
  \xrightarrow[\;N\to\infty\;]{\mathrm{a.s.}}
  \bar{A}(P)
  \triangleq
  \mathbb{E}_{\mu_R}[\mathrm{adv}_t(P,\zeta)].
\]
Because the state space is finite and $\tau(P,\zeta)\triangleq\gamma(I-\gamma P)^{-1}\zeta$ is obtained by a bounded linear map applied to the bounded random vector $\zeta$, the functional $\mathrm{adv}(P,\zeta)$ is measurable and integrable.
 
The positive-frequency assumption $|\mathcal{B}_K^{\mathrm{sel},t}|\geq 1$ implies that
\[
  \mu_R\bigl(\mathrm{adv}_t(P,\zeta_t)>0\bigr)>0
\]
under the frozen-kernel process. Since each summand in the definition of $\mathrm{adv}$ is nonnegative, and $\mathbb{E}[X]\geq\mathbb{E}[X\,\mathbf{1}_{X>0}]>0$ whenever $X\geq 0$ and $P(X>0)>0$, it follows that
\[
  \bar A(P)>0.
\]
 
\textbf{Decomposition around a reference kernel.}
By Assumption~\ref{app:a-assm-slow-drift}, there exists a frozen reference
kernel $\bar P$ such that
$\|P_t-\bar P\|_\infty\leq\varepsilon_{\mathrm{pol}}$ for all $t$.
Then
\[
  \mathrm{adv}_t(P_t,\zeta_t)
  =
  \mathrm{adv}_t(\bar P,\zeta_t)
  +
  \Bigl(\mathrm{adv}_t(P_t,\zeta_t)-\mathrm{adv}_t(\bar P,\zeta_t)\Bigr).
\]
Summing over $t$ and dividing by $N$ yields
\[
  \frac{1}{N}\sum_{t=1}^{N}\mathrm{adv}_t(P_t,\zeta_t)
  =
  \frac{1}{N}\sum_{t=1}^{N}\mathrm{adv}_t(\bar P,\zeta_t)
  +
  \frac{1}{N}\sum_{t=1}^{N}\Bigl(\mathrm{adv}_t(P_t,\zeta_t)-\mathrm{adv}_t(\bar P,\zeta_t)\Bigr).
\]
The first term converges almost surely to $\bar A(\bar P)$ by Step~1; write the convergence error as $e_N \triangleq \frac{1}{N}\sum_t\mathrm{adv}_t(\bar P,\zeta_t) - \bar{A}(\bar{P}) = o(1)$ a.s.
 
\textbf{Drift error bound.}
For two kernels $P,P'$, the resolvent identity gives
\begin{align*}
  (I-\gamma P)^{-1}-(I-\gamma P')^{-1}
  &= (I-\gamma P)^{-1}\,\gamma(P-P')\,(I-\gamma P')^{-1},
\end{align*}
and therefore, recalling $\tau(P,\zeta)=\gamma(I-\gamma P)^{-1}\zeta$,
\[
  \tau(P,\zeta)-\tau(P',\zeta)
  = \gamma^2(I-\gamma P)^{-1}(P-P')(I-\gamma P')^{-1}\zeta.
\]
Since $\|P\|_\infty=1$ implies $\|(I-\gamma P)^{-1}\|_\infty\leq 1/(1-\gamma)$ via the Neumann series,
\[
  \norm{\tau(P,\zeta)-\tau(P',\zeta)}_\infty
  \leq \frac{\gamma^2}{(1-\gamma)^2}\norm{P-P'}_\infty\norm{\zeta}_\infty.
\]
\medskip
\noindent\textbf{1-Lipschitz property of the order-statistic map $x \mapsto x_{(k)}$.}
Let $x, y \in \bbR^n$ and $\varepsilon \triangleq \norm{x - y}_\infty$. Then $\abs{x_i - y_i} \le \varepsilon$, which gives $x_i - \varepsilon \le y_i \le x_i + \varepsilon$ for all $i \in \{1, \dots, n\}$.
Let $x_{(1)} \ge x_{(2)} \ge \dots \ge x_{(n)}$ and $y_{(1)} \ge y_{(2)} \ge \dots \ge y_{(n)}$ be the ordered values. We prove that $\abs{x_{(k)} - y_{(k)}} \le \norm{x - y}_\infty$.
Suppose for contradiction that $x_{(k)} - y_{(k)} > \varepsilon$, which implies $x_{(k)} - \varepsilon > y_{(k)}$.
Define the index set:
\[
  I \triangleq \{ i \in \{1, \dots, n\} \mid x_i \ge x_{(k)} \}.
\]
By definition of the $k$-th order statistic, $\abs{I} \ge k$.
For every index $i \in I$:
\[
  y_i \ge x_i - \varepsilon \ge x_{(k)} - \varepsilon > y_{(k)}.
\]
Thus, $y_i > y_{(k)}$ for at least $k$ distinct indices $i \in I$.
However, by definition of the $k$-th largest entry, there can be at most $k - 1$ entries of $y$ strictly greater than $y_{(k)}$, which is a contradiction.
By symmetry, $y_{(k)} - x_{(k)} \le \varepsilon$. Therefore:
\[
  \abs{x_{(k)} - y_{(k)}} \le \varepsilon = \norm{x - y}_\infty.
\]
Applying this property to the $(q+1)$-th component gives:
\begin{equation}\label{eq:order-stat-1lip}
  \abs{\tau_{(q+1)}(P,\zeta)-\tau_{(q+1)}(P',\zeta)} \leq \norm{\tau(P,\zeta)-\tau(P',\zeta)}_\infty.
\end{equation}

\medskip
\noindent\textbf{1-Lipschitz property of the positive-part map $x \mapsto (x)_+$.}
For any $a, b \in \bbR$, recall $(x)_+ \triangleq \max(0, x) = \frac{x + \abs{x}}{2}$. Then:
\begin{align}\label{eq:pos-part-1lip}
  \abs{(a)_+ - (b)_+} &= \abs{\frac{a + \abs{a}}{2} - \frac{b + \abs{b}}{2}} = \abs{\frac{(a - b) + (\abs{a} - \abs{b})}{2}} \nonumber \\
  &\le \frac{\abs{a - b} + \abs{\abs{a} - \abs{b}}}{2} \le \frac{\abs{a - b} + \abs{a - b}}{2} = \abs{a - b}.
\end{align}

\medskip
\noindent\textbf{Evaluation of the advantage difference $\abs{\mathrm{adv}(P,\zeta) - \mathrm{adv}(P',\zeta)}$.}
Recalling $\mathrm{adv}(P,\zeta)=\frac{1}{\gamma}\sum_{s\in\mathcal{B}_K(s_r)}\bigl(\tau(P,\zeta)(s)-\tau_{(q+1)}(P,\zeta)\bigr)_+$, we apply~\eqref{eq:order-stat-1lip} and~\eqref{eq:pos-part-1lip}:
\begin{align*}
  &\abs{\mathrm{adv}(P,\zeta)-\mathrm{adv}(P',\zeta)} \\
  &= \frac{1}{\gamma} \abs{ \sum_{s \in \mathcal{B}_K(s_r)} \Bigl( \tau(P,\zeta)(s) - \tau_{(q+1)}(P,\zeta) \Bigr)_+ - \sum_{s \in \mathcal{B}_K(s_r)} \Bigl( \tau(P',\zeta)(s) - \tau_{(q+1)}(P',\zeta) \Bigr)_+ } \\
  &\le \frac{1}{\gamma} \sum_{s \in \mathcal{B}_K(s_r)} \abs{ \Bigl( \tau(P,\zeta)(s) - \tau_{(q+1)}(P,\zeta) \Bigr)_+ - \Bigl( \tau(P',\zeta)(s) - \tau_{(q+1)}(P',\zeta) \Bigr)_+ } \\
  &\le \frac{1}{\gamma} \sum_{s \in \mathcal{B}_K(s_r)} \abs{ \Bigl( \tau(P,\zeta)(s) - \tau_{(q+1)}(P,\zeta) \Bigr) - \Bigl( \tau(P',\zeta)(s) - \tau_{(q+1)}(P',\zeta) \Bigr) } \\
  &= \frac{1}{\gamma} \sum_{s \in \mathcal{B}_K(s_r)} \abs{ \Bigl( \tau(P,\zeta)(s) - \tau(P',\zeta)(s) \Bigr) - \Bigl( \tau_{(q+1)}(P,\zeta) - \tau_{(q+1)}(P',\zeta) \Bigr) } \\
  &\le \frac{1}{\gamma} \sum_{s \in \mathcal{B}_K(s_r)} \left( \abs{ \tau(P,\zeta)(s) - \tau(P',\zeta)(s) } + \abs{ \tau_{(q+1)}(P,\zeta) - \tau_{(q+1)}(P',\zeta) } \right) \\
  &\le \frac{1}{\gamma} \sum_{s \in \mathcal{B}_K(s_r)} \left( \norm{\tau(P,\zeta) - \tau(P',\zeta)}_\infty + \norm{\tau(P,\zeta) - \tau(P',\zeta)}_\infty \right) \\
  &= \frac{2 |\mathcal{B}_K(s_r)|}{\gamma} \norm{\tau(P,\zeta) - \tau(P',\zeta)}_\infty \\
  &\le \frac{2 |\mathcal{B}_K(s_r)|\,\gamma}{(1-\gamma)^2} \norm{\zeta}_\infty \norm{P-P'}_\infty.
\end{align*}
Since $\norm{\zeta_t}_\infty=\Delta_{r,t}\leq \Delta_{\max}$ (the maximum absolute reward perturbation across states at time $t$) and $|\mathcal{B}_K(s_r)|\leq |\mathcal{S}|$, the functional is uniformly Lipschitz in $P$ with constant
\[
  L \triangleq \frac{2|\mathcal{S}|\gamma\Delta_{\max}}{(1-\gamma)^2}.
\]
Therefore
\[
  \abs{\mathrm{adv}_t(P_t,\zeta_t)-\mathrm{adv}_t(\bar P,\zeta_t)}
  \leq L \norm{P_t-\bar P}_\infty
  \leq L\varepsilon_{\mathrm{pol}}
\]
throughout the drift envelope. Hence
\[
  \frac{1}{N}\sum_{t=1}^{N}\Bigl(\mathrm{adv}_t(P_t,\zeta_t)-\mathrm{adv}_t(\bar P,\zeta_t)\Bigr)
  \geq -L\varepsilon_{\mathrm{pol}}.
\]
Combining with Step~2 gives
\[
  \frac{1}{N}\sum_{t=1}^{N}\mathrm{adv}_t(P_t,\zeta_t)
  \geq \bar A(\bar P)-L\varepsilon_{\mathrm{pol}}+e_N
  = \bar A(\bar P)-L\varepsilon_{\mathrm{pol}}+o(1)
  \qquad\text{a.s.}
\]
Renaming $\bar A(\bar P)$ as $\bar A$ yields the displayed conclusion.
\end{proof}

\subsection*{Proof of Theorem~\ref{thm:cumulative-tracking}}
\label{app:proof-cumulative-tracking}
\begin{proof}
\textbf{Part (i): per-episode GTA-PS residual.} By \corref{cor:exact-sorter},
\[
  \mathrm{err}^{\mathrm{GTA}}_t=\frac{\tau_{(q+1),t}}{\gamma}=\Phi_{(q+1),t}
  \leq \frac{\|\xi_t\|_\infty}{1-\gamma}
  \leq \frac{\Delta_{r,t}}{1-\gamma}.
\]
The first equality is the definition of $\mathrm{err}^{\mathrm{GTA}}_t$. The second is the exact-sorter identity. For the first inequality,
\[
  \Phi_t=(I-\gamma P_t)^{-1}\xi_t
  = \sum_{k=0}^{\infty}(\gamma P_t)^k\xi_t,
\]
so
\[
  \norm{\Phi_t}_\infty
  \leq \sum_{k=0}^{\infty}\gamma^k \norm{\xi_t}_\infty
  = \frac{1}{1-\gamma}\norm{\xi_t}_\infty.
\]
Since $\Phi_{(q+1),t}$ is one coordinate of $\Phi_t$, this implies
$\Phi_{(q+1),t}\leq \|\xi_t\|_\infty/(1-\gamma)$.
At episode start, under the single-shift assumption:
$\xi_t=\Delta_{r,t}e_{s_r}$.
Hence $\norm{\xi_t}_\infty=\Delta_{r,t}$ and the
final inequality follows.
 
\textbf{Part (ii): cumulative advantage reduction.}
Recall the formal definitions from Theorem~\ref{thm:cumulative-tracking}:
\begin{align}
  \mathrm{err}_t^{\mathrm{GTA}} &\triangleq \frac{\tau_{(q+1), t}}{\gamma}, \\
  \mathrm{TE}^{\mathrm{GTA}}(N) &\triangleq \sum_{t=1}^N \mathrm{err}_t^{\mathrm{GTA}} = \sum_{t=1}^N \frac{\tau_{(q+1), t}}{\gamma}, \\
  \mathrm{TE}^{\mathrm{PS}}(N) &\triangleq \sum_{t=1}^N \sum_{s^* \in \mathcal{B}_K^{\mathrm{sel}, t}} \Phi_t(s^*).
\end{align}
By \propref{prop:dense-support}(ii), reactive PS leaves all blocked states $s^* \in \mathcal{B}_K^{\mathrm{sel}, t}$ unresolved after $q = K$ backups, maintaining the future residual $\Phi_t(s^*) = \frac{\tau_t(s^*)}{\gamma}$. 
Substituting this into $\mathrm{TE}^{\mathrm{PS}}(N)$ gives:
\begin{equation}
  \mathrm{TE}^{\mathrm{PS}}(N) = \sum_{t=1}^N \sum_{s^* \in \mathcal{B}_K^{\mathrm{sel}, t}} \frac{\tau_t(s^*)}{\gamma}.
\end{equation}

Taking the difference between the two tracking errors:
\begin{equation}
  \mathrm{TE}^{\mathrm{PS}}(N) - \mathrm{TE}^{\mathrm{GTA}}(N) = \sum_{t=1}^N \sum_{s^* \in \mathcal{B}_K^{\mathrm{sel}, t}} \frac{\tau_t(s^*)}{\gamma} - \sum_{t=1}^N \frac{\tau_{(q+1), t}}{\gamma}.
\end{equation}
Since the selected blocked set satisfies $|\mathcal{B}_K^{\mathrm{sel}, t}| \ge 1$ at positive frequency by assumption, and because $\tau_{(q+1), t} \ge 0$, we have:
\begin{equation}
  \frac{\tau_{(q+1), t}}{\gamma} \le |\mathcal{B}_K^{\mathrm{sel}, t}| \cdot \frac{\tau_{(q+1), t}}{\gamma} = \sum_{s^* \in \mathcal{B}_K^{\mathrm{sel}, t}} \frac{\tau_{(q+1), t}}{\gamma}.
\end{equation}
Distributing the threshold subtraction across the summation over $\mathcal{B}_K^{\mathrm{sel}, t}$ yields the lower bound:
\begin{align}
  \mathrm{TE}^{\mathrm{PS}}(N) - \mathrm{TE}^{\mathrm{GTA}}(N) 
  &\ge \sum_{t=1}^N \left( \sum_{s^* \in \mathcal{B}_K^{\mathrm{sel}, t}} \frac{\tau_t(s^*)}{\gamma} - \sum_{s^* \in \mathcal{B}_K^{\mathrm{sel}, t}} \frac{\tau_{(q+1), t}}{\gamma} \right) \nonumber \\
  &= \sum_{t=1}^N \sum_{s^* \in \mathcal{B}_K^{\mathrm{sel}, t}} \frac{\tau_t(s^*) - \tau_{(q+1), t}}{\gamma}. \label{eq:diff-distributed}
\end{align}
Now, recall the explicit definition of the priority advantage functional:
\begin{align}
  \mathrm{adv}_t(P_t, \zeta_t) 
  &= \frac{1}{\gamma} \sum_{s \in \mathcal{B}_K(s_{r,t})} \Bigl( \tau_t(s) - \tau_{(q+1), t} \Bigr)_+ \nonumber \\
  &= \frac{1}{\gamma} \Biggl( \sum_{s \in \mathcal{B}_K^{\mathrm{sel}, t}} \Bigl( \tau_t(s) - \tau_{(q+1), t} \Bigr)_+ + \sum_{s \in \mathcal{B}_K(s_{r,t}) \setminus \mathcal{B}_K^{\mathrm{sel}, t}} \Bigl( \tau_t(s) - \tau_{(q+1), t} \Bigr)_+ \Biggr). \label{eq:adv-two-parts}
\end{align}
We analyze each partition directly:
\begin{itemize}
  \item For any state $s \in \mathcal{B}_K^{\mathrm{sel}, t}$, by definition $\tau_t(s) > \tau_{(q+1), t}$, meaning $\tau_t(s) - \tau_{(q+1), t} > 0$. Hence:
  \[
    \Bigl( \tau_t(s) - \tau_{(q+1), t} \Bigr)_+ = \tau_t(s) - \tau_{(q+1), t}.
  \]
  \item For any state $s \in \mathcal{B}_K(s_{r,t}) \setminus \mathcal{B}_K^{\mathrm{sel}, t}$, we have $\tau_t(s) \le \tau_{(q+1), t}$, which implies $\tau_t(s) - \tau_{(q+1), t} \le 0$. Thus:
  \[
    \Bigl( \tau_t(s) - \tau_{(q+1), t} \Bigr)_+ = 0.
  \]
\end{itemize}
The second sum in~\eqref{eq:adv-two-parts} vanishes completely, leaving:
\begin{equation}\label{eq:adv-simplified}
  \mathrm{adv}_t(P_t, \zeta_t) = \frac{1}{\gamma} \sum_{s \in \mathcal{B}_K^{\mathrm{sel}, t}} \Bigl( \tau_t(s) - \tau_{(q+1), t} \Bigr) = \sum_{s^* \in \mathcal{B}_K^{\mathrm{sel}, t}} \frac{\tau_t(s^*) - \tau_{(q+1), t}}{\gamma}.
\end{equation}

Substituting~\eqref{eq:adv-simplified} into~\eqref{eq:diff-distributed} produces the fundamental advantage relation:
\begin{equation}
  \mathrm{TE}^{\mathrm{PS}}(N) - \mathrm{TE}^{\mathrm{GTA}}(N) \ge \sum_{t=1}^N \mathrm{adv}_t(P_t, \zeta_t).
\end{equation}

Finally, applying the ergodic drift lower bound from Proposition~\ref{prop:cumulative-advantage}:
\begin{equation}
  \sum_{t=1}^N \mathrm{adv}_t(P_t, \zeta_t) \ge N \bigl( \bar{A} - L \varepsilon_{\mathrm{pol}} \bigr) + o(N) \quad \text{almost surely}.
\end{equation}
Under Assumption~\ref{app:a-assm-positive-local-advantage}, $\bar{A} > 0$. Since the policy drift satisfies $L \varepsilon_{\mathrm{pol}} < \bar{A}$, the linear rate $\bar{A} - L \varepsilon_{\mathrm{pol}} > 0$. Taking the limit as $N \to \infty$:
\begin{equation}
  \mathrm{TE}^{\mathrm{PS}}(N) - \mathrm{TE}^{\mathrm{GTA}}(N) \xrightarrow{N \to \infty} +\infty \quad \text{almost surely}.
\end{equation}
\end{proof}

\appendixsubsection{Horizon significance of the diffusion support}
\label{app:a-horizon}

Proposition~\ref{prop:dense-support} establishes that the forward field assigns positive mass to every state. This subsection quantifies how much of that support is useful at a finite budget, and when it degenerates. Fix a single reward site $s_r$, write $\rho_\eta=(1+\eta)^{-1}$, and let $h^*(s;s_r)$ be the predecessor backup distance used in Proposition~\ref{prop:dense-support}.

\paragraph{Outer-barrier mass.}
For a localized shift $\Delta_t e_{s_r}$, the Green's entry expands as
\[
  \tau_{\eta,t}^{\mathrm{out}}(s)
  = \Delta_t\left[((1+\eta)I-P_t)^{-1}\right]_{s,s_r}
  = \Delta_t\sum_{k\ge0}\rho_\eta^{k+1}[P_t^k]_{s,s_r}.
\]
With reactive budget $K$ and outer horizon $H\ge h^*(s;s_r)$,
\[
  \tau_{\eta,t}^{\mathrm{out}}(s)
  \ge \Delta_t\,\rho_\eta^{H+1}\sum_{k=K+1}^{H}[P_t^k]_{s,s_r},
\]
which retains only route mass beyond the reactive barrier. We term $\rho_\eta^{H+1}$ the \emph{horizon decay} and the summed path mass the \emph{route multiplicity}. Large $\eta$ and sparse kernels give the worst case, so the bound is instance-dependent and no kernel-independent positive lower bound exists.

\paragraph{Counterexample to the positive-advantage assumption.}
Assumption~\ref{app:a-assm-positive-local-advantage} can fail at logarithmic scale. Consider a deterministic binary tree with a single reward at the root. Every non-root state has one action that moves one level upward, and the root has $c$ structurally equivalent actions, each transitioning to every leaf with probability $2^{-h}$, where $h$ is the tree height. Starting from a leaf with probability approximately $1/2$ under a uniform start, let $s_d$ be a depth-$d$ state and $M_H(d)=\lfloor(H-d)/(h+1)\rfloor$ for $H\ge d$. For any policy over the root actions,
\[
  \xi^{(j)}(s_d)=\Delta_t\gamma^{d+j(h+1)},
  \qquad
  \tau_{\eta,t;H}^{\mathrm{out}}(s_d)
  = \frac{\Delta_t\rho_\eta^{d+1}\bigl(1-\rho_\eta^{(M_H(d)+1)(h+1)}\bigr)}{1-\rho_\eta^{h+1}}
  \xrightarrow[H\to\infty]{}
  \tau^{\mathrm{out}}(s_0)\,\rho_\eta^{d}.
\]
Top-$q$ entries cannot exceed $O(\log q)$ hops from $s_0$, so the admissible $K$ under the assumption is far below the $K=q$ significance invoked by Theorem~\ref{thm:cumulative-tracking}. The example deliberately combines exponential diffusion, minimal route multiplicity, and periodicity.

\paragraph{Horizon significance.}
Let $K_q=K_{P_t}(s_r,q)$ be the largest $h$ such that the top-$q$ frontier contains an admissible predecessor route $s_h\to\cdots\to s_0=s_r$. For a near-greedy policy, contractivity makes the field monotone along the route, so every path state enters the top $q$ and $K_q<q$. Define the \emph{horizon-significance} metric
\[
  \mathrm{HS}(K,q)=K_q/q\le 1,
\]
the useful proactive capacity of the field once the reactive residual has covered the $K$-hop ball. The binary tree gives $\mathrm{HS}(K,q)=\mathcal{O}(\log q/q)$.

\paragraph{Volume growth.}
Let $b_x:=\#\{y:\exists a,\ p(x\mid y,a)>0\}$ with global average $B$, and define
\[
  \mathrm{Vol}_h=\#\{s':\operatorname{dist}(s';s_r)\le h\},
  \qquad
  h_q:=\max\{r:\mathrm{Vol}_r<q\}.
\]
Any top-$q$ distribution is contained in a contiguous manifold inside this ball, so a small volume certifies that the frontier can look far ahead:

\begin{table}[h]
\centering
\small
\begin{tabular}{ll}
\toprule
Environment & $\mathrm{Vol}_h$ \\
\midrule
FourRooms & $\Theta(h^2)$ in-room, $\Theta(h)$ across corridors \\
GARNET & approximately $\Theta(B^h)$, lower after collisions and saturation \\
Binary tree & $\Theta(2^h)$ \\
\bottomrule
\end{tabular}
\caption{\footnotesize Volume growth of the backward ball around the reward site. This combinatorial effect, rather than kernel-independent positivity, is the primary source of useful lookahead.}
\label{tab:volume-growth}
\end{table}

\paragraph{Path concentration under near-greediness.}
Signal dissipates through kernel rows that spread mass: $\sum_y[P_t^h]_{x,y}=1$ and $\max_y[P_t^h]_{x,y}\ge1/\mathrm{Vol}_h$. In a near-greedy epoch there exists a route alignment from $s_h$ to $s_r$ with the highest transition-product mass. Assigning every preferred transition probability $1-\varepsilon_{\mathrm{greedy}}$ gives
\[
  [P_t^h]_{s_h,s_r}
  \ge \frac{(1-\varepsilon_{\mathrm{greedy}})^h}{\mathrm{Vol}_h}
  = \frac{1-o(\varepsilon_{\mathrm{greedy}})}{\mathrm{Vol}_h}.
\]

\paragraph{Strict rate for horizon significance.}
With greedy mass $p_\star=1-\varepsilon_{\mathrm{greedy}}$ and non-greedy mass $\tilde p=\varepsilon_{\mathrm{greedy}}/(a_x-1)$, where $a_x$ is the action count at $x$, the strongest off-ball competitor deviates once at depth $h_q+1$, so the critical first-arrival term at depth $h>h_q$ outranks it whenever $(\rho_\eta p_\star)^{h-h_q}>\rho_\eta\tilde p$. For fixed $\varepsilon_{\mathrm{greedy}}$ this certifies
\[
  K_q\ge\min\left\{q,\ h_q+\left\lceil\frac{\log(\rho_\eta\tilde p)}{\log(\rho_\eta p_\star)}\right\rceil-1\right\},
\]
so the fixed-policy horizon significance grows with the volume radius $h_q$. Recovering the full horizon $h=q$ requires, with global average action factor $A$,
\[
  \varepsilon_{\mathrm{greedy}}
  <\frac{A-1}{\rho_\eta}\bigl[\rho_\eta(1-\varepsilon_{\mathrm{greedy}})\bigr]^{q-h_q}.
\]
Ideal significance therefore demands exploration that decays exponentially in $q$; practical effective horizons rest on the combinatorial volume growth of the environment, and the binary tree stays excluded because $a_x=1$ along its first-arrival path. For recurrent kernels the same conclusion is certified by the full-score cutoff
\[
  \Gamma_q(\varepsilon_{\mathrm{greedy}})
  :=\bar\tau_{(q)}-\bar\tau_{(q+1)}
  >2\norm{\tau_{\eta,t}^{\mathrm{out}}(P_t)-\tau_{\eta,t}^{\mathrm{out}}(\hat P_t)}_\infty.
\]
Taken together, these statements separate \emph{availability} of the dense signal from its \emph{useful magnitude}: the gate determines which states can be executed, and the horizon-significance quantities determine how much proactive ordering survives the top-$q$ filter. The top-$q$ frontier is an optional heap-efficiency device, not part of the guarantee; $q=n$ recovers the untruncated rule.

\newpage
\appendixheading{Spectral Prioritized Sweeping: Algorithms and Implementation Details}
\label{app:b}

\appendixsubsection{Algorithmic settings for MDP observability}
\label{app:b-host-intel}

As to aid the integration of GTA-PS onto DP and Dyna, we can not ignore what each of them is built for in the first place.

DP can only be chosen when the full oracle distributional setting is exposed. The host agent's belief cannot exceed that of exact planning model, as every intel: the known transition kernel, the current reward distribution, and the ideal Bellman residual can be traced at ease. The host does not need exploratory action selection onto the MDP to collect data. The one actual committal process -- being Policy Improvement -- serves as the host step.

In the sample-based Dyna setting, because of \ref{par:agent-setting}, the host agent's belief must be empirical: it can only be formed through observed experience, by typically maintaining a learned model plus the current reward estimate. The available information is therefore local and noisy, best managed by one-step TD residuals rather than other more sample consuming ones. The host step is action execution under an exploratory move, as this leads for subsequent learning process.

\noindent\textit{Note.} In Dyna, the \emph{host policy} should be an $\varepsilon$-greedy one, so that new experience continues to update the sample model and not get stuck in stale knowledge.

\appendixsubsection{Computational methods}
\label{app:b-computational-methods}

\appendixsubsubsection{Construction of $L^{out}_t$, $L^{in}_t$.}
\label{app:b-laplacian}
Given per-action matrices $P^a$ and topology policy $\pi^{\mathrm{PS}}_t$, construct $P_t = \sum_a \mathrm{diag}(\pi^{\mathrm{PS}}_t(\cdot,a))P^a$ in $O(|\mathcal{S}|\cdot|\mathcal{A}|\cdot\mathrm{deg})$. Then $L^{out}_t = I - P_t$ and $L^{in}_t = D^{in}_t - P_t^\top$ where $D^{in}_t = \mathrm{diag}(P_t^\top\mathbf{1})$. Both are stored in CSR format.

\appendixsubsubsection{Resolvent computation.}
\label{app:b-sherman-morrison}
$L^{out}_t = I - P_t$ is singular (null vector $\mathbf{1}$). Adding $\eta I$ shifts all eigenvalues by $\eta$, so $\mathrm{Re}(\kappa_k(L^{out}_t + \eta I)) \geq \eta > 0$ for any ergodic chain. The inverse is thus well-defined without any symmetry requirement. The forward and backward potentials are residual diffusions:
\[
  \mathbf{f}^{out}_t = (L^{out}_t + \eta I)^{-1}\xi_t, \qquad
  \mathbf{f}^{in}_t  = (L^{in}_t  + \eta I)^{-1}\xi_t.
\]
For small problems the sparse systems are factored once with \texttt{scipy.sparse.linalg.splu}; repeated resolvent evaluations then reduce to triangular solves through \texttt{lu.solve(rhs)}. The standalone spectral helper switches to the iterative path at $|\mathcal{S}|\le 300$, while the deployed planner module keeps cached LU up to its configured threshold of 750. Both paths serve the tabular FourRooms and small GARNET experiments.

For larger problems, the iterative path is restarted GMRES with an ILU preconditioner \citep{saad1986gmres, meijerink1977iterative}. In SciPy notation the helper uses \texttt{spla.spilu(A.tocsc(), fill\_factor=10, drop\_tol=1e-4)} followed by \texttt{spla.gmres(A, rhs, M=M, restart=20, maxiter=200, atol=1e-4)}, while the deployed module uses \texttt{restart=30}, \texttt{maxiter=500}, and \texttt{atol=1e-4}.
If ILU fails because of a singular pivot, or if GMRES stalls, the implementation falls back to \texttt{spla.spsolve} or a fresh LU factorization. The rank-8 component audit in Appendix~\ref{app:d-post-review} places the measured crossover near 650 states with overlapping intervals, so both thresholds remain on the conservative side of the measured meeting point.

Reward shifts are especially cheap in the LU regime because the operator is unchanged: if $\delta\mathbf{r}_t$ is the new reward perturbation, the update is $\mathbf{f}^{out}_t \leftarrow \mathbf{f}^{out}_t + (L^{out}_t+\eta I)^{-1}\delta\mathbf{r}_t$ and analogously for $\mathbf{f}^{in}_t$. A full column cache $g(\cdot,s)$ would reduce repeated reward-shift queries to $O(|\mathrm{supp}(\delta\mathbf{r})|\cdot|\mathcal{S}|)$, but at $O(|\mathcal{S}|^2)$ memory cost; the reference path instead reuses sparse factors and solves on demand.

\begin{algorithm}[ht]
\caption{\footnotesize Resolvent Solve Path}
\begin{algorithmic}[1]
\State $P_{\mathrm{sp}} \gets \texttt{sp.csr\_array}(P_t)$
\State $A^{out} \gets \texttt{sp.eye}(n,\texttt{format="csr"})\cdot(1+\eta)-P_{\mathrm{sp}}$
\State $A^{in} \gets \texttt{in\_laplacian}(P_t)+\texttt{sp.eye}(n,\texttt{format="csr"})\cdot\eta$
\If{$n \leq \texttt{solver\_threshold}$}
    \State $\texttt{lu\_out} \gets \texttt{spla.splu}(A^{out}\texttt{.tocsc()})$
    \State$\texttt{lu\_in} \gets \texttt{spla.splu}(A^{in}\texttt{.tocsc()})$
    \State $\mathbf{f}^{out}_t \gets \texttt{lu\_out.solve}(\xi_t)$
\EndIf
\If{$n > \texttt{solver\_threshold}$}
    \State $\texttt{ilu} \gets \texttt{spla.spilu}(A^{\mathrm{out}}\texttt{.tocsc()}, \texttt{fill\_factor}=10, \texttt{drop\_tol}=10^{-4})$
    \State $M \gets \texttt{spla.LinearOperator}((n,n), \texttt{ilu.solve})$
    \State $\mathbf{f}^{\mathrm{out}}_t, \texttt{info} \gets \texttt{spla.gmres}(A^{\mathrm{out}}, \xi_t, M, \texttt{restart}=30, \texttt{maxiter}=500, \texttt{atol}=10^{-4})$
    \If{$\texttt{info} \neq 0$}
        \State $\mathbf{f}^{\mathrm{out}}_t \gets \texttt{spla.spsolve}(A^{\mathrm{out}}\texttt{.tocsc()}, \xi_t)$
    \EndIf
\EndIf
\State $\mathbf{f}^{in}_t$ is solved analogously with $A^{in}$
\end{algorithmic}
\end{algorithm}

\appendixsubsubsection{SLEM computation via Arnoldi.}
The SLEM is recomputed from the current topology kernel $P_t$ using the Arnoldi process underlying \texttt{scipy.sparse.linalg.eigs}. Each matrix-vector product costs $O(\mathrm{supp}(P_t))$, so a $k$-vector Krylov run costs $O(k\,\mathrm{supp}(P_t)+k^2|\mathcal{S}|)$. In practice we use $k=15$ and pass the dominant eigenvector from the previous recomputation as a warm start. The non-unit eigenvalue with the largest modulus is taken as the SLEM estimate. Recomputing every $T_{\mathrm{arn}}=\lceil \sqrt{|\mathcal{S}|}\rceil$ policy updates is usually sufficient; between those checkpoints, the scheduler reuses the last estimate.

\begin{algorithm}[ht]
\caption{\footnotesize SLEM Update}
\begin{algorithmic}[1]
\State $P_{\mathrm{sp}} \gets \texttt{sp.csr\_array}(P_t)$;\quad $P_{\mathrm{sp}}\texttt{.eliminate\_zeros()}$
\State $\kappa,\;V \gets \texttt{spla.eigs}(P_{\mathrm{sp}},k=16,\texttt{which="LM"},v_0=v_{\mathrm{prev}})$
\State $m_i \gets |\kappa_i|$ for all returned eigenvalues
\State discard the eigenvalue with $|\kappa_i-1| \leq 10^{-4}$
\State $\kappa_2 \gets \arg\max_i m_i$ over the remaining set
\State $\mathrm{SLEM}(P_t) \gets |\kappa_2|$
\State $v_{\mathrm{prev}} \gets \operatorname{Re}(V[:,0])$ for the next warm start
\State recompute only if $t \bmod T_\mathrm{arn} = 0$
\end{algorithmic}
\end{algorithm}

\appendixsubsubsection{Topology signal assembly.}
Once $\xi_t$ is available, the computational chain is: solve the forward and backward resolvents, mix them with the fixed coefficient $\alpha$, filter the dense signal to a top-$q_t$ frontier, and rekey the residual queue with the aggregate topology score. In the exact-model regime $\xi_t(s)=|(\mathcal{T}V_t)(s)-V_t(s)|$; in sample-based Dyna this is the magnitude of the most recent TD(0) error. The reference implementation materializes the dense frontier with \texttt{np.argpartition} and keeps the active residual and topology-only candidates in separate heaps (\texttt{heapq.heapify}), which avoids sorting all states at every backup.

\begin{algorithm}[ht]
\caption{\footnotesize Topology Assembly}
\begin{algorithmic}[1]
\State $\xi_t(s) \gets |(\mathcal{T}V_t)(s)-V_t(s)|$
\State $\fout_t \gets \texttt{solve\_out}(P_t, \xi_t)$; \quad $\fin_t \gets \texttt{solve\_in}(P_t, \xi_t)$
\State $\boldsymbol{\tau}_t \gets (1-\alpha)|\fout_t|+\alpha|\fin_t|$
\State $\beta_t \gets \beta$ or $\beta_1((1-\mathrm{SLEM}(P_t))+\beta_2)$
\State $I_q \gets \texttt{np.argpartition}(\boldsymbol{\tau}_t, n-q_t)[n-q_t:]$
\State $\texttt{Q}_{\xi} \gets \texttt{heapq.heapify}\{(s,|\xi_t(s)|+\beta_t\tau_t(s)):|\xi_t(s)|>0\}$
\State $\texttt{H}_{q} \gets \texttt{heapq.heapify}\{(s,\beta_t\tau_t(s)):s\in I_q\}$
\end{algorithmic}
\end{algorithm}

\appendixsubsubsection{Two-queue priority with top-$q$ filtering.}
\label{app:b-elitist-two-queue}
Because $\boldsymbol{\tau}_t$ is dense, inserting it directly into the residual heap would crowd the queue. The implementation uses two persistent structures: $\mathcal{Q}_{\xi}$ over the active residual set $\mathcal{I}_t$, keyed by the aggregate score $|\xi_t(s)|+\beta_t\tau_t(s)$, and an auxiliary dense index $\mathcal{Q}_{\tau}$ over all states keyed by $\tau_t(s)$. After each \textsc{RefreshTau}, a filtered frontier $\mathcal{F}_q = \textsc{TopQ}(\mathcal{Q}_{\tau}, q_t)$ is materialized (via argpartition, $O(|\mathcal{S}|)$) and stored as a small heap $\mathcal{H}_q$ for storage and rekeying. Selection is residual-first, $s_t\in\arg\max_{s\in\mathcal{I}_t}\{|\xi_t(s)|+\beta_t\tau_t(s)\}$, and the frontier supplies the exploratory fallback only once $\mathcal{Q}_{\xi}$ is exhausted. Setting $q_t=|\mathcal{S}|$ recovers the unfiltered algorithm.

\appendixsubsubsection{Computation chain.}
\label{app:b-computation-chain}
The safest way to read the deployed module is as a directed acyclic computation graph.

At a fixed host step, the topology-policy state determines the induced topology kernel and its two linear systems:
\[
  \pi_t^{\mathrm{PS}}
  \;\longrightarrow\;
  P_t
  \;\longrightarrow\;
  (L_t^{out}, L_t^{in}).
\]
From that point the computation splits into three downstream branches. The spectral branch is
\[
  P_t
  \;\longrightarrow\;
  \mathrm{SLEM}(P_t)
  \;\longrightarrow\;
  \beta_t.
\]
The dense topology branch is
\[
  (\xi_t, L_t^{out}, L_t^{in})
  \;\longrightarrow\;
  (\mathbf{f}^{out}_t,\mathbf{f}^{in}_t)
  \;\longrightarrow\;
  \boldsymbol{\tau}_t
  \;\longrightarrow\;
  \mathcal{Q}_{\tau}
  \;\longrightarrow\;
  \mathcal{F}_q
  \;\longrightarrow\;
  \mathcal{H}_q.
\]
The residual branch is simply
\[
  \xi_t
  \;\longrightarrow\;
  \mathcal{Q}_{\xi}.
\]
These branches meet only at the selection stage through the score $|\xi_t(s)|+\beta_t\tau_t(s)$ used by \textsc{PopAndPrioritize}, which takes the head of $\mathcal{Q}_{\xi}$ whenever one exists and falls back to the head of $\mathcal{H}_q$ only when $\mathcal{Q}_{\xi}$ is exhausted. In particular, $\beta_t$ does not enter the resolvent solves, $\boldsymbol{\tau}_t$ does not enter Arnoldi, and top-$q_t$ filtering is applied only after the dense topology field has already been formed.

\appendixsubsubsection{Per-step cost summary.}
\label{app:b-cost-summary}
\begin{center}
\renewcommand{\arraystretch}{1.15}
\begin{tabular}{ll}
\toprule
Procedure & Cost \\
\midrule
\textsc{GtaPolicyUpdate} & $O(|\mathcal{A}|)$ per backed-up state \\
\textsc{TopologyUpdate} (per policy step) & $O(|\mathcal{S}|)$ (LU tri-solves + SM) \\
\textsc{RefreshTau} (per shift / lazy) & $O(|\mathcal{S}| + q_t)$ \\
\textsc{RewardUpdate} (per shift) & $O(|\mathcal{S}| + q_t)$ \\
\textsc{PopAndPrioritize} (per backup) & $O(\log|\mathcal{S}| + \log q_t)$ \\
\textsc{PushPredecessors} (per backup) & $O(|\mathrm{Pred}(s)|\log|\mathcal{S}|)$ \\
Arnoldi (every $T_{\mathrm{arn}}$) & $O(j|\mathcal{S}|\cdot\mathrm{deg}+j^2|\mathcal{S}|)$, $j=15$ \\
\bottomrule
\end{tabular}
\end{center}

Dense refreshes (\textsc{RefreshTau}, Arnoldi) are triggered only at reward shifts and lazily every $T_{\tau}=|\mathcal{S}|$ backups; the amortized incremental cost per plan step is $O(|\mathcal{S}|)$ for the LU tri-solves and $O(|\mathcal{A}|\cdot\mathrm{deg})$ for the host Bellman backup.

\appendixsubsection{The unified Prioritized Sweeping interface}
\label{app:b-ps-interface}

We should view any priority-based system as a \emph{database} for the host. The host agent gains complete control over when entries are pushed, queried, refreshed, and removed during planning. The priority module should expose a small set of stateful operations that the host can invoke inside DP or Dyna-style procedures, while preserving the same residual and topology bookkeeping across all variants. This makes the interface reusable across different replanning schedules, reward-update patterns, transition models and stopping rules. We do not specify how this should be implemented programmatically. The GTA-PS + DP/Dyna pseudocode in \ref{app:e-dp-dyna} shows the concrete host-side call sequence.

\appendixsubsection{The SPS (GTA-PS) module}
\label{app:b-algorithm-pseudocode}

\appendixsubsubsection{Residual gating mechanism}
\label{app:b-gating-mechanism}
GTA-PS applies the topology term as an ordering correction over the residual support. At each planning step, define the residual-active set
\[
  \mathcal I_t \triangleq
  \{s\in\mathcal S: |\xi_t(s)|>0\}.
\]
If \(\mathcal I_t\neq\varnothing\), the next backup state is selected only from
\(\mathcal I_t\):
\[
  s_t \in
  \arg\max_{s\in\mathcal I_t}
  \left\{
    |\xi_t(s)|+\beta_t\tau_t(s)
  \right\}.
\]
Otherwise the planner draws an exploratory state from the frontier.

Because the gate keeps every executable backup inside \(\mathcal I_t\), the term
\(\beta_t\tau_t(s)\) can reorder states with positive residual but cannot promote
a topology-only state while residual-active states exist.

This preserves the anti-starvation condition behind the asynchronous contraction proof \citep{bertsekas1989parallel, tsitsiklis1994asynchronous}, so GTA-PS keeps the standard infinite-residual visit condition.

\appendixsubsubsection{Detailed pseudocode}

The module exposes five procedures that any model-based host can call, and maintains:

\begin{enumerate}[label=(\roman*),leftmargin=*,itemsep=0pt,topsep=0.25em]
    \item the Bellman residual array $\xi[s]$ and a residual heap $\mathcal{Q}_{\xi}$ over the active set $\mathcal{I}_t=\{s:\xi[s]>0\}$ \ref{app:b-gating-mechanism};
    \item the topology array $\tau[s]$, a dense heap $\mathcal{Q}_{\tau}$ over all states, and a filtered frontier
    $\mathcal{F}_q=\textsc{TopQ}(\mathcal{Q}_{\tau},q_t)$ as a small heap $\mathcal{H}_q$ \ref{app:b-elitist-two-queue};
    \item topology state $(\mathbf{f}^{out},\mathbf{f}^{in},\mathrm{SLEM},v_2,u_2)$;
    \item the topology policy $\pi^{\mathrm{PS}}_t$ \textup{(Boltzmann over host advantages, temperature~$\lambda$)}.
\end{enumerate}

During planning, the host calls \textsc{PopAndPrioritize} to pick the next backup state and \textsc{PushPredecessors} to push the resulting residual to its predecessors. At each backed-up state it also calls \textsc{GtaPolicyUpdate} for the temperature-softened policy row and \textsc{TopologyUpdate} for the incremental Sherman--Morrison and Arnoldi work. After a reward shift it calls \textsc{RewardUpdate}, which diffuses the new residual through both resolvents and rebuilds the priority queues.

\appendixsubsubsection{Four-state walkthrough}
\label{app:b-walkthrough}

\vspace{-0.5cm}
\begin{table}[H]
\centering
\small
\caption{\footnotesize Forward field after a single shift at $s_4$ ($\gamma=0.9$, $\eta=\eta^\star$, $\alpha=0$ for the displayed column, restart $\varepsilon=0.01$, shift size $\Delta=1$). The field profile matches the discount $\gamma^d$ that reactive PS delivers after $d$ backups.}
\label{tab:walkthrough}
\begin{tabular}{ccccc}
\toprule
state & hop $d$ to $s_4$ & $\xi_t$ (now) & $\tau_t/\tau_t(s_4)$ & wave $\gamma^d$ \\
\midrule
$s_1$ & 3 & 0.00 & 0.72 & 0.73 \\
$s_2$ & 2 & 0.00 & 0.81 & 0.81 \\
$s_3$ & 1 & 0.00 & 0.90 & 0.90 \\
$s_4$ & 0 & 1.00 & 1.00 & 1.00 \\
\bottomrule
\end{tabular}
\end{table}

We propose a toy example for prototyping the method. Running a 4-chain $s_1\to s_2\to s_3\to s_4$ with a restart at $s_1$ at rate $\varepsilon=0.01$ keeps the chain irreducible and aperiodic; take $\gamma=0.9$ and the canonical $\eta^\star=(1-\gamma)/\gamma$. A shift at $s_4$ sets $r_t(s_4)=1$ from a zero estimate, so the residual is sparse: $\xi_t=(0,0,0,1)$. Standard PS can queue only $s_4$; the wave reaches $s_3$ after one backup at $\gamma=0.90$, $s_2$ after two at $\gamma^2=0.81$, and $s_1$ after three.

Under $K=2$ it spends the whole budget on $s_4$ and $s_3$, leaving $s_1,s_2$ unqueued. At the shift instant the forward field already tracks that wave: Table~\ref{tab:walkthrough} reports the normalized profile against $\gamma^{d}$, the geometric-discount reading of Eq.~\eqref{eq:gta-discounted-residual-propagation}. Because the gate admits only $s_4$ and then $s_3$, GTA-PS reorders the same disciplined schedule; the gain is informational, since upstream states carry relative mass from the first instant and co-active ties resolve by downstream propagation. Unnormalized, the field at $s_4$ measures $2.59$, so the ratios carry the ranking while the magnitudes follow the resolvent scale.

\appendixsubsection{Complexity analysis.}
\label{app:b-complexity}

The complexity account assigns one meaning and one event frequency to each symbol. Let $n=|\mathcal{S}|$, let $\mathrm{deg}$ denote the average number of supported successors per state-action pair, and let $p_s$ denote the number of cached predecessor state-action entries of state $s$, so that $\mathbb{E}_s[p_s]=|\mathcal{A}|\,\mathrm{deg}$ under the checked-in predecessor cache.

Define $C_{\mathrm{backup}}^{\mathrm{DP}}(s)=O(|\mathcal{A}|\,\mathrm{deg})$ for the host Bellman backup, $C_{\mathrm{res}}(u)=O(|\mathcal{A}|\,\mathrm{deg})$ for one predecessor residual recomputation, $C_{\mathrm{row}}(s)$ for the row-local policy and topology update attached to the backed state (cached triangular solves plus the Sherman--Morrison correction), and $C_{\mathrm{pot}}$ for one pair of cached resolvent applications. One literal backed-state operation then satisfies the per-backup upper bound
\[
\begin{aligned}
  C_{\mathrm{local}}^{\mathrm{GTA}}(s)
  ={}& C_{\mathrm{backup}}^{\mathrm{DP}}(s)
  + O(\log n+\log q_t) \\
  &+ \sum_{u\in\mathrm{Pred}(s)}\bigl(C_{\mathrm{res}}(u)+O(\log n)\bigr)
  + C_{\mathrm{row}}(s).
\end{aligned}
\]

The logarithmic terms count heap operations only: priority comparisons read stored scalar priorities, so Bellman and residual computations are never multiplied by heap depth. If residual computation carries the same sparse Bellman cost at every predecessor, the bound simplifies to
\[
  C_{\mathrm{local}}^{\mathrm{GTA}}(s)
  = O\!\left((p_s+1)|\mathcal{A}|\,\mathrm{deg}
  +(p_s+1)\log n+\log q_t+C_{\mathrm{row}}(s)\right).
\]
Cancelling the terms that Standard-PS also pays, namely $C_{\mathrm{backup}}^{\mathrm{DP}}$ and the predecessor residual sums, the incremental GTA-PS overhead per backed state is
\[
  O\!\left((p_s+1)\log n+\log q_t\right)+C_{\mathrm{row}}(s).
\]
The queue logarithms are multiplied by the number of heap operations, that is, by one push per predecessor plus the pop itself; they stay additive relative to Bellman and residual work. The average predecessor count matches the declared branching convention: a simple state graph with average successor count $b_{\mathrm{state}}$ gives $\mathbb{E}_s[p_s]=b_{\mathrm{state}}$, while the checked-in state-action cache gives $\mathbb{E}_s[p_s]=|\mathcal{A}|\,\mathrm{deg}$.

Periodic work is charged separately from the local bound. Let $C_{\mathrm{ref}}$ be one \textsc{RefreshTau} call, $C_{\mathrm{topo}}$ one dense topology-operator rebuild, $C_{\mathrm{arn}}$ one Arnoldi recomputation with Krylov depth $j$, and $C_{\mathrm{shift}}$ one reward-shift refresh. In the reference implementation,
\[
  C_{\mathrm{ref}} = O\!\left(C_{\mathrm{pot}}+n+q_t+|\mathcal{I}_t|\right)
\]
with cached sparse LU triangular solves below the configured module threshold (750 in the deployed planner) and restarted GMRES above, so $C_{\mathrm{pot}}$ is $O(n)$ in the LU regime and iteration-dependent otherwise; the measured rank-8 LU-to-GMRES crossover appears in Appendix~\ref{app:d-post-review}. With measured event periods $T_\tau,T_{\mathrm{topo}},T_{\mathrm{arn}}$ and reward-shift period $T_{\mathrm{shift}}$, the amortized per-backup bound is
\[
\bar C_{\mathrm{GTA}}(s) = C_{\mathrm{local}}^{\mathrm{GTA}}(s)
 + \frac{C_{\mathrm{ref}}}{T_\tau} + \frac{C_{\mathrm{topo}}}{T_{\mathrm{topo}}} + \frac{C_{\mathrm{arn}}}{T_{\mathrm{arn}}} + \frac{C_{\mathrm{shift}}}{T_{\mathrm{shift}}}.
\]
For a Dyna interaction with $K$ backed states $s_1,\dots,s_K$,
\[
  C_{\mathrm{Dyna}} = C_{\mathrm{real}} + \sum_{k=1}^{K} C_{\mathrm{local}}^{\mathrm{GTA}}(s_k) + C_{\mathrm{periodic}},
\]
where $C_{\mathrm{periodic}}$ collects the refresh, rebuild, Arnoldi, and shift terms advanced by the counters that the real and simulated steps actually move. Per-backup, per-policy-update, per-shift, and per-interaction quantities therefore stay in separate expressions, and the incremental GTA overhead excludes every cost shared with Standard-PS.
The checked-in implementation uses $T_{\mathrm{arn}}=\lceil\sqrt{n}\rceil$ and $T_{\mathrm{topo}}=T_\tau=n$. These cadences are implementation heuristics, not asymptotic optima. The $\sqrt{n}$ scaling balances the amortized $O(n/T_{\mathrm{arn}})$ refresh cost against the $O(T_{\mathrm{arn}})$ staleness of the estimate; constant offsets are implementation-specific, and cadence ablations favor a few multiples of $\sqrt{n}$ in practice.

\begin{algorithm}[!ht]
\caption{\footnotesize Spectral Prioritized Sweeping (SPS / GTA-PS) Module}
\small
\begin{algorithmic}[1]
\Statex \textbf{Precompute} (once, $O(|\mathcal{S}|^3)$)
\State $P^{\mathrm{PS}}_0 \gets \sum_a \mathrm{diag}(\pi^{\mathrm{PS}}_0(\cdot,a))\,P^a$
\State $L^{out}_0 \gets I - P^{\mathrm{PS}}_0$;\quad $L^{in}_0 \gets D^{in}_0 - (P^{\mathrm{PS}}_0)^\top$
\State $(L^{out}_0+\eta I)^{-1}$ via LU \Comment{cached for all subsequent tri-solves}
\State $(v_2, u_2, \mathrm{SLEM}) \gets \textsc{Arnoldi}(P^{\mathrm{PS}}_0, (P^{\mathrm{PS}}_0)^\top)$, $j=15$
\State $\beta_0 \gets \beta_1\cdot(1-\mathrm{SLEM}+\beta_2)$;\quad choose $q_0 \in [K,|\mathcal{S}|]$
\State \textbf{for each} $s \in \mathcal{S}$: $\xi[s] \gets |(\mathcal{T}V_0)(s)-V_0(s)|$
\State \textsc{RefreshTau}() \Comment{initial $\boldsymbol{\tau}$ from $\xi$; builds all queues}

\Statex
\Procedure{GtaPolicyUpdate}{$s, A^{\mathrm{PS}}(s,\cdot)$} \Comment{$O(|\mathcal{A}|)$}
  \State $\pi^{\mathrm{PS}}(s,\cdot) \gets \mathrm{softmax}(A^{\mathrm{PS}}(s,\cdot)/\lambda)$
\EndProcedure

\Statex
\Procedure{TopologyUpdate}{$s_{\mathrm{upd}}, \delta\pi^{\mathrm{PS}}$}
  \State $w \gets \sum_a \delta\pi^{\mathrm{PS}}(a)\,p(\cdot|s_{\mathrm{upd}},a)$;\quad $\delta P^{\mathrm{PS}} \gets e_{s_{\mathrm{upd}}}\,w^\top$ \Comment{rank-1, $O(\mathrm{deg})$}
  \State $u \gets (L^{out}_t+\eta I)^{-1} w$;\quad $v \gets (L^{out}_t+\eta I)^{-\top} e_{s_{\mathrm{upd}}}$ \Comment{tri-solves: $O(|\mathcal{S}|)$}
  \State $\mathbf{f}^{out} \mathrel{+}= \frac{v^\top\mathbf{f}^{out}}{1 - v^\top w} u$ \Comment{SM exact update, $O(|\mathcal{S}|)$}
  \State $\delta\kappa_2 \gets u_2(s_{\mathrm{upd}})(w^\top v_2)/(u_2^\top v_2)$;\quad $\mathrm{SLEM} \mathrel{+}= \operatorname{Re}(\bar\kappa_2/|\kappa_2|\cdot\delta\kappa_2)$ \Comment{$O(\mathrm{deg})$}
  \If{$t \bmod T_{\mathrm{arn}} = 0$} \Comment{$T_{\mathrm{arn}} = \lceil|\mathcal{S}|^{1/2}\rceil$}
    \State $(v_2,u_2,\mathrm{SLEM}) \gets \textsc{Arnoldi}(P^{\mathrm{PS}}_t, (P^{\mathrm{PS}}_t)^\top)$
    \State $\beta_t \gets \beta_1(1-\mathrm{SLEM}+\beta_2)$
    \State $\mathbf{f}^{in} \gets (L^{in}_t+\eta I)^{-1}\xi$ \Comment{batched rebuild}
    \State \textsc{RefreshTau}()
  \EndIf
\EndProcedure

\Statex
\Procedure{RefreshTau}{} \Comment{$O(|\mathcal{S}|)$ two tri-solves; $O(|\mathcal{S}|+q_t)$ queue rebuild}
  \State $\fout \gets (L^{out}_t+\eta I)^{-1}|\xi|$;\quad $\fin \gets (L^{in}_t+\eta I)^{-1}|\xi|$
  \State \textbf{for each} $s \in \mathcal{S}$: $\tau[s] \gets (1-\alpha)|\fout[s]| + \alpha\,|\fin[s]|$
  \State Rebuild $\mathcal{Q}_{\tau}$ from $\{(s,\tau[s])\}_{s\in\mathcal{S}}$
  \State $\mathcal{F}_q \gets \textsc{TopQ}(\mathcal{Q}_{\tau}, q_t)$; build $\mathcal{H}_q$ from $\{(s,\xi[s]+\beta_t\tau[s])\}_{s\in\mathcal{F}_q}$
  \State Rebuild $\mathcal{Q}_{\xi}$ from $\{(s,\xi[s]+\beta_t\tau[s])\}_{s:\xi[s]>0}$
\EndProcedure

\Statex
\Procedure{RewardUpdate}{$\delta\mathbf{r}$} \Comment{called after each reward shift}
  \State $\mathbf{r} \mathrel{+}= \delta\mathbf{r}$;\quad $\mathbf{f}^{out} \mathrel{+}= (L^{out}_t+\eta I)^{-1}\delta\mathbf{r}$;\quad $\mathbf{f}^{in} \mathrel{+}= (L^{in}_t+\eta I)^{-1}\delta\mathbf{r}$
  \State \textbf{for each} $s \in \mathcal{S}$: $\xi[s] \gets |(\mathcal{T}V_t)(s)-V_t(s)|$
  \State \textsc{RefreshTau}()
\EndProcedure

\Statex
\Procedure{PopAndPrioritize}{} \Comment{$O(\log|\mathcal{S}|)$}
  \State Let $s_{\xi}$ be the head of $\mathcal{Q}_{\xi}$; if $\mathcal{Q}_{\xi}$ is empty, let $s_{\xi}$ be the head of $\mathcal{H}_q$ \Comment{exploratory fallback}
  \State \textbf{return} $s_{\xi}$, removing it from all queues
\EndProcedure

\Statex
\Procedure{PushPredecessors}{$s, \Delta V$} \Comment{$O(|\mathrm{Pred}(s)|\log|\mathcal{S}|)$}
  \For{$s' \in \mathrm{Pred}(s)$}
    \State $\xi[s'] \gets |(\mathcal{T}V)(s')-V(s')|$;\quad $\mathcal{Q}_{\xi}.\textsc{Update}(s', \xi[s']+\beta_t\tau[s'])$
  \EndFor
  \If{$k \bmod T_{\tau} = 0$} \textsc{RefreshTau}() \EndIf \Comment{$T_{\tau}=|\mathcal{S}|$}
\EndProcedure
\end{algorithmic}
\end{algorithm}

\newpage
\appendixheading{Intuitions and Design Decisions}
\label{app:c}

\appendixsubsection{Key decisions and their rationale}
\label{app:c-design-decisions}

\paragraph{Why nonstationary reward is the chosen setting}
\label{par:why-nonstationary-reward}

With stationary reward, standard PS runs to convergence. At convergence $\xi_t \to \mathbf{0}$, so $\mathbf{f}^{out}_t = (L^{out}_t+\eta I)^{-1}\xi_t \to \mathbf{0}$ and the topology signal vanishes identically (asymptotic-zero property). For deterministic shortest-path problems, Dijkstra's algorithm computes the optimal policy in $O(|\mathcal{S}|\log|\mathcal{S}|+|E|)$ -- strictly superior. GTA-PS can excel in both regimes, but the extra cost is not justified when standard queues already suffice.

Nonstationary reward with bottlenecks is the setting where GTA-PS dominates. When $r_t$ shifts at frequency $1/T$, the planner has $T$ steps before the next shift. A full PS sweep costs $O(|\mathcal{S}|^2|\mathcal{A}|)$; under budget $T \ll |\mathcal{S}|^2|\mathcal{A}|$, the Bellman wavefront has not reached all states and the reactive residual signal is zero at predecessor-blocked states. GTA-PS's topology signal is nonzero at all states immediately after a reward shift (\propref{prop:dense-support}), so it can assign nonzero topology scores to predecessor-blocked states before the wavefront arrives. A sub-$O(|\mathcal{S}|^2)$ update multiplier keeps this regime economical.

\paragraph{Why the resolvent, among the diffuser family?}
\label{par:why-resolvent}
There are numerous choices for a diffuser operator, such as $k$-truncated smoothing $(I - \eta L)^k$, heat diffusion $\exp(-\tau L)$, the Laplacian pseudoinverse, and the regularized resolvent. GTA-PS uses the resolvent because it is both controllable and economical.
\begin{itemize}
    \item Truncated smoothing has a hard finite horizon, heat diffusion is costly to update incrementally, and the pseudoinverse lacks explicit horizon control while depending on the Laplacian nullspace. In the regularized resolvent \((L+\eta I)^{-1}\), \(\eta\) acts as a geometric-discounting factor.
    \item A concise, exact localized update also exists (\ref{app:b-computational-methods}), which we expect heat diffusion or pseudoinverse constructions to lack.
\end{itemize}

\paragraph{Directed versus symmetrized Laplacians.}
A symmetrized construction, such as $(L+L^\top)/2$ or an additive reversiblization of $P_t$, would hide the source-sink asymmetry that a localized reward shift creates, which is exactly the structure the scheduler should see. An earlier version began from symmetrized-$P$ material and separated the two graph directions after observing this subtlety; the undirected theory \citep{chung1997spectral,doyle1984random} remains background for the spectral language, not the deployed operator.

\paragraph{Why low-rank update assumption?}
A topology-policy change at state $s_{\mathrm{upd}}$ induces a row-local perturbation $\delta P_t = e_{s_{\mathrm{upd}}} w^\top$. Low-rank resolvent identities are therefore natural here: they match a learning process that is local rather than global. This does not restrict the reward nonstationarity, because multiple localized shifts can be batched into one vector shift per host step. A simultaneous rank-$k$ change admits a Woodbury extension of the same identity: the update costs multiple factor solves plus a $k\times k$ correction, so the cost grows with $k$ and stays outside the measured rank-one maintenance regime.

\paragraph{Relative to competing topology-aware works.}
This design also separates GTA-PS from adjacent methods. Relative to standard PS and VaST~\citep{corneil2018vast}, GTA-PS is proactive: the resolvent field is dense immediately after a reward shift, so distant states receive nonzero topology scores before the Bellman wavefront reaches them. Relative to TVI/FTVI~\citep{dai2011topological}, GTA-PS uses a soft, interruptible topology signal rather than an SCC batch schedule; it can be stopped after any backup and re-used after a reward shift without recomputing a decomposition. Relative to the successor representation~\citep{dayan1993improving}, reward updates remain cheap while the topology operator is applied through sparse linear solves without storing a full dense successor matrix.

\paragraph{Two-queue architecture with top-$q$ frontier.}
Because the mixed field $\boldsymbol{\tau}_t$ is dense (nonzero everywhere when $\xi_t \neq \mathbf{0}$), inserting it directly into the residual heap would crowd the queue with low-priority states, adding an excessive $O(|\mathcal{S}|)$ cost multiplier. The capacity is therefore limited by $q$. Additionally, the implementation uses two heaps: $\mathcal{Q}_{\xi}$ over the active residual set and $\mathcal{H}_q$ over a filtered top-$q$ frontier extracted from the dense $\mathcal{Q}_{\tau}$. This \emph{decouples the two signals}: the residual heap guarantees that states with large Bellman error remain eligible at all times, while the topology frontier keeps the proactive signal visible for rekeying and serves as the exploratory fallback without flooding the heap. Setting $q_t = |\mathcal{S}|$ recovers the unfiltered version analyzed in the theory.

\paragraph{Budget priority: residual-backups first.}
In \textsc{PopAndPrioritize}, residual-reducing backups are mandatory for convergence, while topology-only backups (states with $\xi=0$ but $\tau>0$) are exploratory. Under tight budget, the implementation ensures that $\mathcal{H}_q$ never starves $\mathcal{Q}_{\xi}$. This keeps the topology signal from spending budget on no-op backups at converged states.

\paragraph{LU factorization threshold.}
For small state spaces, sparse LU factorization provides exact triangular solves at $O(|\mathcal{S}|)$ each and is the default path in the tabular experiments. Deployment fixes the threshold explicitly: \texttt{\_LU\_THRESHOLD = 750}, with a conservative switch at 300 inside the standalone spectral helper. Above their thresholds the routine switches to GMRES with a sparse-direct fallback. The component audit in Appendix~\ref{app:d-post-review} measures the rank-8 crossover near 650 states with overlapping intervals, so both thresholds sit on the conservative side of the measured meeting point.

\appendixsubsection{Failed approaches}
\label{app:c-failed-approaches}

\paragraph{Column cache abandoned.}
An earlier design maintained an explicit $O(|\mathcal{S}|^2)$ column cache $\mathbf{g}_s = G^{out}_{\eta,t}(\cdot, s)$ (the $s$-column of the Green's operator) to accelerate incremental updates. It was abandoned in favour of the Sherman--Morrison dual tri-solve, which matches its update expressiveness at $O(|\mathcal{S}|)$ per step with no persistent cache and no per-policy column maintenance.

\paragraph{Stale-column incremental strategy.}
An intermediate design proposed maintaining stale approximate columns with periodic refresh, balancing $O(|\mathcal{S}|)$ local cost against $O(|\mathcal{S}|^2)$ full recomputation. This became unnecessary once Sherman--Morrison provided exact updates at the same $O(|\mathcal{S}|)$ cost, with no refresh lag to accumulate.

\paragraph{Full recomputation of in-field per policy step.}
The in-field perturbation $\delta L^{in}_t = \mathrm{diag}(w) - w\,e_{s_{\mathrm{upd}}}^\top$ is not rank-1, so Sherman--Morrison does not apply. An early implementation attempted to track $\mathbf{f}^{in}$ incrementally through a generalised rank-$(d+1)$ update; the formula's fill-in made it costlier than periodic batch recomputation. The settled design rebuilds $\mathbf{f}^{in}$ in batch at Arnoldi sync points, accepting $O(|\mathcal{S}|)$ periodic cost amortised over $T_{\mathrm{arn}}$ steps.

\paragraph{Forced Arnoldi on every reward shift.}
An early version triggered a full SLEM recomputation (Arnoldi) at every reward shift. This was wasteful: the transition matrix $P_t$ does not change at a reward shift (only $\mathbf{r}_t$ changes), so $\mathrm{SLEM}(P_t)$ and its eigendecomposition remain valid. The current design lets Arnoldi fire on its natural cadence $T_{\mathrm{arn}}$, independent of reward shifts.

\paragraph{Forward-only ablation is not uniformly best.}
The repository keeps a forward-only comparison by setting $\alpha=0$ in the asymmetry sweeps (RQ4, RQ6), as an ablation rather than the deployed default. In directed environments, relying only on $\fout_t$ can miss upstream source--sink structure that appears once $\fin_t$ is mixed in, so the code tunes a fixed $\alpha$ for the full variants instead of collapsing to the forward-only case.

\paragraph{LU fill-in at large $|\mathcal{S}|$.}
For very large state spaces, sparse LU factorization fill-in grows beyond memory or incurs prohibitive one-time cost. The GMRES fallback is present precisely for this regime. Checked-in benchmark scripts probe this overhead on FourRooms and on GARNET instances up to $|\mathcal{S}|=200$, and the crossover itself is measured in Appendix~\ref{app:d-post-review}.

\newpage
\appendixheading{Environments, Hyperparameters, and Experimental Protocol}
\label{app:d}

\appendixsubsection{Environment setup}
\label{app:d-environment-setup}

To isolate each term, we restrict the experiments to two environments that keep full control over the MDP characteristics.
\begin{description}
    \item[Minigrid Fourrooms \citep{chevalier2023minigrid}] \texttt{FourRoomsTabular} implements the environment as a grid navigation field with four rooms separated by middle walls and joined through narrow one-cell corridors, with default action set size $|\mathcal{A}|=4$. Under the default geometry the state space varies from $20$ to $120$ passable cells. Checked-in configs cover $k\in\{1,4\}$ active-room settings for most scripts and a separate $k=2$ for fine-grained time-series observations. Optional asymmetry controls come from the \texttt{drift} and \texttt{one\_way\_prob} arguments.

    \item[GARNET~\citep{archibald1995generation}.] \texttt{GarnetEnv} implements the synthetic benchmark. Checked-in configs use $(|\mathcal{S}|,|\mathcal{A}|)=(75,4)$ and $(200,5)$, with branching factor $3$, the neighbor-rate \texttt{n\_nbr\_rate} controlling locality, and optional \texttt{reversible}, \texttt{bipartite}, and \texttt{reward\_locality} switches. This is the main environment for controlled sweeps over mixing rate, reversibility, reward support, and solver overhead.
\end{description}

Both classes expose the attributes swept later: mixing rate, reward span, reversibility, branching factor, and planning budget.

\appendixsubsection{Experiment Configurations}
\label{app:d-experiment-config}

All reported curves aggregate independent random seeds. Sweeps run via the main orchestrator with at least $1000$ seeds and $\geq10$ logical processors. Confidence bands are $95\%$ Student-$t$ intervals over the seed means.

Hyperparameter tuning uses the standalone Optuna driver \citep{akiba2019optuna}. Each tuning run uses $200$ trials with \texttt{seed\_offset}=100000, keeping tuning seeds disjoint from the evaluation seeds and avoiding selection bias.

Linear-solver settings stay fixed across experiments: exact sparse LU up to the module threshold of 750 states, otherwise GMRES with \texttt{restart}$=30$, \texttt{maxiter}$=500$, and absolute tolerance \texttt{atol}=$10^{-4}$, falling back to \texttt{spsolve} on nonzero status. A rank-8 component audit in Appendix~\ref{app:d-post-review} places the crossover near 650 states, so the threshold stays conservative.

For the Dyna host we keep an exploration rate of $\epsilon=0.2$ to escape stale local loops: with probability $1-\epsilon$ it samples the current greedy action, and with probability $\epsilon$ a random admissible action.

\appendixsubsection{Reward protocol}
\label{app:d-reward-protocol}

All environments use state-based rewards, but the patterns differ by design across settings.

\begin{description}
    \item[FourRooms.] Rewards are not single-cell goals in the checked-in environment. Instead, \texttt{FourRoomsTabular} samples a Gaussian-mixture reward field: it chooses $k$ active rooms, places \texttt{n\_peaks\_per\_room} Gaussian bumps inside each selected room, samples bump heights from the room-specific \texttt{room\_peak\_ranges}, and normalizes the final vector to unit maximum. A reward shift simply resamples this field.

    \item[GARNET.] By default the generator draws sparse rewards over \texttt{reward\_support} states, with amplitudes uniform in $[0.5,1.5]$. When \texttt{reward\_locality=True}, those rewarded states are drawn as a BFS-local connected cluster from the union graph of the transition tensor rather than as independent global samples.
\end{description}

\appendixsubsection{Hyperparameters Tuning and Selection}
\label{app:d-hparam}

We tune the module with a standalone Optuna driver \citep{akiba2019optuna}. Each search minimizes the mean $L_\infty$ value error to the exact DP solution over $30$ tuning seeds, $4$ reward shifts, $K=20$ planning backups, and $50$ host steps per shift. Search intervals and selected values appear in the supplement code.

\appendixsubsection{Baselines and fair-comparison controls}
\label{app:d-baselines}

The checked-in experiment harness instantiates three method variants:
\begin{itemize}
  \item \textbf{\texttt{standard\_ps}:} the Moore-style prioritized-sweeping baseline \citep{moore1993prioritized}, using Bellman residual magnitude as the queue key.
  \item \textbf{\texttt{gta\_base}:} GTA-PS with fixed $\alpha$ and fixed $\beta$.
  \item \textbf{\texttt{gta\_beta}:} GTA-PS with fixed $\alpha$ and adaptive $\beta_t$.
\end{itemize}

All three variants are implemented through the unified priority-based planning
interface \ref{app:b-ps-interface}. This lets us run them under the exact same host setting, MDP instance,
backup budget, and evaluation protocol, with no method-specific entanglement.
The differences can only happen in their inner states themselves.

\appendixsubsection{Why Standard-PS is the primary baseline}
\label{app:d-why-always-ps}

Standard-PS is the primary baseline because it matches our assumption set and intervention scope without extra confounds. We treat the neighboring methods discussed in the paper, TVI~\citep{dai2011topological}, VaST~\citep{corneil2018vast}, and Dyna-2~\citep{silver2008dyna2}, as context, reserving matched comparison for Standard-PS and GPS-H2.

Under the communicating-MDP assumption, TVI reduces to a Standard-PS update schedule. VaST adds supervised representation learning that is orthogonal to our fixed-MDP priority design. Dyna-2 concerns host-agent architecture and model-based decision organization, whereas our study manages sequential value backups inside a given planning interface.

Generalized prioritized solvers \citep{wingate2005prioritization} are the closest candidate, since their priority score already combines value and residual. In nonstationary settings, stale value entries can mislead that queue, which is the mechanism the post-review comparison tests; the GPS-H2 audit is reported in Appendix~\ref{app:d-post-review}.

\appendixsubsection{Computational details and optimizations}
\label{app:d-computational-details}

We implement several nontrivial optimizations that lower the wall-clock cost of GTA-PS:

\paragraph{Sparse representation.} The topology module stores the policy-induced systems in SciPy sparse form. In \texttt{GtaPS}, $P_\pi$ is converted to CSR, near-zero entries below $10^{-6}$ are dropped, and the forward system is assembled as $(1+\eta)I-P_\pi$. The in-field system uses \texttt{in\_laplacian(P\_pi)} with an additional diagonal stabilizer of $\eta+10^{-6}$ in the initialization and lazy-rebuild path.

\paragraph{Solver selection in the deployed module.} The experiment implementation uses the threshold \texttt{\_LU\_THRESHOLD = 750}. Below the threshold, both forward and backward systems are factored with \texttt{scipy.sparse.linalg.splu} and then reused for all subsequent solves. Above it, the module calls \texttt{gmres} with \texttt{restart=30}, \texttt{maxiter=500}, and \texttt{atol=1e-4}, with a fallback to \texttt{spsolve} if GMRES reports nonzero status. The separate helper functions in \texttt{utils/spectral.py} use a conservative 300-state switch with restarted GMRES and an ILU preconditioner for the resolvent routines, together with the ARPACK-based SLEM estimator used by the topology layer.

\paragraph{Queue and predecessor maintenance.} The module rebuilds the dense top-$q$ frontier with \texttt{np.argpartition}, uses Python \texttt{heapq} heaps for both residual and topology queues, and filters predecessor edges at threshold $10^{-2}$ when building the cached predecessor sets. This threshold is used both when the oracle graph is first cached and when empirical model rows are updated online.

\begin{table}[!t]
\centering
\small
\caption{\footnotesize Empirical wall-clock planning cost, reported as median milliseconds over $120$ repetitions. GTA-PS incurs a moderate but stable runtime multiplier from topology refreshes and dual-queue maintenance.}
\label{tab:d7-wallclock}
\begin{tabular}{lcccccc}
\toprule
Environment & $|\mathcal{S}|$ & $K$ & Standard-PS & GTA-PS & GTA-PS$^\beta$ & Overhead \\
\midrule
FourRooms & 40  & 10 & 4.10 & 9.73  & 9.31  & $2.37\times / 2.27\times$ \\
GARNET    & 75  & 18 & 18.79 & 48.41 & 48.85 & $2.58\times / 2.60\times$ \\
GARNET    & 200 & 50 & 51.20 & 112.39 & 114.00 & $2.20\times / 2.23\times$ \\
\bottomrule
\end{tabular}
\end{table}

\appendixsubsection{Additional experiment results.}
\label{app:d-additional-results}

\begin{figure}[htbp]
\centering
\includegraphics[width=0.65\linewidth]{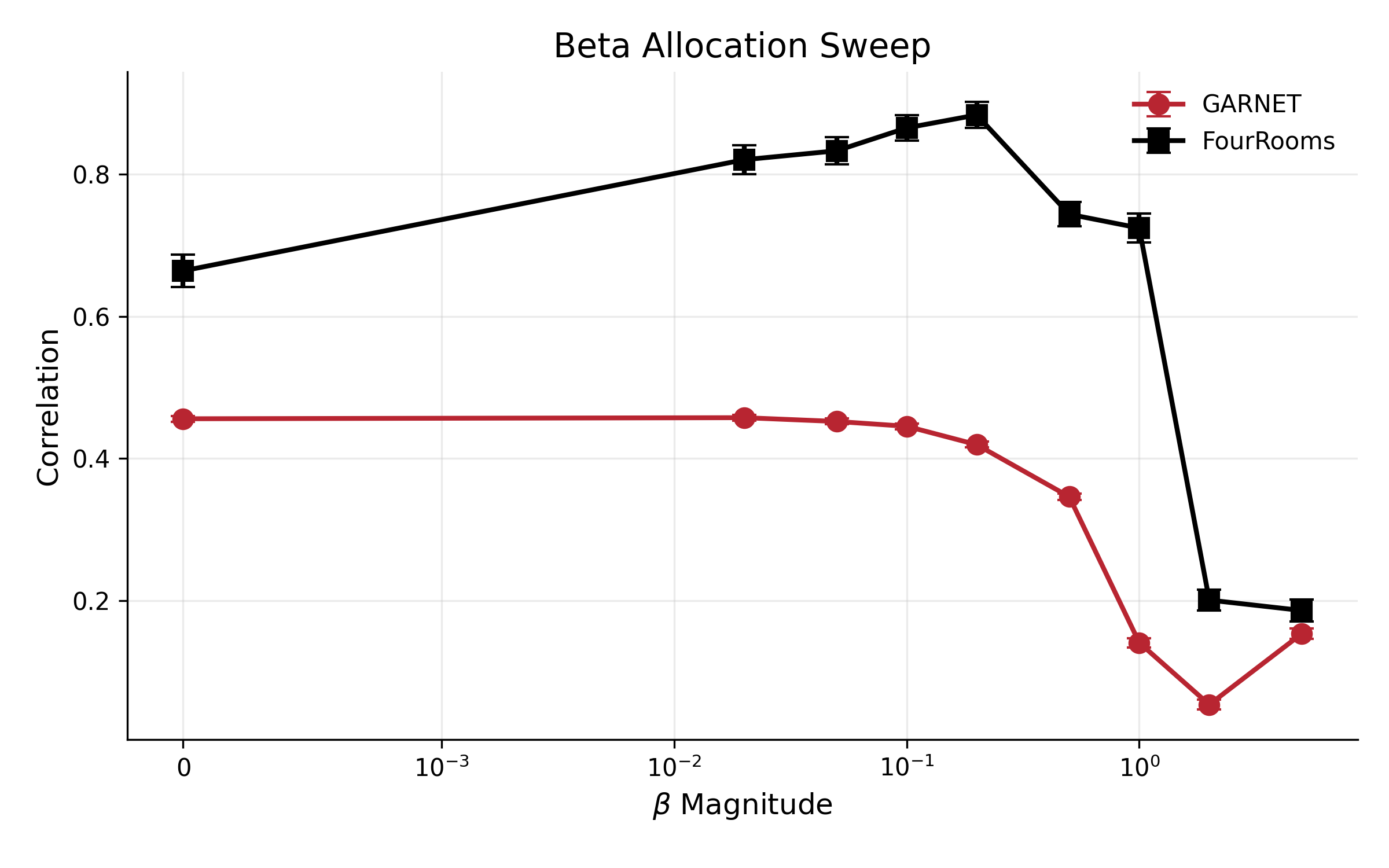}
\caption{
    \footnotesize
    Figure~\ref{fig:d-rq2-beta} studies a single reward-shift event from a zero-initialized reward field in two environments: GARNET with $|\mathcal{S}|=75$ and FourRooms with $|\mathcal{S}|=40$ when varying over calibration weight $\beta$. Vertical axis shows Spearman correlation $\rho$ between initial topology ranking and the realized pop order during the first $K$ backups. Larger values indicate better priority alignment. Moderate $\beta$ gives the strongest agreement between the topology field and the realized backup order, peaking at $\beta=0.02$ in GARNET and $\beta=0.1$ in FourRooms. Larger topology gain sharply degrades the ordering quality, which complements the smallness condition in \propref{prop:gtaps-infinite-visit}.}
\label{fig:d-rq2-beta}
\end{figure}

\begin{figure}[htbp]
\centering
\includegraphics[width=0.65\linewidth]{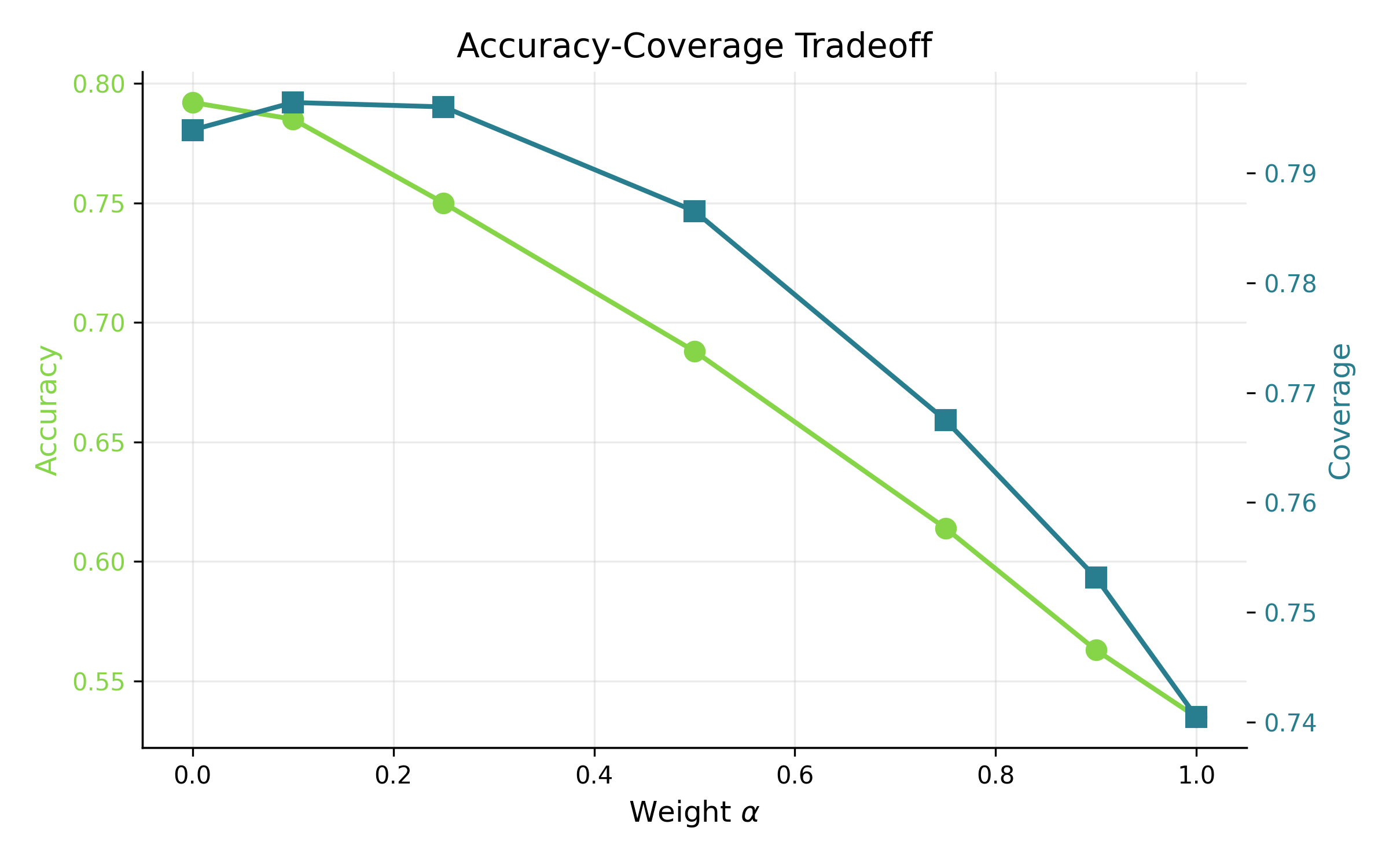}
\caption{
    \footnotesize
    Figure~\ref{fig:d-rq3-alpha} is a GARNET design that calibrates over fixed $\alpha = \alpha_t$.
    The left vertical axis reports top-$1$ accuracy, namely whether the state ranked first by the topology score matches the oracle state ranked first by the greedy propagation surrogate. The right vertical axis reports how much of the oracle's high-priority predecessor set is recovered by the top-ranked topology states. The results favor small $\alpha$, keeping the forward (out-Laplacian) signal dominant for proactive propagation.}
\label{fig:d-rq3-alpha}
\end{figure}

\begin{figure}[htbp]
\centering
\includegraphics[width=0.85\linewidth]{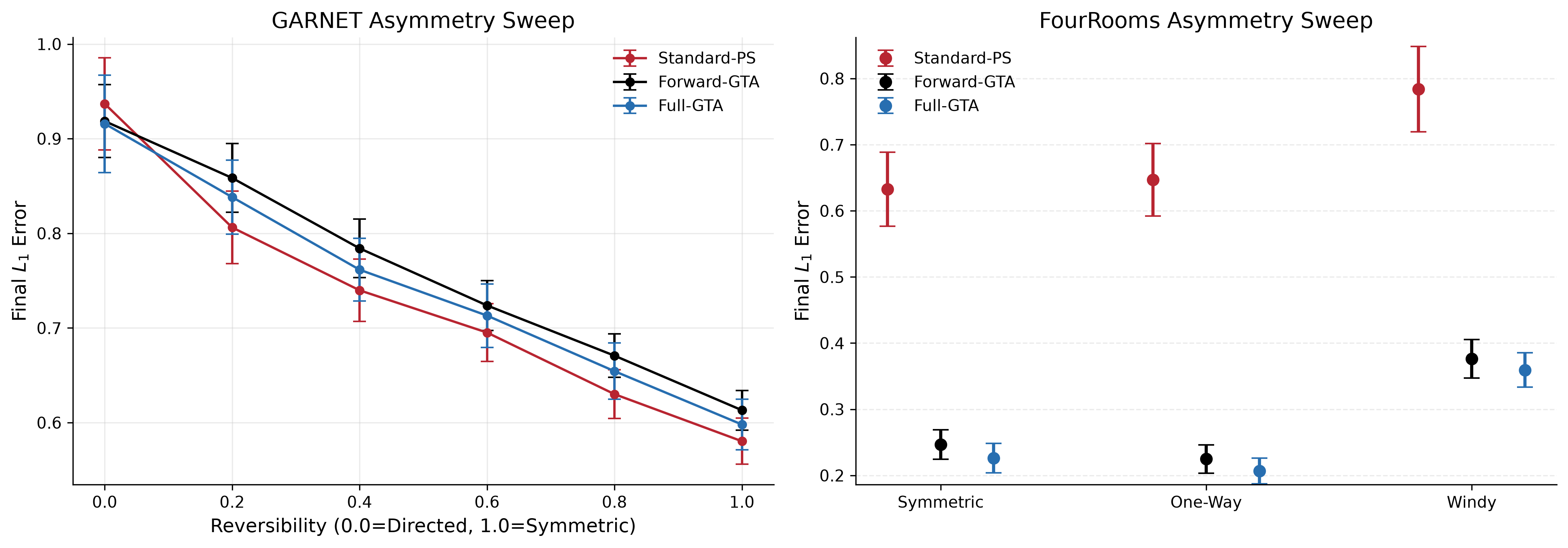}
\caption{
    \footnotesize
    Figure~\ref{fig:d-rq4-asymmetry} reports shift-end $L_1$ value error after repeated replanning under a fixed budget.
    In the left panel, the GARNET horizontal axis is the reversibility coefficient, where $0$ is maximally directed and $1$ is exact symmetry (reversible chain); lower values correspond to more strongly directional random dynamics. In the right panel, FourRooms is perturbed by three transport modes: symmetric dynamics, a one-way corridor bias, and a windy drift field. GTA-PS remains competitive across directed, one-way, and drift-heavy transports; the benefit is not monotone in asymmetry, and the drift-heavy panel retains a clear margin over Standard-PS.}
\label{fig:d-rq4-asymmetry}
\end{figure}

\begin{figure}[htbp]
\centering
\includegraphics[width=0.85\linewidth]{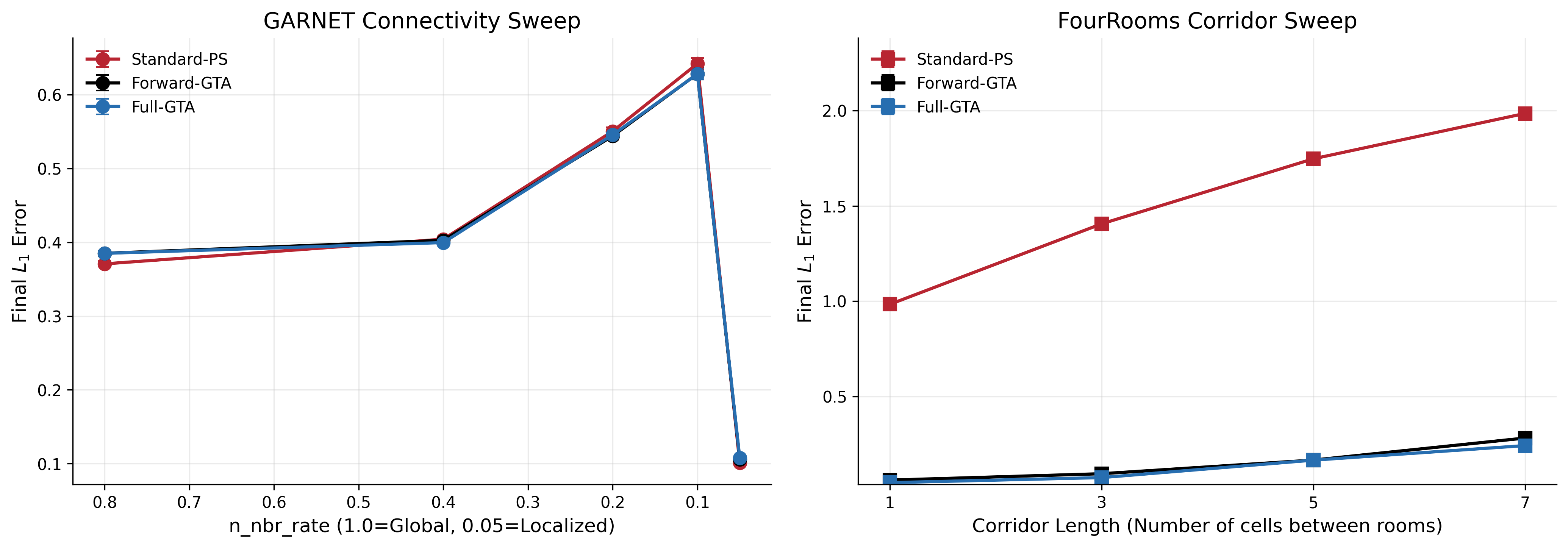}
    \caption{
    \footnotesize
    Figure~\ref{fig:d-rq5-bottlenecks} also measures shift-end $L_1$ value error under the same repeated-shift budgeted replanning protocol. In the left GARNET panel, the horizontal axis is the neighbor-rate parameter \texttt{n\_nbr\_rate}, which controls how globally connected the random transdition graph is; moving left-to-right corresponds to increasingly localized connectivity. The right FourRooms panel ablates over the depth of the bottleneck corridor, measured in grid cells. Lower error means that the planner is coping better with either random sparsification in GARNET or geometric clogging in FourRooms. In corridor-heavy FourRooms layouts, topology-aware prioritization remains substantially more stable than Standard-PS and excels on long-corridor bottlenecks.}
\label{fig:d-rq5-bottlenecks}
\end{figure}

We spend our remaining supporting sweeps to this subsection. Together, Figs.~\ref{fig:d-rq2-beta}--\ref{fig:d-rq5-bottlenecks} provide the broader experimental ground, showing how to calibrate topology strength, how tolerable backward mixing is, and in which stochastic or navigational regimes the topology signal is most effective.

\appendixsubsection{Post-review audits}
\label{app:d-post-review}

This subsection collects the audits added after review, each backed by stored evidence from the released code, and the tables quote that evidence directly.

\appendixsubsubsection{GPS-H2 comparison.}

Among topology-aware comparators, the GPS partition planner of Wingate and Seppi~\citep{wingate2005prioritization} sits closest to our setting. We implement its H2 branch, denoted GPS-H2, which selects a partition block by the normalized priority $H_{2,\upsilon}(s)=B(s)+\upsilon V(s)$ with $\upsilon=1-\gamma=0.05$, then solves the block with a fixed-order Gauss-Seidel sweep. A value term keeps the priority on the scale of a one-step residual, while under nonstationarity the block solve can revisit stale pre-shift neighborhoods, which is the mechanism this comparison tests. Standard-PS and GTA-PS freeze their top-$m$ choices and commit simultaneous updates, while GPS-H2 keeps its block solve.

On FourRooms-104, GTA-PS lowers shift-end error against Standard-PS at every width by $22.2\%$ to $41.4\%$, and against GPS-H2 for $m\ge2$ by $29.4\%$ to $46.3\%$ (Table~\ref{tab:d-gps-fourrooms}). In the low-stochasticity GARNET regime, GPS-H2 leads at $m\le2$ while GTA-PS overtakes it from $m=4$ (Table~\ref{tab:d-gps-garnet}). GPS-H2 is ahead at $m=1$ by $0.091$ absolute in FourRooms, so batch width decides the comparison.

\begin{table}[h]
\centering
\small
\caption{\footnotesize FourRooms-104 shift-end $L_1$ value error across frozen batch widths $m$. Protocol: oracle DP, six reward fields, 25 replanning calls of 20 backups per shift, $q=|\mathcal{S}|$ for GTA-PS, 100 paired seeds, 95\% intervals.}
\label{tab:d-gps-fourrooms}
\begin{tabular}{cccc}
\toprule
$m$ & Standard-PS & GPS-H2 & GTA-PS \\
\midrule
1 & $1.780\pm0.081$ & $0.952\pm0.073$ & $1.043\pm0.062$ \\
2 & $1.817\pm0.081$ & $1.793\pm0.078$ & $1.133\pm0.059$ \\
4 & $1.883\pm0.087$ & $2.163\pm0.071$ & $1.161\pm0.058$ \\
8 & $1.922\pm0.085$ & $2.234\pm0.075$ & $1.304\pm0.063$ \\
16 & $2.004\pm0.092$ & $2.209\pm0.069$ & $1.559\pm0.079$ \\
\bottomrule
\end{tabular}
\end{table}

\begin{table}[h]
\centering
\small
\caption{\footnotesize Low-stochasticity GARNET-75 shift-end $L_1$ error under the same protocol: four actions, branching factor 3, neighborhood rate 0.15, localized reward support 3, exact-DP host, 100 paired seeds.}
\label{tab:d-gps-garnet}
\begin{tabular}{ccc}
\toprule
$m$ & GPS-H2 & GTA-PS \\
\midrule
1 & $1.279\pm0.050$ & $1.771\pm0.076$ \\
2 & $1.625\pm0.058$ & $1.756\pm0.077$ \\
4 & $1.830\pm0.070$ & $1.683\pm0.071$ \\
8 & $2.081\pm0.084$ & $1.431\pm0.057$ \\
16 & $2.141\pm0.088$ & $1.233\pm0.046$ \\
\bottomrule
\end{tabular}
\end{table}

\appendixsubsubsection{Compact learned abstraction.}
This paired study replaces the identity abstraction with one learned compact tabular model. A deterministic linear projection maps the 104 states to abstract nodes, and both planners plan over the same abstract model with the same lifted-value evaluation, differing only in the backup ordering. Values are 25th-call $L_1$ error with $m=1$ and 100 paired seeds.

\begin{table}[h]
\centering
\small
\caption{\footnotesize Last-call $L_1$ value error under a shared learned compact abstraction. The result supports compatibility with one compact tabular representation; general function approximation and partial observability await separate study.}
\label{tab:d-compact-abstraction}
\begin{tabular}{cccc}
\toprule
Abstract states & Linear VaST-VI & GTA-PS-FA & Mean reduction \\
\midrule
24 & $4.918\pm0.105$ & $5.100\pm0.121$ & $-3.7\%$ \\
40 & $4.384\pm0.095$ & $4.436\pm0.112$ & $-1.2\%$ \\
64 & $2.873\pm0.112$ & $2.694\pm0.126$ & $6.2\%$ \\
\bottomrule
\end{tabular}
\end{table}

GTA-PS-FA matches the linear ordered VI baseline within overlapping 95\% intervals at 24 and 40 abstract states ($-3.7\%$ and $-1.2\%$ mean change) and reduces error by 6.2\% at 64. The mechanism transfers to a coarser representation once the abstraction grows fine enough to expose topological structure, and the claim spans exactly this setting.

\appendixsubsubsection{Frontier and cutoff diagnostics.}
Diagnostics below characterize the computed frontier. Under the deployed residual gate, frontier-only states act as candidates, while executed backups draw from the residual queue. Setup: FourRooms-104 with the exact-DP host, $q=20$, uniform starts, and 30 seeds. We inspect the frontier right after the shifted reward root $s_0$ is backed up; the outside-$K$ set holds the states outside its $K$-hop backward ball.

Whenever the exterior frontier holds states, it lines up with the oracle ranking, with mean nDCG above 0.99. Realized cutoff scores vary strongly across shifts while adjacent scores co-vary, so the magnitude of the topology bonus is an instance-level property. These measurements establish useful frontier mass; an executed-backup utility claim would need a matched-time ablation comparing residual-gated selection with frontier-inclusive selection.

\begin{table}[h]
\centering
\small
\caption{\footnotesize Exterior frontier size and ranking quality after the reward root is backed up, 30-seed 95\% intervals. nDCG compares the frontier order with the oracle discounted residual ranking outside the $K$-ball.}
\label{tab:d-frontier}
\begin{tabular}{cccc}
\toprule
Barrier $K$ & $\mathrm{Vol}_K$ & Top-20 states outside $K$ & Outside-$K$ nDCG \\
\midrule
1 & $4.30\pm0.70$ & $15.70\pm0.70$ & $0.993\pm0.005$ \\
2 & $9.60\pm1.99$ & $10.60\pm1.75$ & $0.991\pm0.005$ \\
3 & $16.07\pm3.26$ & $5.33\pm2.32$ & $0.993\pm0.008$ \\
4 & $22.93\pm4.11$ & $1.50\pm1.59$ & $0.994\pm0.008$ \\
5 & $29.87\pm5.64$ & $0.33\pm0.48$ & $1.000\pm0.000$ \\
6 & $37.10\pm7.58$ & $0.00\pm0.00$ & -- \\
7 & $44.70\pm9.20$ & $0.00\pm0.00$ & -- \\
8 & $53.27\pm10.19$ & $0.00\pm0.00$ & -- \\
\bottomrule
\end{tabular}
\end{table}

\begin{table}[h]
\centering
\small
\caption{\footnotesize Realized top-$q$ cutoff sensitivity across localized reward shifts, with $q=20$ and $T_{\mathrm{shift}}=25$. Reward met counts seeds whose shift arrives while the reward root is active. Values are 30-seed means with 95\% intervals.}
\label{tab:d-tau-cutoff}
\begin{tabular}{ccccc}
\toprule
Shift & Reward met & $\tau_{20}$ & $\tau_{21}$ & Cutoff gap \\
\midrule
1 & 30/30 & $1.225\pm0.100$ & $1.193\pm0.096$ & $0.031\pm0.033$ \\
2 & 30/30 & $26.259\pm14.369$ & $25.996\pm14.371$ & $0.263\pm0.532$ \\
5 & 29/30 & $13.147\pm5.639$ & $13.079\pm5.611$ & $0.067\pm0.081$ \\
10 & 27/30 & $16.531\pm14.715$ & $16.465\pm14.681$ & $0.067\pm0.087$ \\
20 & 27/30 & $13.348\pm6.099$ & $13.106\pm5.676$ & $0.242\pm0.946$ \\
30 & 28/30 & $15.945\pm12.062$ & $15.878\pm12.060$ & $0.067\pm0.091$ \\
40 & 26/30 & $13.916\pm9.337$ & $13.831\pm9.272$ & $0.085\pm0.125$ \\
\bottomrule
\end{tabular}
\end{table}

\appendixsubsubsection{Scheduler checkpoints.}
We record scheduler statistics along fixed-reward DP trajectories with $q=|\mathcal{S}|$, $T=50$ backups, and Arnoldi after each policy improvement (30 seeds, point estimates). Scheduler parameters read $(\beta_1,\beta_2)=(0.1596,1.3615)$ for FourRooms and $(0.3469,0.2690)$ for GARNET.

\begin{table}[h]
\centering
\small
\caption{\footnotesize Mean SLEM and scheduler weight at the start, midpoint, and end of a fixed-reward trajectory. SLEM near one at small sizes tracks a premature policy assessment and can shift as the policy settles.}
\label{tab:d-slem-beta}
\begin{tabular}{lccc}
\toprule
Environment & $t=0$ & $t=T/2$ & $t=T$ \\
\midrule
FourRooms-20 & $0.9611$ / $0.2236$ & $0.9237$ / $0.2296$ & $0.8372$ / $0.2434$ \\
FourRooms-40 & $0.9811$ / $0.2204$ & $0.9856$ / $0.2197$ & $0.9408$ / $0.2269$ \\
FourRooms-68 & $0.9900$ / $0.2190$ & $0.9948$ / $0.2182$ & $0.9913$ / $0.2188$ \\
FourRooms-104 & $0.9937$ / $0.2184$ & $0.9978$ / $0.2178$ & $0.9957$ / $0.2181$ \\
GARNET-75 & $0.9684$ / $0.1043$ & $0.9761$ / $0.1016$ & $0.9789$ / $0.1007$ \\
GARNET-200 & $0.9636$ / $0.1060$ & $0.9675$ / $0.1046$ & $0.9734$ / $0.1026$ \\
\bottomrule
\end{tabular}
\end{table}

\begin{table}[h]
\centering
\small
\caption{\footnotesize Mean topology mass $B_t=\norm{\beta_t\tau_t}_1$ and total priority mass $M_t=\norm{\xi_t+\beta_t\tau_t}_1$ at the same checkpoints. The topology contribution varies mainly with $\tau_t$, with $\beta_t$ playing a secondary role.}
\label{tab:d-priority-mass}
\begin{tabular}{lccc}
\toprule
Environment & $t=0$ & $t=T/2$ & $t=T$ \\
\midrule
FourRooms-20 & $65.862$ / $71.174$ & $234.542$ / $256.385$ & $176.441$ / $204.324$ \\
FourRooms-40 & $94.530$ / $102.265$ & $435.271$ / $468.805$ & $325.239$ / $362.997$ \\
FourRooms-68 & $115.477$ / $124.988$ & $533.318$ / $567.564$ & $530.407$ / $577.551$ \\
FourRooms-104 & $128.367$ / $138.968$ & $705.985$ / $744.238$ & $672.026$ / $724.554$ \\
GARNET-75 & $21.260$ / $24.322$ & $87.278$ / $98.702$ & $99.749$ / $112.968$ \\
GARNET-200 & $27.859$ / $31.854$ & $171.764$ / $195.969$ & $202.778$ / $229.852$ \\
\bottomrule
\end{tabular}
\end{table}

Checkpoints show the scheduler operating consistently: the product $\beta_t\tau_t$ stays active even when SLEM approaches one, and the topology mass tracks the field, with the weight playing a secondary role. In small tasks the global SLEM trend declines as the greedy policy concentrates flow on critical routes, which raises conductance. One transition provides sparse evidence for the sign of a SLEM change; regional statistical control would settle that question, and we leave it for future work. Consequently, the scheduler adjusts $\beta_t$ gradually and keeps topology engaged throughout.

\appendixsubsubsection{Solver crossover.}
Comparing cached LU plus Woodbury corrections with restarted GMRES, the rank-8 topology-update audit runs on GARNET instances with 100 seeds and three timing repeats per seed. It isolates the component update from end-to-end planning.

\begin{table}[h]
\centering
\small
\caption{\footnotesize Rank-8 topology-update time including the $\tau$ refresh. Entries are means with 95\% Student-$t$ intervals. The crossover near 650 states is specific to this implementation and machine.}
\label{tab:d-solver}
\begin{tabular}{ccc}
\toprule
States & LU + Woodbury (ms) & GMRES (ms) \\
\midrule
400 & $21.58\pm0.46$ & $36.59\pm1.27$ \\
600 & $41.72\pm0.86$ & $44.63\pm1.26$ \\
650 & $46.97\pm0.81$ & $47.19\pm1.42$ \\
700 & $54.02\pm0.94$ & $49.71\pm1.54$ \\
750 & $63.99\pm1.19$ & $56.10\pm1.79$ \\
800 & $68.71\pm1.08$ & $55.20\pm1.41$ \\
850 & $79.04\pm1.22$ & $60.67\pm1.52$ \\
900 & $85.82\pm1.15$ & $62.92\pm1.49$ \\
\bottomrule
\end{tabular}
\end{table}

LU plus Woodbury leads through 650 states, where the intervals overlap, and GMRES takes the lead from 700 states onward. Growth differs mainly in the constant term, so the switch is implementation-specific; the deployed code keeps the sparse LU path for small instances and GMRES above, with the cutoff set conservatively around the measured meeting point.

\clearpage
\appendixheading{Miscellaneous Supplement}
\label{app:e}

\appendixsubsection{Dyna/DP + SPS (GTA-PS) pseudocode}
\label{app:e-dp-dyna}

To deeply specifying the algorithms, we provide detailed pseudocodes describing how the host agent interacts with the module in each setting of DP and Dyna \ref{app:b-host-intel}

\begin{algorithm}[!ht]
\caption{\footnotesize DP + SPS (GTA-PS)}
\small
\begin{algorithmic}[1]
\Statex \textbf{Require:} $A^{\mathrm{PS}}_0(s,a) \gets \gamma\sum_{s'} p(s'|s,a)\,V_0(s') - \max_{a'}\gamma\sum_{s'} p(s'|s,a')\,V_0(s')$; \quad module temperature $\lambda>0$; \quad $\pi^{\mathrm{PS}}_0(a|s) \gets \mathrm{softmax}(A^{\mathrm{PS}}_0(s,\cdot)/\lambda),\ \forall s$
\While{not converged}
  \If{\textsc{RewardShifted}()}
    \State $r \mathrel{+}= \delta\mathbf{r}$;\quad \textsc{RewardUpdate}$(\delta\mathbf{r})$
  \EndIf
  \For{$k = 1$ \textbf{to} $K$} \Comment{$K$ Bellman backups available per replanning}
    \State $s \gets \textsc{PopAndPrioritize}()$
    \If{$s = \varnothing$} \textbf{break} \EndIf
    \State $\Delta V(s) \gets \textsc{BellmanBackup}(s, P, r, V)$ \Comment{host backup, $O(|A|\cdot\mathrm{deg})$}
    \State \textsc{UpdateValue}$(s, V(s))$ \Comment{sync current host value into the module}
    \State $A^{\mathrm{PS}}(s,a) \gets \gamma\sum_{s'} p(s'|s,a)\,V(s') - \max_{a'}\gamma\sum_{s'} p(s'|s,a')\,V(s')$ for each $a$ \Comment{$O(|A|\cdot\mathrm{deg})$}
    \State \textsc{NotifyBackup}$(s, A^{\mathrm{PS}}(s,\cdot), \Delta V(s))$ \Comment{encapsulates local policy/topology update; no dense queue refresh}
    \State \textsc{PushPredecessors}$(s,\, \Delta V(s))$
  \EndFor
\EndWhile
\end{algorithmic}
\end{algorithm}

\begin{algorithm}[!ht]
\caption{\footnotesize Dyna (one-step TD) + SPS (GTA-PS)}
\small
\begin{algorithmic}[1]
\Statex \textbf{Require:} module temperature $\lambda>0$;\quad TD learning rate $\eta_{\mathrm{td}}\in(0,1]$;\quad $A^{\mathrm{PS}}_0(s,a) \gets Q_0(s,a) - \max_{a'}Q_0(s,a')$;\quad $\pi^{\mathrm{PS}}_0(a|s) \gets \mathrm{softmax}(A^{\mathrm{PS}}_0(s,\cdot)/\lambda),\ \forall s$
\While{not converged}
  \State $(s,a,r,s') \gets \textsc{RealStep}()$ \Comment{interact with environment; $a \sim \pi_t$ (host policy)}
  \State Update $\hat{p}(\cdot|s,a), \hat{r}$ from $(s,a,s')$ \Comment{model learning; uses empirical model}
  \State $\xi_s \gets r + \gamma \max_{a'} Q(s',a') - Q(s,a)$;\quad $Q(s,a) \mathrel{+}= \eta_{\mathrm{td}}\,\xi_s$
  \State \textsc{UpdateModel}$(\hat{p}, \hat{r})$;\quad \textsc{UpdateValue}$(s, \max_a Q(s,a))$
  \State \textsc{PushPredecessors}$(s, |\xi_s|)$ \Comment{real step activates local residuals through the module interface}
  \If{\textsc{RewardShifted}()} \textsc{RewardUpdate}$(\delta\mathbf{r})$ \EndIf
  \For{$k = 1$ \textbf{to} $K$} \Comment{$K$ simulated planning backups available}
    \State $s \gets \textsc{PopAndPrioritize}()$
    \If{$s = \varnothing$} \textbf{break} \EndIf
    \State $a \gets \arg\max_{a'} Q(s,a')$
    \State $(r_{\mathrm{sim}}, s'_{\mathrm{sim}}) \sim \hat{p}(\cdot|s,a)$
    \State $\xi_s \gets r_{\mathrm{sim}} + \gamma\max_{a'}Q(s'_{\mathrm{sim}},a') - Q(s,a)$;\quad $Q(s,a) \mathrel{+}= \eta_{\mathrm{td}}\,\xi_s$
    \State \textsc{UpdateValue}$(s, \max_a Q(s,a))$
    \State $A^{\mathrm{PS}}(s,a) \gets Q(s,a) - \max_{a'}Q(s,a')$ for each $a$
    \State \textsc{NotifyBackup}$(s, A^{\mathrm{PS}}(s,\cdot), |\xi_s|)$ \Comment{encapsulates local policy/topology update; no dense queue refresh}
    \State \textsc{PushPredecessors}$(s, |\xi_s|)$
  \EndFor
\EndWhile
\end{algorithmic}
\end{algorithm}

\newpage
\appendixsubsection{Additional ablations}
\label{app:e-ablations}

We report a focused subset of the ablation study, highlighting the most influential insights, as additional appendix supplements.
These ablations isolate short-horizon error tracking, host-level reward accumulation, and the effect of the mixing-time threshold.
The remaining ablations, implementation details, and extended results are provided in the code supplement.

\begin{figure}[!ht]
\centering
\includegraphics[width=0.85\linewidth]{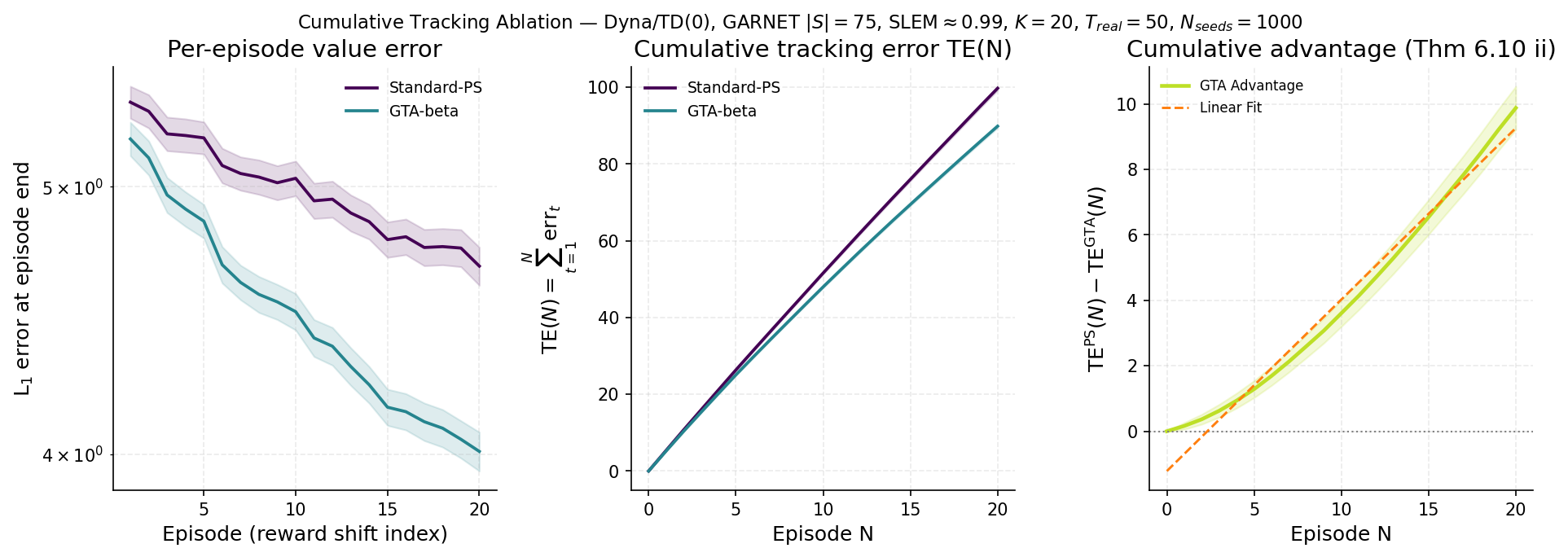}
\caption{\footnotesize GARNET cumulative tracking ablation. The left panel shows per-episode value error, the middle panel accumulates the tracking error over episodes, and the right panel plots the theorem-facing cumulative advantage. GTA-PS$^\beta$ reduces the episode-end error faster than Standard-PS, and the gap widens rather than collapsing. The cumulative tracking error $\mathrm{TE}(N)$ gap increases throughout the series, empirically backing the analysis of \ref{app:proof-cumulative-advantage}. The advantage curve remains close to linear with a positive slope, matching the intended interpretation of \ref{thm:cumulative-tracking}. This is the cleanest evidence that GTA-PS is doing useful work when reward shifts are frequent and the planner must repeatedly recover from fresh residual fields.}
\label{fig:e-cumulative-ablation}
\end{figure}

\begin{figure}[!ht]
\centering
\includegraphics[width=0.85\linewidth]{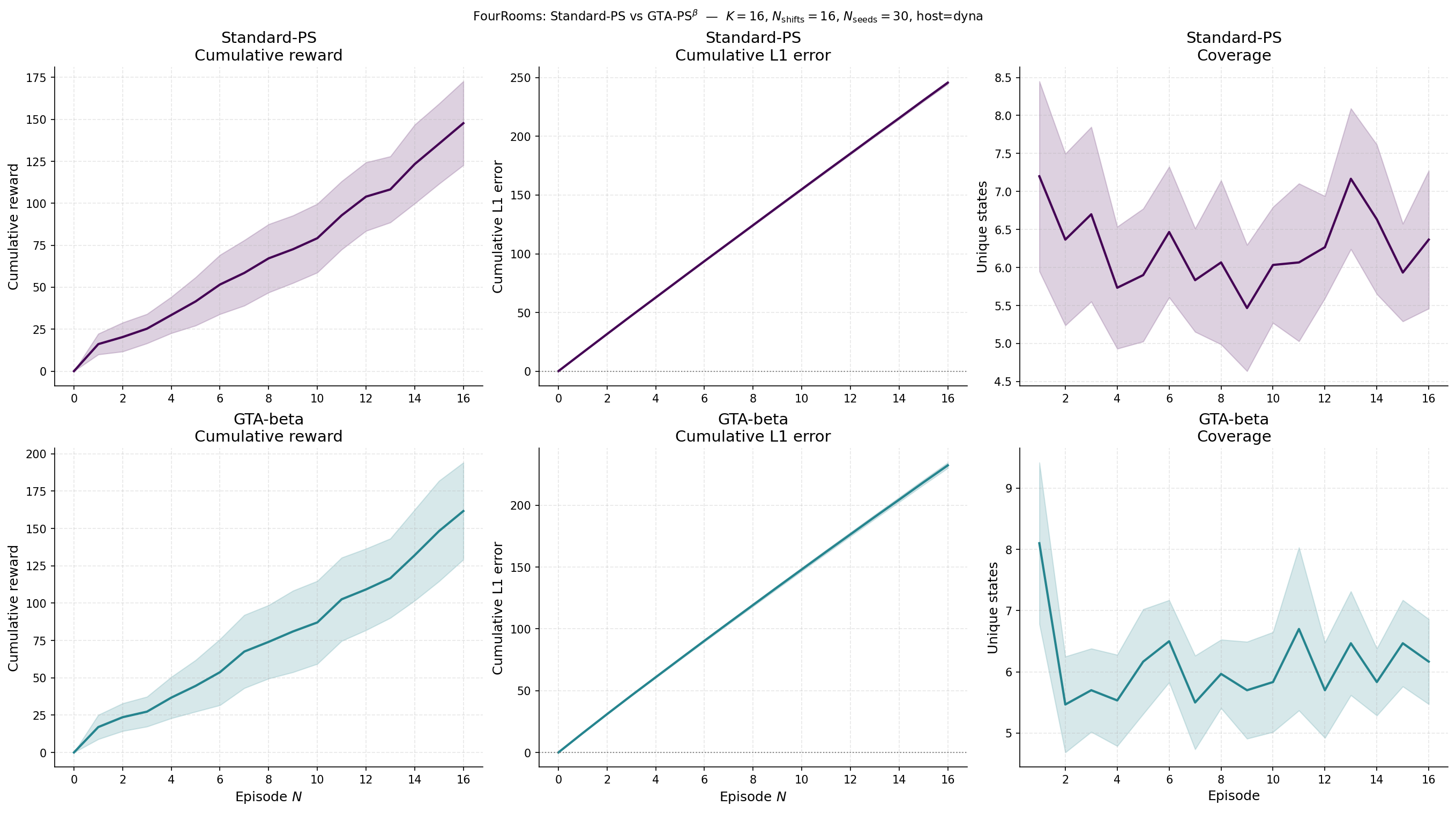}
\caption{\footnotesize FourRooms ablation under a Dyna host. The top row reports Standard-PS and the bottom row reports GTA-PS$^\beta$ on the same seeds, budgets, and shift schedule. GTA-PS$^\beta$ achieves higher cumulative reward while simultaneously lowering cumulative $\ell_1$ error, so the improvement was multi-faceted. The coverage traces remain in a comparable band for both methods, which shows that GTA-PS does not exploit aggressively. In the corridor-heavy FourRooms geometry, this is the right signal to look at because the topology module should help recover from bottlenecked propagation.}
\label{fig:e-fourrooms-ablation}
\end{figure}

\begin{figure}[!ht]
\centering
\includegraphics[width=\linewidth]{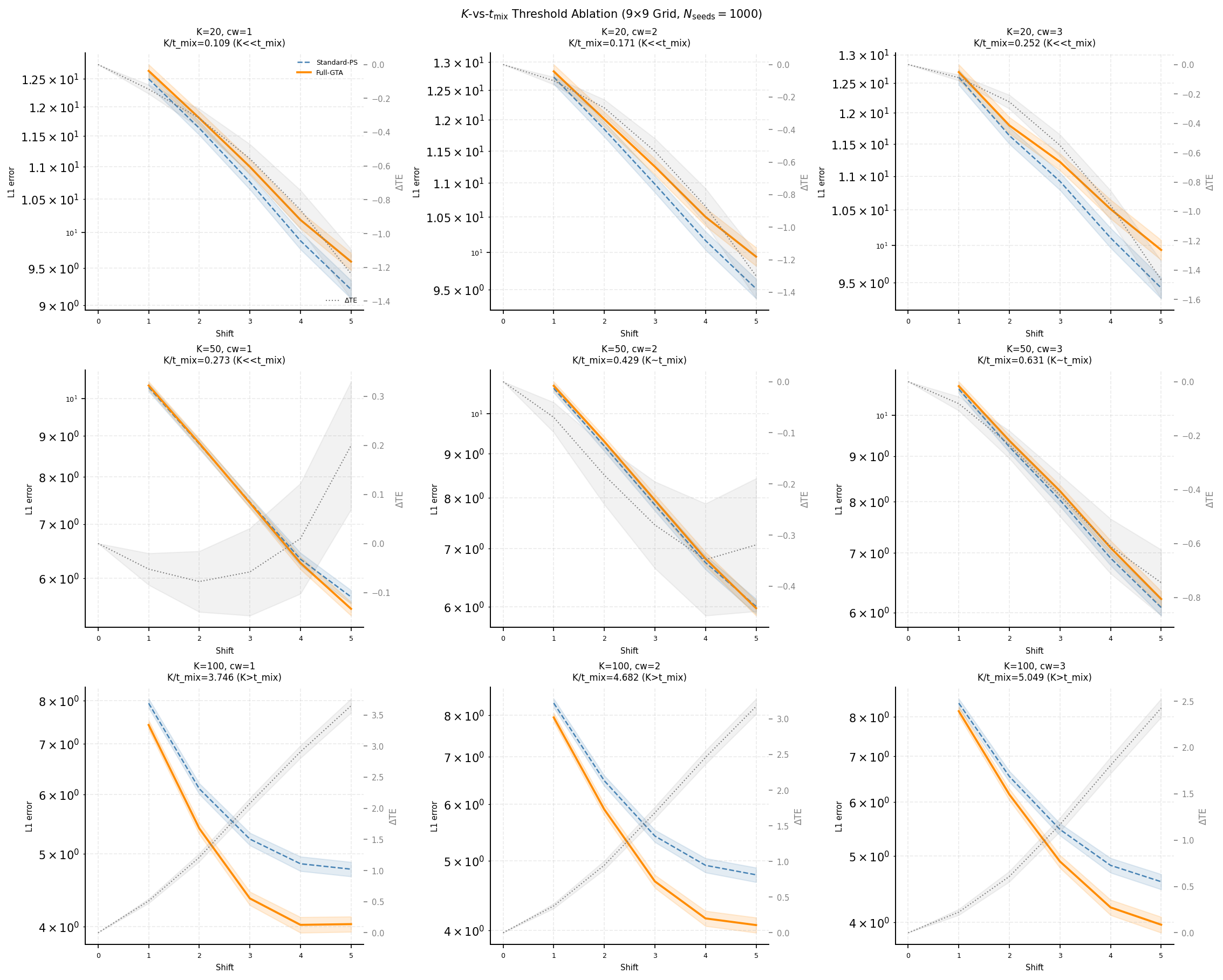}
\caption{\footnotesize Threshold test against the $K$ versus $t_{\mathrm{mix}}$ claim. Each panel fixes $K=20,50,100$ and varies the corridor width, which changes the mixing scale and therefore the ratio $K/t_{\mathrm{mix}}$. When $K \ll t_{\mathrm{mix}}$, Standard-PS performs significantly worse than GTA-PS. As the ratio increases, the curves cross and the full GTA variant becomes competitive or superior. GTA-PS performs well on narrow corridors, empirically backing the proactiveness intuition, and it also excels when given more compute, enabling scaling capabilities.}
\label{fig:e-threshold-ablation}
\end{figure}

\end{document}